\def\year{2022}\relax
\documentclass[letterpaper]{article} 
\usepackage{aaai22}  
\usepackage{times}  
\usepackage{helvet}  
\usepackage{courier}  
\usepackage[hyphens]{url}  
\usepackage{graphicx} 
\usepackage{natbib}  
\usepackage{caption} 
\DeclareCaptionStyle{ruled}{labelfont=normalfont,labelsep=colon,strut=off} 
\usepackage{booktabs}
\usepackage{microtype}
\usepackage{amsmath}
\usepackage{amsfonts}
\usepackage{array}
\usepackage{makecell}
\usepackage{algorithm,algpseudocode}

\usepackage{newfloat}
\usepackage{listings}
\floatstyle{ruled}
\newfloat{listing}{tb}{lst}{}
\floatname{listing}{Listing}
\title{Style Mixing and Patchwise Prototypical Matching for One-Shot Unsupervised Domain Adaptive Semantic Segmentation}
\author{
	Xinyi Wu,{\textsuperscript{\rm 1}}
	Zhenyao Wu,\textsuperscript{\rm 1}
	Yuhang Lu,\textsuperscript{\rm 1}
	Lili Ju,\textsuperscript{\rm 2,$*$}
	Song Wang\textsuperscript{\rm {1,}}\thanks{Co-corresponding authors. \\Code is available at {https://github.com/W-zx-Y/SM-PPM}.}\\
}

\affiliations {
	\textsuperscript {\rm 1}Department of Computer Science and Engineering, University of South Carolina, USA \\
	\textsuperscript {\rm 2}Department of Mathematics, University of South Carolina, USA\\
	\{xinyiw, zhenyao, yuhang\}@email.sc.edu,  ju@math.sc.edu, songwang@cec.sc.edu
}

\usepackage{bibentry}
\begin{document}
\maketitle
\begin{abstract}
	In this paper, we tackle the problem of one-shot unsupervised domain adaptation (OSUDA) for semantic segmentation where the segmentors only see one unlabeled target image during training. In this case, traditional unsupervised domain adaptation models usually fail since they cannot adapt to the target domain with over-fitting to one (or few) target samples. To address this problem, existing OSUDA methods usually integrate a style-transfer module to perform domain randomization based on the unlabeled target sample, with which multiple domains around the target sample can be explored during training. However, such a style-transfer module relies on an additional set of images as style reference for pre-training and also increases the memory demand for domain adaptation. Here we propose a new OSUDA method that can effectively relieve such computational burden. Specifically, we integrate several style-mixing layers into the segmentor which play the role of style-transfer module to stylize the source images without introducing any learned parameters. Moreover, we propose a patchwise prototypical matching (PPM) method to weighted consider the importance of source pixels during the supervised training to relieve the negative adaptation. Experimental results show that our method achieves new state-of-the-art performance on two commonly used benchmarks for domain adaptive semantic segmentation under the one-shot setting and is more efficient than all comparison approaches.
\end{abstract} 
\section{Introduction}
Semantic segmentation is a basic computer vision task to identify the semantic category of each pixel from a set of pre-defined categories.
It benefits a variety of tasks such as autonomous driving~\cite{treml2016speeding}, medical imaging~\cite{taghanaki2021deep}, and image editing~\cite{aksoy2018semantic}. Using deep learning, state-of-the-art semantic segmentation can be obtained in the form of good prediction at each pixel by training the well-designed segmentor network~\cite{chen2017deeplab} on large-scale labeled datasets and testing on the same domain. However, constructing datasets for such a dense prediction task is both very time-consuming and labor-intensive, which makes it often impossible to prepare  a high-quality large-scale labeled training set for all different scenarios/domains, e.g., different cities or different illumination conditions. As a result, the generalization ability of a trained model is limited, i.e., it usually suffers from a drastic performance drop on an unseen testing domain due to the different data distributions from the training set. 
\begin{figure*}[htbp]
	\centering
	\includegraphics[width=1\linewidth]{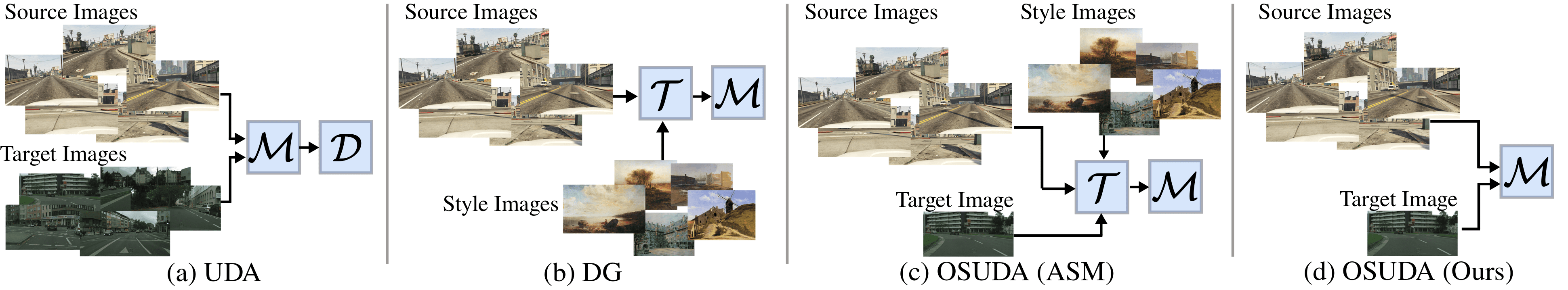}
	\caption{An illustration of general UDA, DG and OSUDA for semantic segmentation. The difference is mainly in the number of unlabeled target samples that are used for adaptation. Here, $\mathcal M$, $\mathcal D$ and $\mathcal T$ represent the segmentor, discriminator and style-transfer module, respectively.}
	\label{fig:fig1}
\end{figure*}

Recently, some unsupervised domain adaptation approaches were proposed to overcome the domain discrepancy and reduce the demand for labeled data in new unseen test domains.  Synthetic-to-real is a common setting in domain adaptive semantic segmentation, which was first proposed by~\citeauthor{hoffman2016fcns}~\shortcite{hoffman2016fcns}. In this setting, source domains with labeled synthetic data~\cite{richter2016playing,RosCVPR16} are constructed by using computer graphics techniques and in the meantime, a sufficient number of samples in the target domain are also provided without labels. In many applications, such as medical imaging, a large collection of unlabeled target data may be unavailable or difficult to obtain, which leads to the introduction of a new setting, the one-shot unsupervised domain adaptation (OSUDA)~\cite{Luo2020ASM} for semantic segmentation. The difference of the general unsupervised domain adaptation (UDA), domain generalization (DG) and OSUDA is illustrated in Fig.~\ref{fig:fig1}. In this paper, we aim to tackle this challenging but practical setting of OSUDA.

Existing UDA approaches, especially those which employ discriminators to distinguish whether the content, e.g., image feature~\cite{hoffman2016fcns}, segmentation prediction~\cite{Tsai_adaptseg_2018} or entropy map~\cite{vu2019advent}, is from the source or target domains (Fig.~\ref{fig:fig1}(a)), are prone to over-fitting on only one target sample -- discriminators can easily distinguish the over-fit target domain from the source domain. Other style-transfer-based approaches cannot handle this one-shot setting either, since the source images can only be stylized by only one target sample. To solve this problem, Luo {et.al} proposed an adversarial style mining (ASM) algorithm~\cite{Luo2020ASM}, as illustrated in Fig.~\ref{fig:fig1}(c), by mutually optimizing the style-transfer module and the semantic segmentation network via an adversarial regime. However, the style-transfer module itself requires additional data for pre-training and also increases the demand for GPU memory for adaptation. 

In this paper, we propose a new OSUDA approach, as illustrated in Fig.~\ref{fig:fig1}(d), which does not require additional data to pre-train a style-transfer module and explicitly synthesizes stylized images for semantic segmentation. First, we design a style-mixing segmentor which can simultaneously augment the source domain conditioned on feature statistics of the target sample and produce the semantic segmentation results. In addition, to relieve the negative adaptation~\cite{li2020content}, i.e., not all source samples/pixels have a positive effect for domain adaptation, the source images are weightedly trained based on their similarity with patchwise prototypes of the sole target sample during domain adaptation. 

The main contributions of this paper are summarized as follows. We propose a simple and effective method for OSUDA semantic segmentation, which makes full use of the sole target image in two aspects: (1) implicitly stylizing the source domain in both image and feature levels; (2) softly selecting the source training pixels. No additional images and training parameters are introduced in the whole process. It is worth mentioning that, with a pre-trained model on the source domain, our method only needs 20 minutes (500 iterations) to adapt to the target domain and obtains comparable results to the current best OSUDA approach (200k iterations without a pre-trained model, and additional training iterations for style-transfer model). Experimental results on two commonly-used synthetic-to-real scenarios demonstrate the effectiveness and efficiency of the proposed method. 

\section{Related work} 
In this section, we briefly review the previous related works on UDA/OSUDA, DG, prototypical representation and style transfer, especially their applications to semantic segmentation.

\subsection{Unsupervised domain adaptation and domain generalization}
OSUDA is developed from the general UDA setting. The first UDA approach for semantic segmentation was proposed in~\citeauthor{hoffman2016fcns}~\shortcite{hoffman2016fcns} using feature-level adversarial learning and category-specific adaptation. After that, adversarial learning has been applied to UDA in feature level~\cite{chen2017no,hoffman2018cycada}, output space~\cite{Tsai_adaptseg_2018,pan2020unsupervised, zhang2017curriculum, perone2019unsupervised} and entropy of the prediction~\cite{vu2019advent,pan2020unsupervised} for alignment. Image translation is another approach for UDA~\cite{sankaranarayanan2018learning,li2019bidirectional, yang2020fda} by exploiting advanced image-to-image translation networks, e.g., CycleGAN~\cite{CycleGAN2017}, to reduce the domain discrepancy. Recently, multiple rounds of self-training with generated pseudo labels of the target domain samples was proved to be a powerful strategy to boost the adaptation performance~\cite{zou2018unsupervised,li2019bidirectional,zhang2019category}. However, these methods cannot be directly applied to the OSUDA setting due to the scarce of the target images.

Also related to OSUDA is the problem of domain generalization where the target domain is totally unknown. Based on the number of source domains involved during adaptive learning, existing DG approaches can be basically divided into multi-source DG~\cite{gong2013reshaping,dou2019domain} and single-source DG~\cite{pan2018two,yue2019domain,huang2021fsdr}. For multi-source DG, Zhou {et.al} proposed a MixStyle~\cite{zhou2021mixstyle} strategy to increase the domain diversity of the source domains. During training, two instances of different domains in a mini-batch are selected to synthesize novel domains leveraging the feature-level style statistics~\cite{huang2017adain}. Single-source DG is more challenging since less labeled source data is accessible for adaptation. A typical solution is to perform domain randomization~\cite{tobin2017domain} on the source training samples via image stylization or translation which can also be treated as a data/domain augmentation strategy. For example, several real-life images from ImageNet~\cite{deng2009imagenet} are picked as randomization references~\cite{yue2019domain} to adjust the source images or their domain invariant frequency components~\cite{huang2021fsdr}. 

\subsection{Prototypical representation}
Prototypes are defined as abstractions of essential semantic feature representations, which were popularly used in computer vision tasks recently. For example, 
\citeauthor{wang2019panet}~\shortcite{wang2019panet} design a prototype alignment regularization for few-shot semantic segmentation, where the class-specific prototypes are computed via a masked average pooling.  More recently, 
\citeauthor{zhang2021prototypical}~\shortcite{zhang2021prototypical} exploited the distances between the target features and the class-wise prototypes to re-weight the predicted probability for better self-training. In this paper, we calculate the prototypes of the patches of the sole target image to weigh the training pixels from the source domain.
\begin{figure*}[t]
	\centering
	\includegraphics[width=1.0\linewidth]{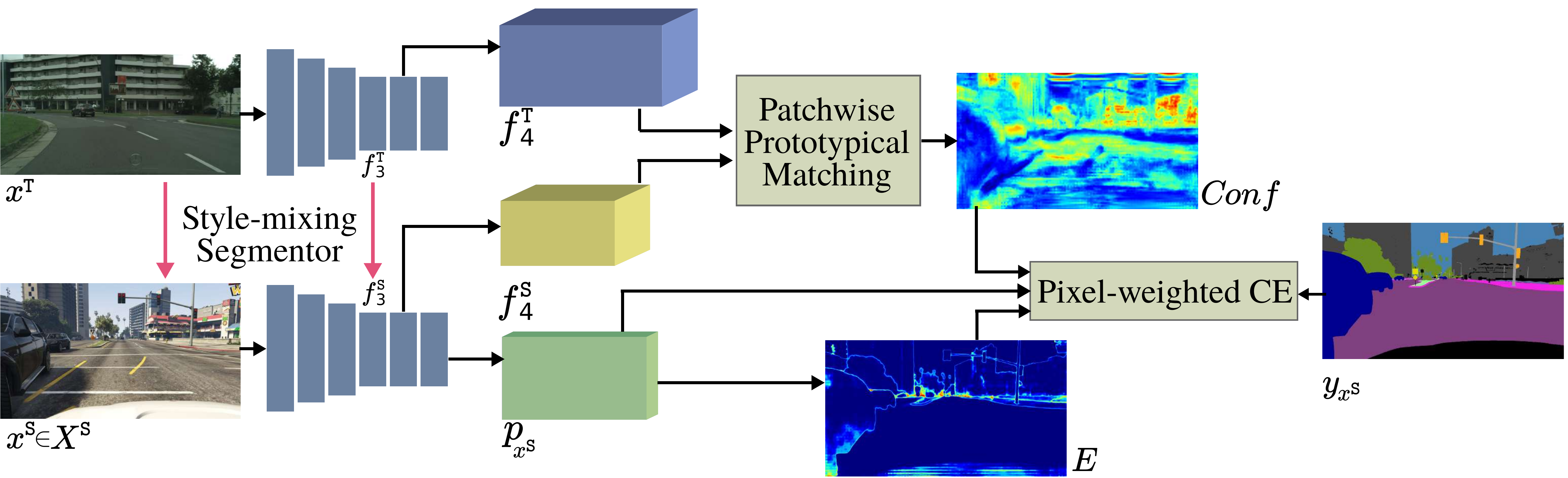}
	\caption{ An illustration of the proposed method for OSUDA semantic segmentation. The pink arrows indicate the positions that style-mixing operation is performed.}
	\label{fig:model}
\end{figure*}

\subsection{Style transfer} 
\citeauthor{gatys2015neural}~\shortcite{gatys2015neural} proposed a neural style-transfer algorithm to generate high-quality artistic images by separating and recombining the content and style of arbitrary images. Later style transfer has become an effective technique, which benefits several real-world applications such as makeup transfer and removal~\cite{chang2018pairedcyclegan} and virtual try-on~\cite{yang2020towards}. Our work is closely related to the adaptive instance normalization (AdaIN) proposed by \citeauthor{huang2017arbitrary}~\shortcite{huang2017arbitrary}, which transfers the mean and variance in the feature space in real-time. The main difference is that we don't synthesize the image with a decoder.

\section{Our approach}
Given labeled samples $(X^\mathtt S, Y^\mathtt S)$ from the source domain $\mathtt S$ and unlabeled samples $X^\mathtt T$ from the target domain $\mathtt T$, the goal of general UDA problem is to learn a mapping  $\mathcal G$  formulated as
\begin{equation}
\mathcal G(X^\mathtt S) \rightarrow Y^\mathtt S;\;\; \mathcal F(\mathtt S) \rightarrow \mathcal F(\mathtt T),
\end{equation}
where $\mathcal F$ represents any function to align the two domains, e.g., a discriminator. Different from general UDA, only one unlabeled target sample $x^\mathtt T \in X^\mathtt T$ is accessible in the OSUDA setting, which can be formulated as:
\begin{equation}
\mathcal G(X^\mathtt S|x^\mathtt T) \rightarrow Y^\mathtt S.
\end{equation}

The network architecture of the proposed one-shot unsupervised domain adaptive semantic segmentation method is illustrated in Fig.~\ref{fig:model}, which is composed of a style-mixing segmentor for both style transfer and semantic segmentation, and a patchwise prototypical matching module for weighting the pixels of the source domain. The details of the two main components and the objective functions are discussed below.

\subsection{Style-mixing segmentor}
For each iteration, the style-mixing segmentor first takes the target sample $x^{\mathtt T}$ as input in the evaluation mode to achieve target features using the current model parameters. Then, a sample $x^{\mathtt S}$ is randomly chosen from the source domain and fed into the segmentor in the training mode.
\begin{figure}[h]
	\centering
	\includegraphics[width=0.38\textwidth]{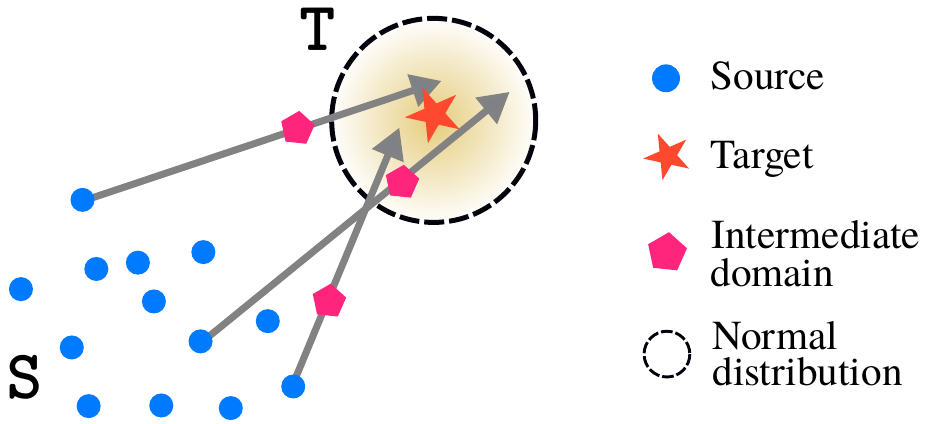}
	\caption{An illustration of the style-mixing operation. We first augment the target feature statistics by adding a perturbation sampled from normal distribution. Then some intermediate domains can be obtained by mixing the feature statistics of the source and the augmented target.}
	\label{fig:mix}
\end{figure}

Inspired by~\citeauthor{zhou2021mixstyle}~\shortcite{zhou2021mixstyle}, we mix the target features from the sole target image with the source features via instance normalization layers to obtain intermediate domain features. 
 
Following~\citeauthor{huang2017arbitrary}~\shortcite{huang2017arbitrary}, we compute the channel-wise spatial mean $\mu(.)$ and standard deviation $\sigma(.)$ of any given sample/feature $f \in \mathbb R^{ C\times H\times W}$ via
\begin{equation}
\mu(f) = \frac{1}{HW}\sum_{h=1}^{H}\sum_{w=1}^{W}f
\end{equation}
and
\begin{equation}
\sigma(f) = \sqrt{\frac{1}{HW}\sum_{h=1}^{H}\sum_{w=1}^{W}(f-\mu(f))^{2}+\epsilon},
\end{equation}
where $C$, $H$ and $W$ are channel, height and width of $f$, respectively. $\epsilon$ is set to $10^{-30}$.
Since the only one target sample is insufficient to describe the whole target feature distribution, we exploit more feature statistics centered around $f^ \mathtt T$ as shown in Fig.~\ref{fig:mix}.
Then, we mix the statistics of source and target domains and calculate the intermediate channel-wise mean $\gamma$ and standard deviation $\beta$ by
\begin{equation}
\gamma = \lambda \sigma ( f^ \mathtt S ) + ( 1- \lambda) \left (\sigma ( f^ \mathtt T) + r_{\sigma} \right ),
\end{equation}
\begin{equation}
\beta = \lambda \mu (f^ \mathtt S ) + (1- \lambda)\left ( \mu (f^ \mathtt T) + r_{\mu} \right ),
\end{equation}
where $\lambda \in \mathbb R^{C}$ are weights to balance the mixing operation which are randomly sampled from uniform distribution for each image/feature pair, $f^\mathtt S \in \{x^ \mathtt S, f^\mathtt S_{3}\}$ and $f^ \mathtt T \in \{x^ \mathtt T, f^\mathtt T_{3}\}$ with $f^\mathtt S_{3}$ and $f^\mathtt T_{3}$ denoting the source and target features achieved from \textit{layer3} respectively. Here we take $r_{\sigma} \sim N(0,\frac{|\sigma ( f^ \mathtt T)-\sigma ( f^ \mathtt S )|}{10})$ and $r_{\mu} \sim N(0,\frac{|\mu ( f^ \mathtt T)-\mu ( f^ \mathtt S )|}{10})$. 
The stylized source feature $\widehat{f^ \mathtt S}$ is then produced  by taking
\begin{equation}
\widehat{f^\mathtt S} = \gamma \left( \frac{f^\mathtt S-\mu (f^\mathtt S)}{\sigma (f^\mathtt S)} \right) + \beta.
\end{equation}

\subsection{Patchwise prototypical matching}
\citeauthor{li2020content}~\shortcite{li2020content} empirically found that some source samples could have negative effect on the adaptation. Based on this observation, they perform both image and pixel-level selections in the source domain to avoid the negative domain adaptation. However, both of their image and pixel-level selections are dependent on the distribution analysis of the target domain predictions, which are not applicable to our one-shot setting, i.e., the sole target image cannot correctly reflect the data distribution in the target domain and many categories are missing in this target sample. 
Inspired by~\citeauthor{li2020content}~\shortcite{li2020content}, we propose a patchwise prototypical matching  (PPM) by softly adjusting the weight of each source pixel during training according to their similarity with the target sample. We do not perform image-level selection since ``negative'' samples might also contain ``positive'' pixels for adaptation. 
\begin{figure}[htbp]
	\centering
	\includegraphics[width=1\linewidth]{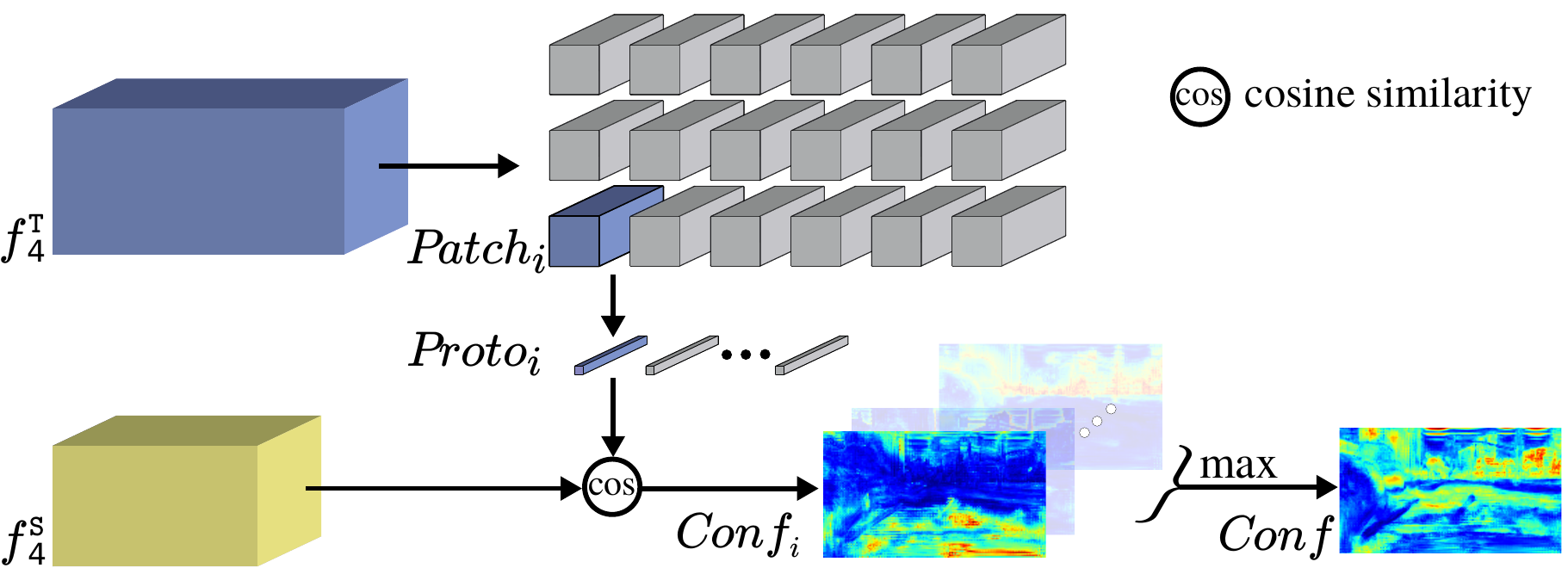}
	\caption{An illustration of the proposed patchwise prototypical matching.}
	\label{fig:conf}
\end{figure}

Specifically, we reshape the target image feature $f^\mathtt T_{4} \in \mathbb R^{C_4 \times H_4 \times W_4}$ obtained from \textit{layer4} of the segmentor into the form of patches $p^\mathtt T_{4} \in \mathbb R^{N \times C_4 \times P^2}$ as shown in Fig.~\ref{fig:conf}, where $H_4$, $W_4$, $C_4$ are the height, width and number of channels of $f^\mathtt T_{4}$, $P$ is the patch size and $N$ is the number of patches. There is no overlap between two patches. Then, we compute the prototype for each patch via: 
\begin{equation}\label{eq1}
Proto_i = \frac{1}{P^2}\sum_{s=1}^P \sum_{t=1}^P Patch_i(s,t),
\end{equation}
where $Proto_i \in \mathbb R^{C_4}$, $i \in [0, N-1]$ and $(s, t)$ specifies each position in the patch. We compute the similarity between each prototype and the source features as a confidence map $Conf_i \in \mathbb R^{C_4 \times H_4 \times W_4}$ for adaptation by
\begin{equation}\label{eq2}
Conf_i = {\mathcal F \left (f^ \mathtt S_{4}, Proto_i \right )},
\end{equation}
where $f^ \mathtt S_{4} \in \mathbb R^{C_4 \times H_4 \times W_4}$ is the source image feature obtained from \textit{layer4}, and we choose the cosine similarity as the distance function $\mathcal F$. We then perform a max operation across all prototypes to obtain the $Conf \in R^{H_4 \times W_4}$ for this source sample in this running iteration by
\begin{equation}\label{eq3}
Conf = \max_{i \in [0, N-1]} Conf_i. 
\end{equation}
The reason for using the prototypical representation of the target features to compute the confidence maps include: 1) it is more efficient than pixel-wise similarity computation; 2) due to the domain gap, the pixel-level similarity usually contains much more noise which can be relieved by using patchwise prototypes. 
Finally, the confidence map is rectified based on the entropy of the source prediction. Given the source prediction $p_{x^\mathtt S}$, its entropy map $E \in \mathbb R^{H_4 \times W_4}$ can be achieved via:
\begin{equation}\label{eq4}
E = - \frac{1}{\mathrm{log}(C)} \sum _{c = 1}^C \left ( \ p_{x^\mathtt S}^{(c)} \cdot \mathrm{log}(p_{x^\mathtt S}^{(c)}) \right ),
\end{equation}
where $C$ is the number of classes. Through this way, the rectified confidence map is achieved by
\begin{equation}\label{eq5}
\widehat{Conf} = Conf \cdot (1-E). 
\end{equation}
High entropy indicates low confidence for the prediction, therefore, $(1-E)$ can highlight the confident region based on the prediction. The thought behind this design is that the source should be confident enough to help the adaptation to the target. A detailed pipeline for the proposed PPM is given in Algorithm~\ref{algorithm}.
\begin{algorithm}
	\caption{Patchwise Prototypical Matching}
	\label{algorithm}
	\begin{algorithmic}[1]
		\Statex
		\textbf{Input}: {Source images $X^\mathtt S$; source labels $Y^\mathtt S$; one-shot target image $x^\mathtt T$; style-mixing segmentor $\mathcal M$ with the parameter $\theta$ and learning rate $lr$} \\
		\textbf{Output:} {Optimal $\theta^{*}$}
		\For{$x^\mathtt S \in X^\mathtt S$}
		\State \textbf{With} no gradients:
		\State \qquad  $(p_{x^{\mathtt T}}, f_3^\mathtt T, f_4^\mathtt T) = \mathcal M \left (x^{\mathtt T}, \mathrm{style}=\mathrm{None} \right )$;
		\State \qquad $Patch$ = rearrange($f_4^\mathtt T$);
		\State \qquad Compute $Proto$ via Eq.(\ref{eq1});
		\State $(p_{x^{\mathtt S}}, f_3^\mathtt S, f_4^\mathtt S) = \mathcal M \left (x^\mathtt S, \mathrm{style}=(x^\mathtt T, f_3^\mathtt T)\right )$;
		\State Compute the $E$ using the $p_{x^{\mathtt S}}$ via Eq. (\ref{eq4});
		\State Compute the $\widehat{Conf}$ using $f_4^\mathtt S$ and $Patch$ via Eqs. (\ref{eq2}), (\ref{eq3}) and (\ref{eq5});
		\State Update the parameter:
		
		$\theta \gets \theta - lr \nabla_\theta \mathcal L(p_{x^{\mathtt S}}, y_{x^{\mathtt S}}, \widehat{Conf}, E$);
		\EndFor
		\State Return $\theta$ as $\theta^{*}$
	\end{algorithmic}
\end{algorithm}  

\begin{table*}[htbp]
	\centering
	\caption{Quantitative comparison results for domain adaptation from GTA5 to Cityscapes. The per-category mIoU (\%) of the Cityscapes-val set are reported. For all methods with one-shot only setting denoted by O, the best results are presented in {\bf bold}, with the second best results \underline{underlined}.}
	\small
	\label{gta5-city}
		\begin{tabular}{p{15mm}|p{6mm}|*{18}{p{2.8mm}}p{5mm}|p{6mm}}
			\toprule
			Method &Extra & \rotatebox{90}{road} & \rotatebox{90}{sidewalk} & \rotatebox{90}{building} & \rotatebox{90}{wall} & \rotatebox{90}{fence} & \rotatebox{90}{pole} & \rotatebox{90}{traffic light \ } & \rotatebox{90}{traffic sign} & \rotatebox{90}{vegetation} & \rotatebox{90}{terrain} & \rotatebox{90}{sky} & \rotatebox{90}{person} & \rotatebox{90}{rider} & \rotatebox{90}{car} & \rotatebox{90}{truck} & \rotatebox{90}{bus} & \rotatebox{90}{train} & \rotatebox{90}{motorcycle} & \rotatebox{90}{bicycle}  & \bf mIoU\\ 
			\midrule
			Source only                            & - &75.8 &16.8 &77.2 &12.5 &21.0 &25.5 &30.1 &20.1 &81.3 &24.6 &70.3 &53.8 &26.4 &49.9 &17.2 &25.9 &6.5 &25.3 &36.0 &36.6\\
			\midrule
			Adaptseg     & O &77.7 &19.2 &75.5 &11.7 &6.4  &16.8 &18.2 &15.4 &77.1 &34.0 &68.5 &55.3 &\underline{30.9} &74.5 &23.7 &28.3 &\underline{2.9} &14.4 &18.9 &35.2\\
			CLAN            & O &77.1 &22.7 &78.6 &17.0 &14.8 &20.5 &\underline{23.8} &12.0 &80.2 &39.5 &74.3 &56.6 &25.2 &78.1 &29.3 &31.2 &0.0 &19.4 &16.7 &37.7\\
			ADVENT            & O &76.1 &15.1 &76.6 &14.4 &10.8 &17.5 &19.8 &12.0 &79.2 &39.5 &71.3 &55.7 &25.2 &76.7 &28.3 &30.5 &0.0 &23.6 &14.4 &36.1\\
			CBST     & O &76.1 &22.2 &73.5 &13.8 &18.8 &19.1 &20.7 &18.6 &79.5 &\underline{41.3} &\underline{74.8} &\underline{57.4} &19.9 &\underline{78.7} &21.3 &28.5 &0.0 &28.0 &13.2 &37.1\\
			CycleGAN          & O &80.3 &\underline{23.8} &76.7 &17.3 &18.2 &18.1 &21.3 &17.5 &\underline{81.5} &40.1 &74.0 &56.2 &\bf38.3 &77.1 &\underline{30.3} &27.6 &1.7 &\bf30.0 &22.2 &39.6\\
			OST        & O &\underline{84.3} &\bf27.6 &\bf80.9 &\bf24.1 &\underline{23.4} &\underline{26.7} &23.2 &\underline{19.4} &80.2 &\bf42.0 &\bf80.7 &\bf59.2 &20.3 &\bf84.1 &\bf35.1 &\bf39.6 &1.0 &\underline{29.1} &\underline{23.2} &\underline{42.3}\\
			\bf{Ours}                              & O &\bf85.0 &23.2 &\underline{80.4} &\underline{21.3} &\bf24.5 &\bf30.0 &\bf32.0 &\bf26.7 &\bf83.2 &34.8 &74.0 &57.3 &29.0 &77.7 &27.3 &\underline{36.5} &\bf5.0 &28.2 &\bf39.4 &\bf 42.8\\
			\midrule                
			DRPC            & S &- &- &- &- &- &- &- &- &- &- &- &- &- &- &- &- &- &- &- &42.5 \\
			FSDR             & S &89.3 &40.5 &79.1 &26.3 &27.8 &29.3 &33.7 &29.0 &83.0 &27.7 &76.0 &57.8 &27.5 &81.0 &32.3 &42.4 &16.8 &21.0 &30.2 &44.8\\
			\midrule
			ASM                 & O+S &89.5 &31.2 &81.3 &27.8 &22.8 &30.6 &32.8 &25.1 &82.6 &35.0 &76.7 &59.2 &26.6 &82.3 &27.7 &34.1 &0.9 &25.6 &29.6 &43.2\\
			\bottomrule
	\end{tabular}
\end{table*}

\subsection{Objective functions}
In general, the semantic segmentation task applies the cross entropy as the loss function: 
\begin{equation}
\mathcal L_{ce} = - \frac{1}{HW}\sum_{h,w}^{HW}  \sum _{c = 1}^C \left ( \ y_{x^\mathtt S}^{(h,w,c)} \cdot \mathrm{log}(p_{x^\mathtt S}^{(h,w,c)}) \right ),
\end{equation}
where $y_{x^\mathtt S}^{(h,w,c)}$ represents the one-hot encoding of the ground-truth label at position $(h,w)$ for the class $c$. In our approach, we employ the final confidence map $\widehat{Conf}$ to adjust the weight of each source sample in pixel-level via
\begin{equation}
\begin{split}
\mathcal L_{pce} = & - \frac{1}{HW}\sum_{h,w}^{HW}   ( \widehat{Conf}^{(h,w)} \cdot \\
& \sum _{c=1}^{C}(\ y_{x^\mathtt S}^{(h,w,c)} \cdot \mathrm{log}(p_{x^\mathtt S}^{(h,w,c)}) ) ).
\end{split}
\end{equation}
Finally, the whole network is trained with 
\begin{equation}
\mathcal L  = \alpha \mathcal L_{ce} + \mathcal L_{pce},
\end{equation}
where $\alpha$ is the balancing factor which is set to 0.5 in all experiments.

\section{Experiments}
\subsection{Datasets and evaluation metric}
We evaluate the proposed OSUDA semantic segmentation method  in two synthetic-to-real scenarios, i.e., GTA5~\cite{richter2016playing} $\rightarrow$ Cityscapes~\cite{Cordts2016Cityscapes} and SYNTHIA~\cite{RosCVPR16} $\rightarrow$ Cityscapes.
Both GTA5 and SYNTHIA datasets are treated as the source domains, where the former contains 24,966 images with a resolution of $1,914 \times 1,052$ and the latter contains 9,400 images with a resolution of $1,280 \times 760$. We use Cityscapes as the target domain which is split into 2,975/500/1,525 images for training/validation/testing purposes. We follow the one-shot setting in~\citeauthor{Luo2020ASM}~\shortcite{Luo2020ASM} where only one unlabeled target image is used for domain adaptation. In GTA5 $\rightarrow$ Cityscapes, 19 common categories are evaluated and in SYNTHIA $\rightarrow$ Cityscapes, 16 common categories are evaluated. We apply the Intersection over Union (IoU) as the evaluation metric.

\subsection{Implementation details} \label{exp_setting}
The proposed method is implemented using PyTorch trained on a single Nvidia 2080Ti GPU. We use the DeepLabV2-Res101~\cite{chen2017deeplab}  initialized with the source-only trained weights provided by~\citeauthor{Tsai_adaptseg_2018}~\shortcite{Tsai_adaptseg_2018} as the segmentor. The source images are resized to $1,280 \times 760$ and the one-shot target sample keeps its original size. We train the network using the SGD~\cite{bottou2010large} optimizer with a momentum of 0.9 and a weight decay of $5 \times 10^{-4}$. The initial learning rate is set to 2.5 $\times 10^{-5}$ and it is decreased gradually following the poly learning rate policy in~\citeauthor{Tsai_adaptseg_2018}~\shortcite{Tsai_adaptseg_2018}. The batch size is set to 1 and the whole network is trained for 500 iterations. Note that we even don't get access to all source images during domain adaptation. Images from Cityscapes validation set are resized to $1,024 \times 512$ for performance evaluation. We run each OSUDA experiment with the same 5 images as~\citeauthor{Luo2020ASM}~\shortcite{Luo2020ASM} (one for each time) and 5 times for each image. Finally, we report the average mIoU of the 25 runs computed using the model weights saved in the last running iteration. All the approaches are evaluated on the Cityscapes-val set.
\begin{table*}[htbp] 
	\centering
	\caption{Quantitative comparison results for domain adaptation from SYNTHIA to Cityscapes. The per-category mIoU (\%) (13 categories) and mIoU* (\%) (16 categories) of Cityscapes-val set are reported. For all methods with one-shot only setting denoted by O, the best results are presented in {\bf bold}, with the second best results \underline{underlined}.}
	\small
	\label{syn2city}
		\begin{tabular}{p{15mm}|p{6mm}|*{15}{p{3.4mm}}p{5mm}|p{6mm}p{6mm}}
			\toprule
			Method & Extra  & \rotatebox{90}{road} & \rotatebox{90}{sidewalk} & \rotatebox{90}{building} & \rotatebox{90}{wall} & \rotatebox{90}{fence} & \rotatebox{90}{pole} & \rotatebox{90}{traffic light \ } & \rotatebox{90}{traffic sign} & \rotatebox{90}{vegetation} & \rotatebox{90}{sky} & \rotatebox{90}{person} & \rotatebox{90}{rider} & \rotatebox{90}{car} & \rotatebox{90}{bus} & \rotatebox{90}{motorcycle} & \rotatebox{90}{bicycle}  & \bf mIoU & \bf mIoU*\\ 
			\midrule
			Source only                            & -&55.6 &23.8 &74.6 &-&-&-&6.1  &12.1 &74.8 &79.0 &55.3 &19.1 &39.6 &23.3 &13.7 &25.0 &38.6 &- \\
			\midrule
			Adaptseg     & O &64.1 &25.6 &\underline{75.3} &-&-&-&4.7  &2.7  &\underline{77.0} &70.0 &52.2 &20.6 &51.3 &\underline{22.4} &19.9 &22.3 &39.1 &-\\
			CLAN             & O &68.3 &26.9 &72.2 &-&-&-&5.1  &5.3  &75.9 &\underline{71.4} &54.8 &18.4 &65.3 &19.2 &\underline{22.1} &20.7 &40.4 &-\\
			ADVENT            & O &65.7 &22.3 &69.2 &-&-&-&2.9  &3.3  &76.9 &69.2 &\bf55.4 &\underline{21.4} &\bf77.3 &17.4 &21.4 &16.7 &39.9 &-\\
			CBST       & O &59.6 &24.1 &72.9 &-&-&-&5.5  &\underline{13.8} &72.2 &69.8 &\underline{55.3} &21.1 &57.1 &17.4 &13.8 &18.5 &38.5 &-\\
			OST       & O &\underline{75.3} &\underline{31.6} &72.1 &-&-&-&\underline{12.3} &9.3  &76.1 &71.1 &51.1 &17.7 &\underline{68.9} &19.0 &\bf26.3 &\underline{25.4} &\underline{42.8} &-\\
			\bf{Ours}                              & O &\bf79.3 &\bf35.3 &\bf75.9 &5.6 &16.6 &29.8 &\bf25.4 &\bf22.7 &\bf79.9 &\bf76.8 &54.6 &\bf23.5 &60.2 &\bf23.9 &21.2 &\bf36.6 &\bf47.3 &\bf41.4\\
			\midrule
			DRPC             & S &- &- &- &- &- &-&- &- &- &- &- &- &- &- &- &-  &- &37.6\\
			FSDR            & S &69.3 &34.9 &77.6 &7.9 &0.2 &29.4 &16.3 &19.2 &72.3 &76.3 &56.7 &22.1 &80.6 &41.5 &19.1 &29.3 &47.3  &40.8\\
			\midrule
			ASM                & O+S &85.7 &39.7 &77.1 &1.1 &0.0 &24.2 &2.1 &9.2 &76.9 &81.7 &43.4 &11.4 &63.9 &15.8 &1.6 &20.3 &40.7 &34.6\\
			\bottomrule
	\end{tabular}
\end{table*}

\begin{figure*}[!ht]
	\begin{center}
		\begin{tabular}{ccccccc}
			\hspace{-.21cm} \rotatebox{90}{\ \ \footnotesize Target} & \hspace{-.45cm}
			\includegraphics[width=.159\textwidth]{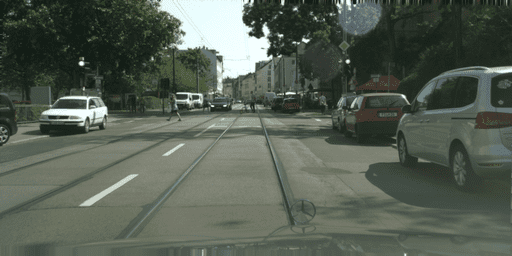} & \hspace{-.45cm}
			\includegraphics[width=.159\textwidth]{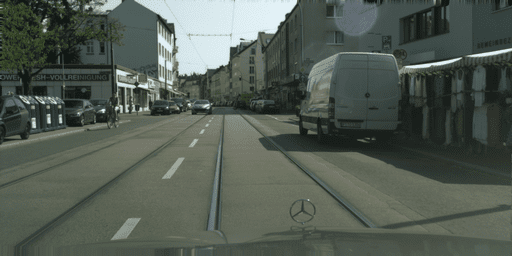} & \hspace{-.45cm}
			\includegraphics[width=.159\textwidth]{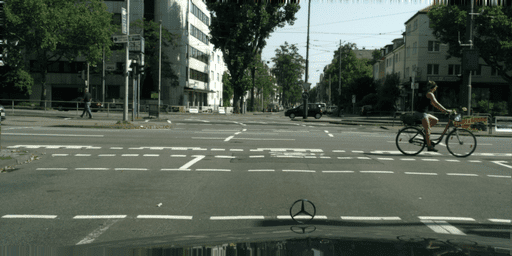} & \hspace{-.45cm}
			\includegraphics[width=.159\textwidth]{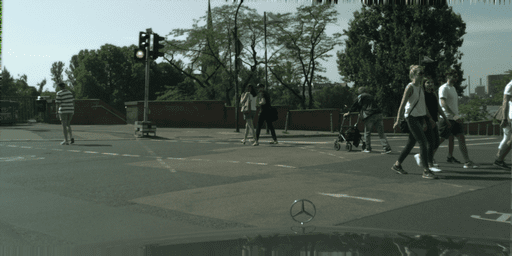} & \hspace{-.45cm}
			\includegraphics[width=.159\textwidth]{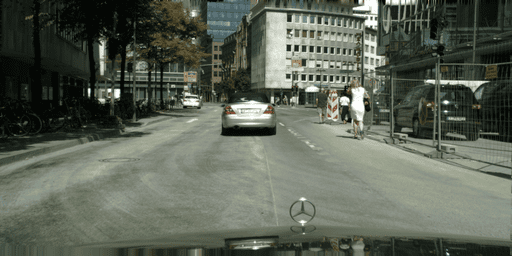} & \hspace{-.45cm}
			\includegraphics[width=.159\textwidth]{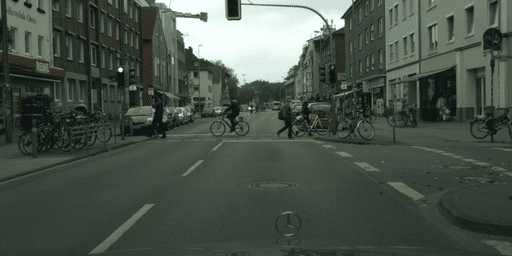} \vspace{-.05cm} \\			
			\hspace{-.21cm} \rotatebox{90}{\ \ \footnotesize ASM} & \hspace{-.45cm}
			\includegraphics[width=.159\textwidth]{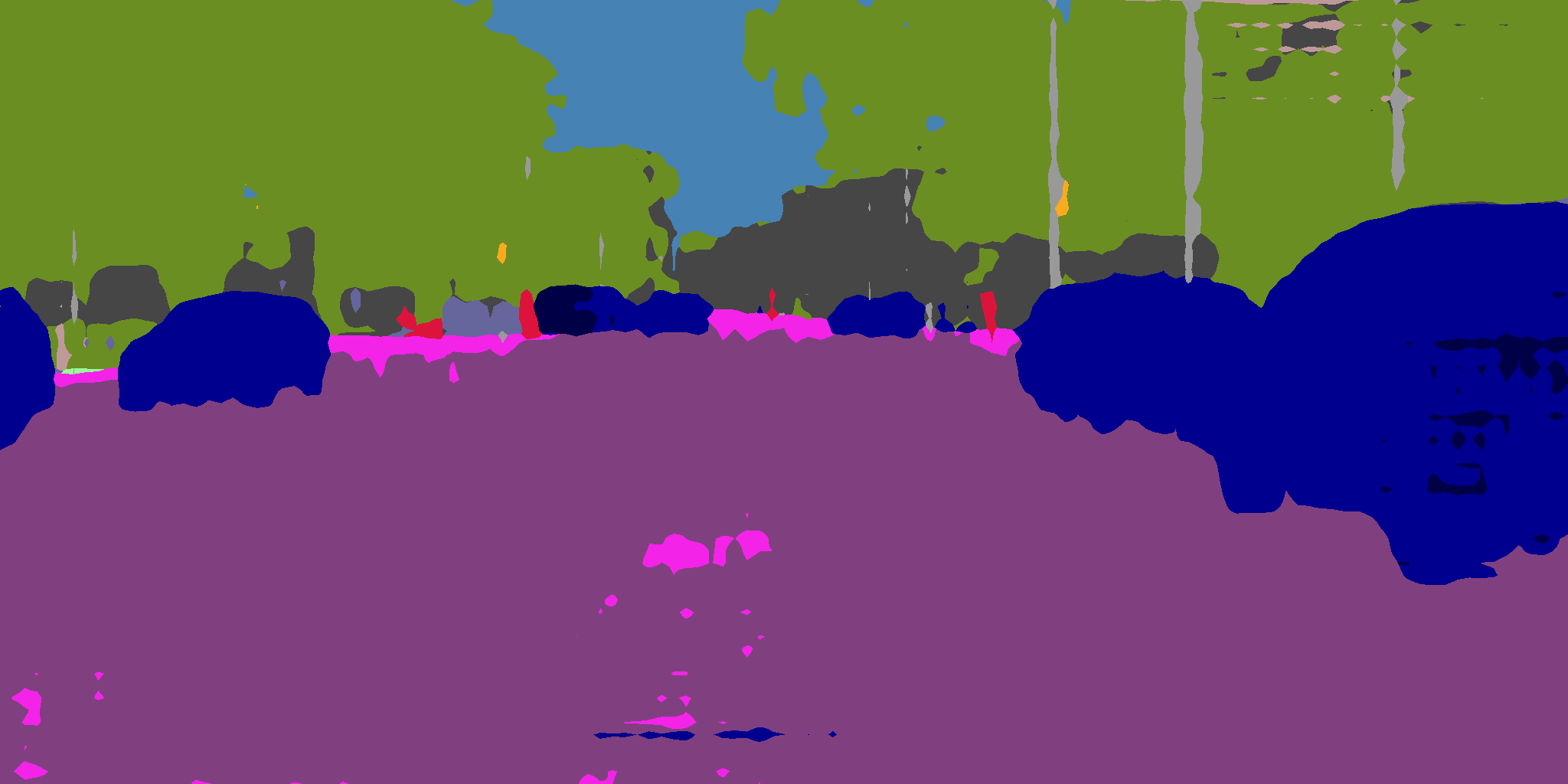} & \hspace{-.45cm}
			\includegraphics[width=.159\textwidth]{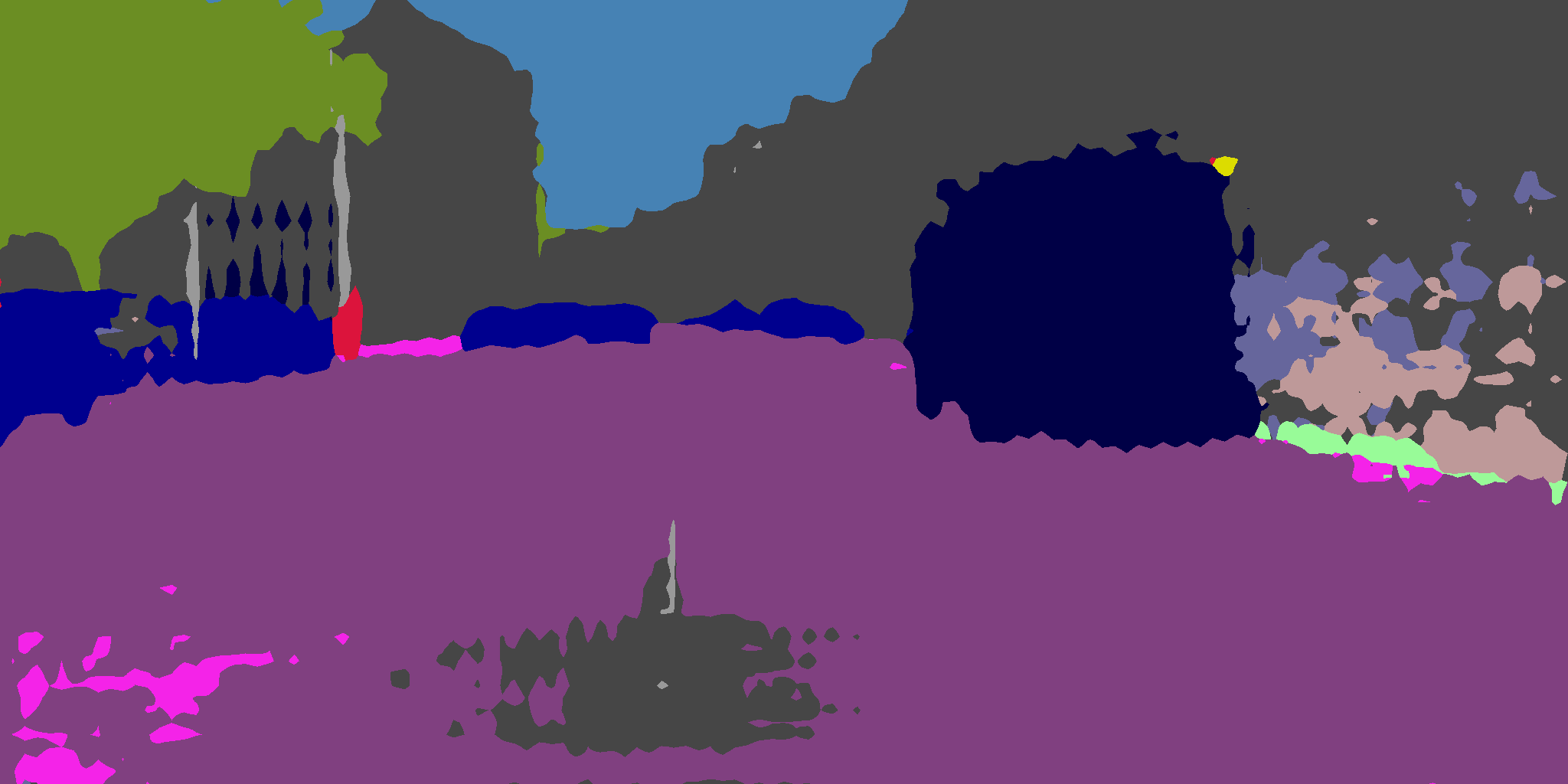} & \hspace{-.45cm}
			\includegraphics[width=.159\textwidth]{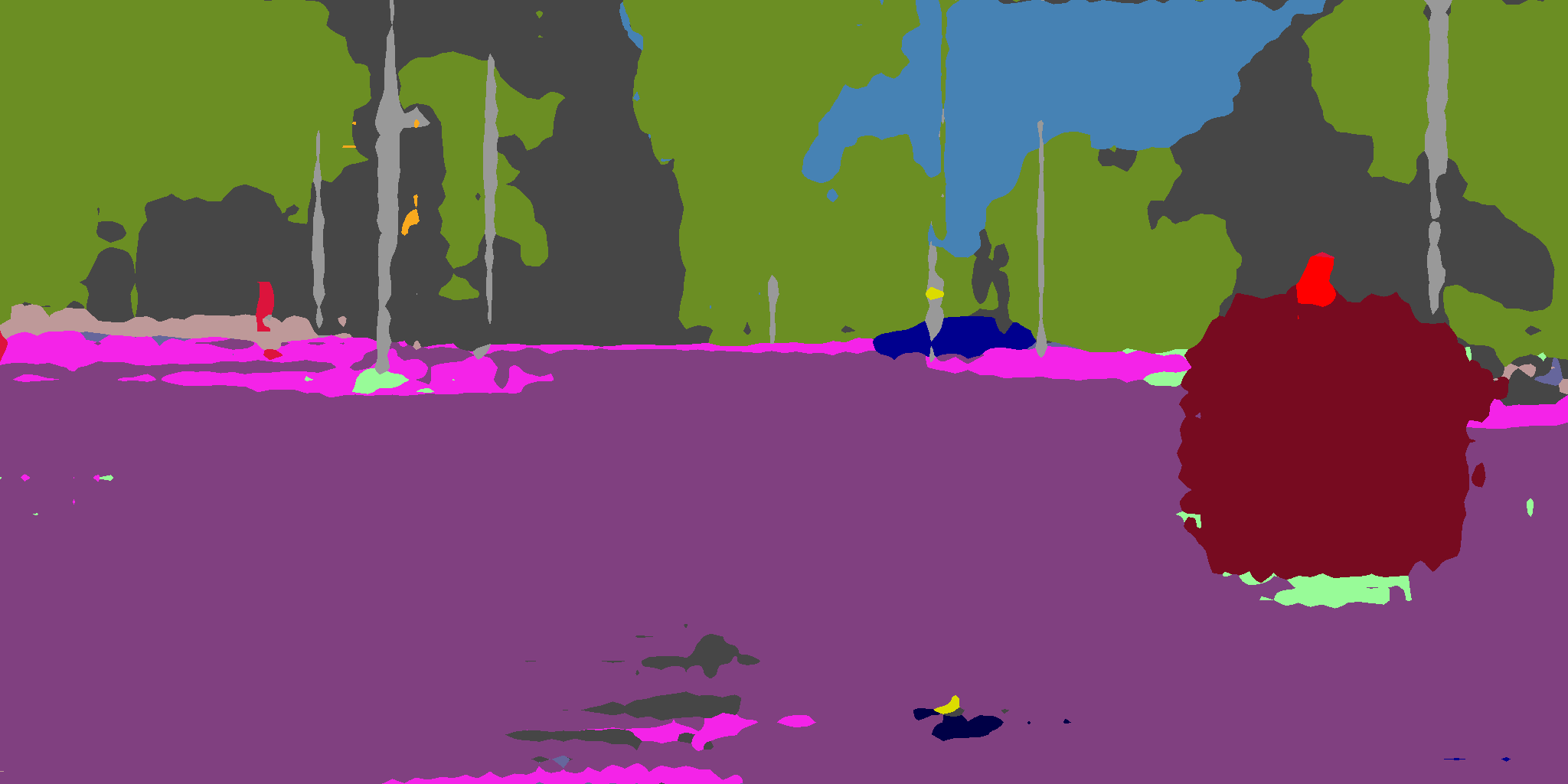} & \hspace{-.45cm}
			\includegraphics[width=.159\textwidth]{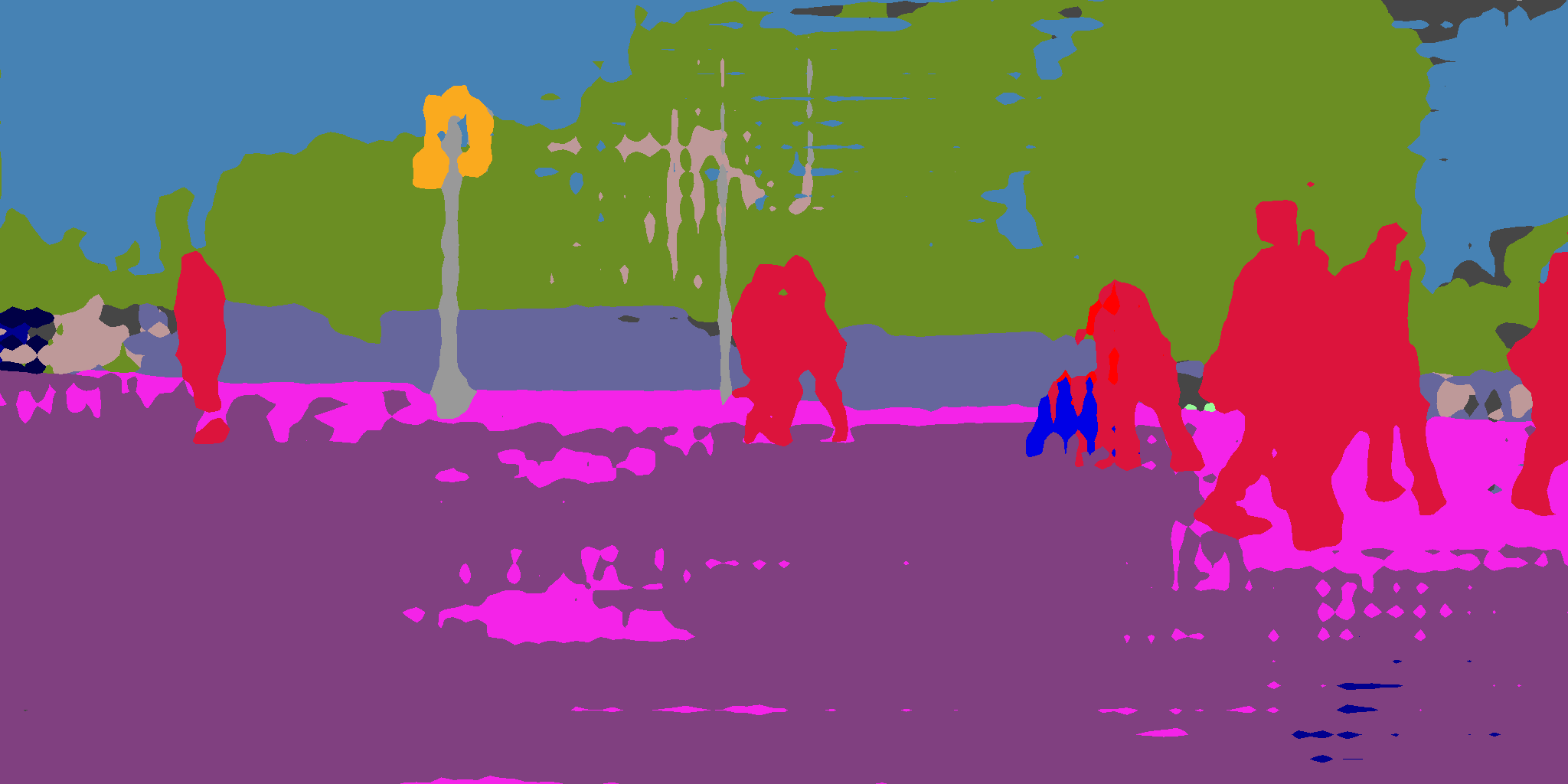} & \hspace{-.45cm}
			\includegraphics[width=.159\textwidth]{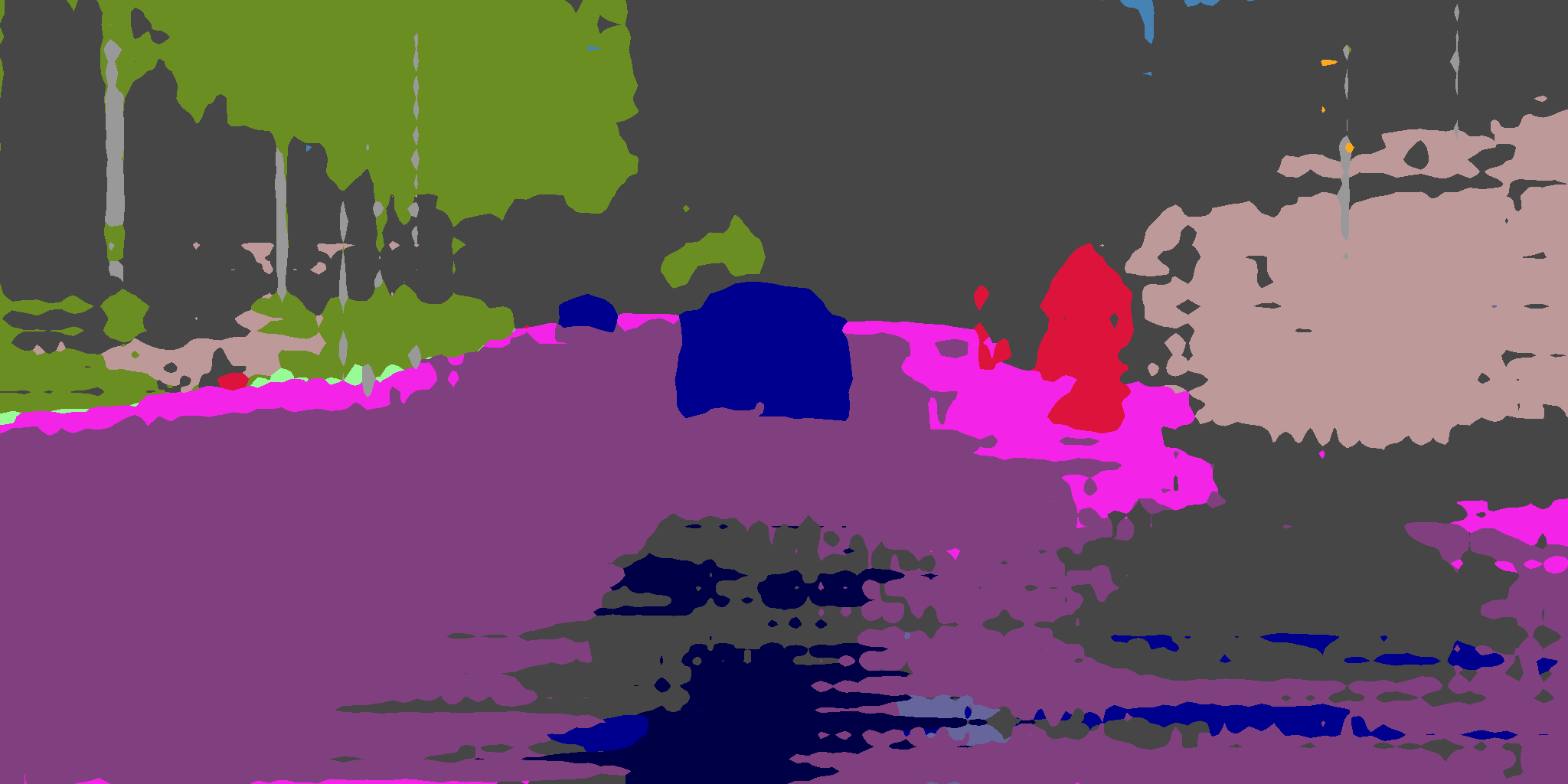} & \hspace{-.45cm}
			\includegraphics[width=.159\textwidth]{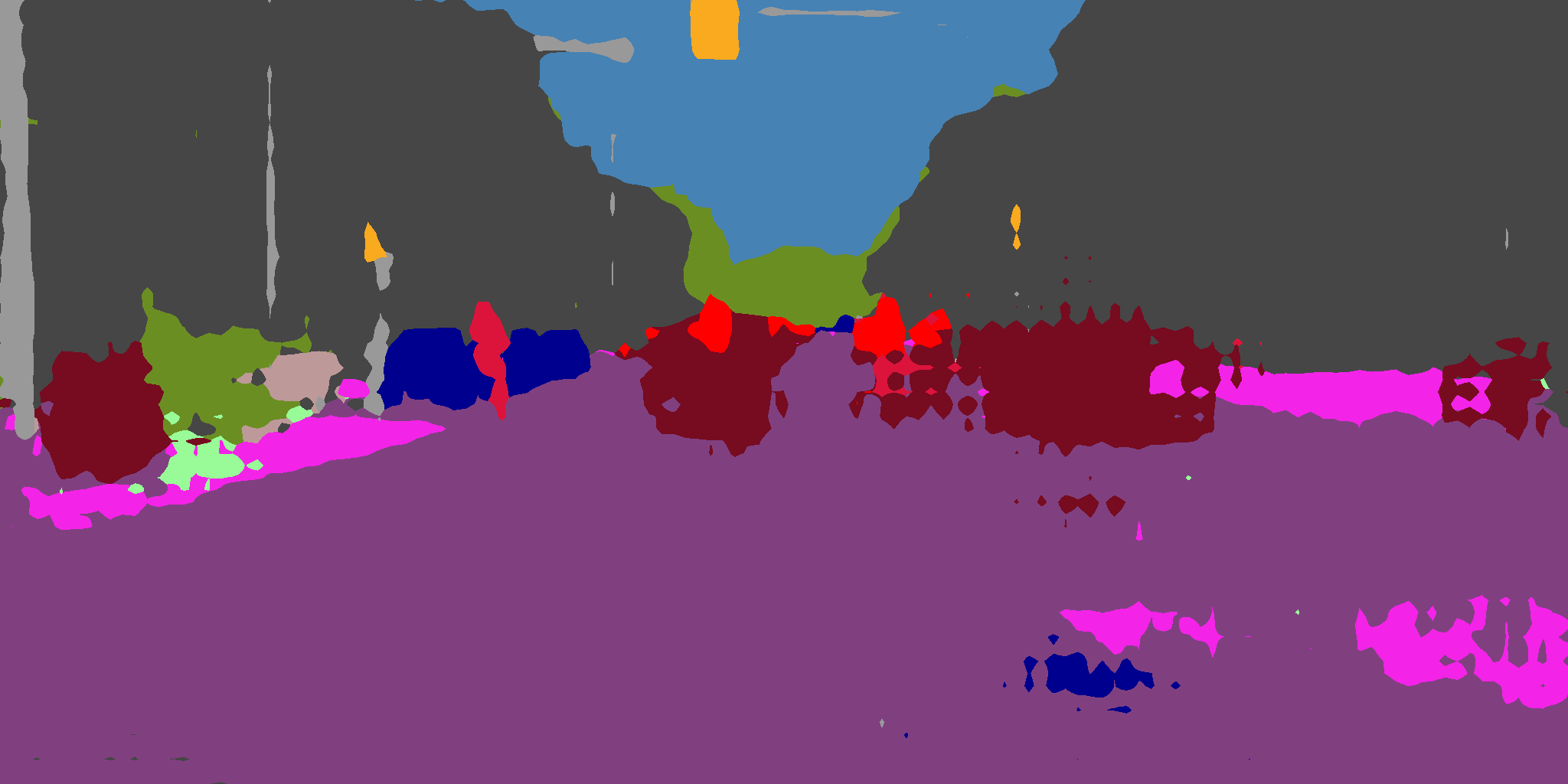} \vspace{-.05cm} \\
			\hspace{-.21cm} \rotatebox{90}{\ \ \footnotesize Ours} & \hspace{-.45cm}
			\includegraphics[width=.159\textwidth]{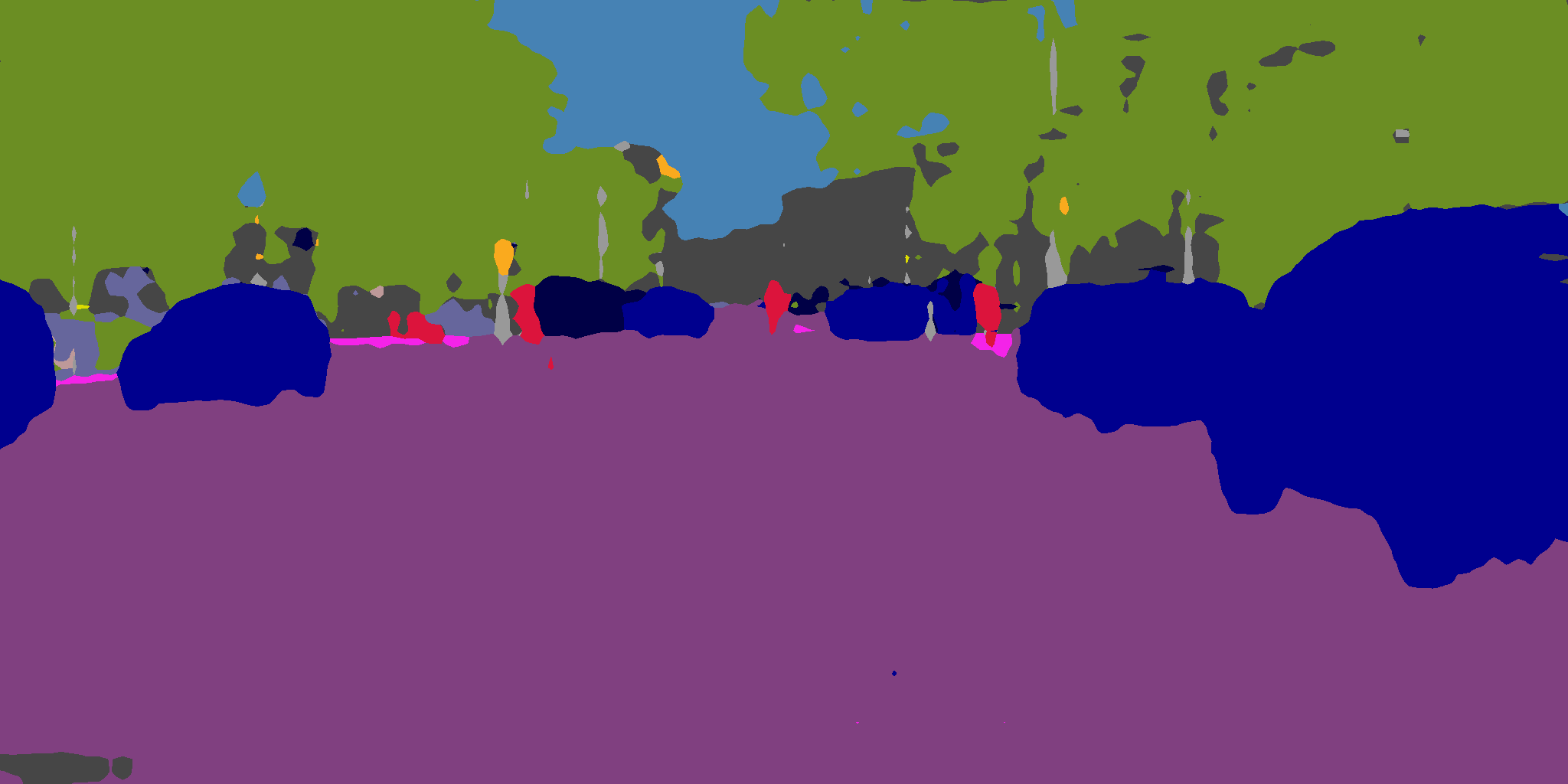} & \hspace{-.45cm}
			\includegraphics[width=.159\textwidth]{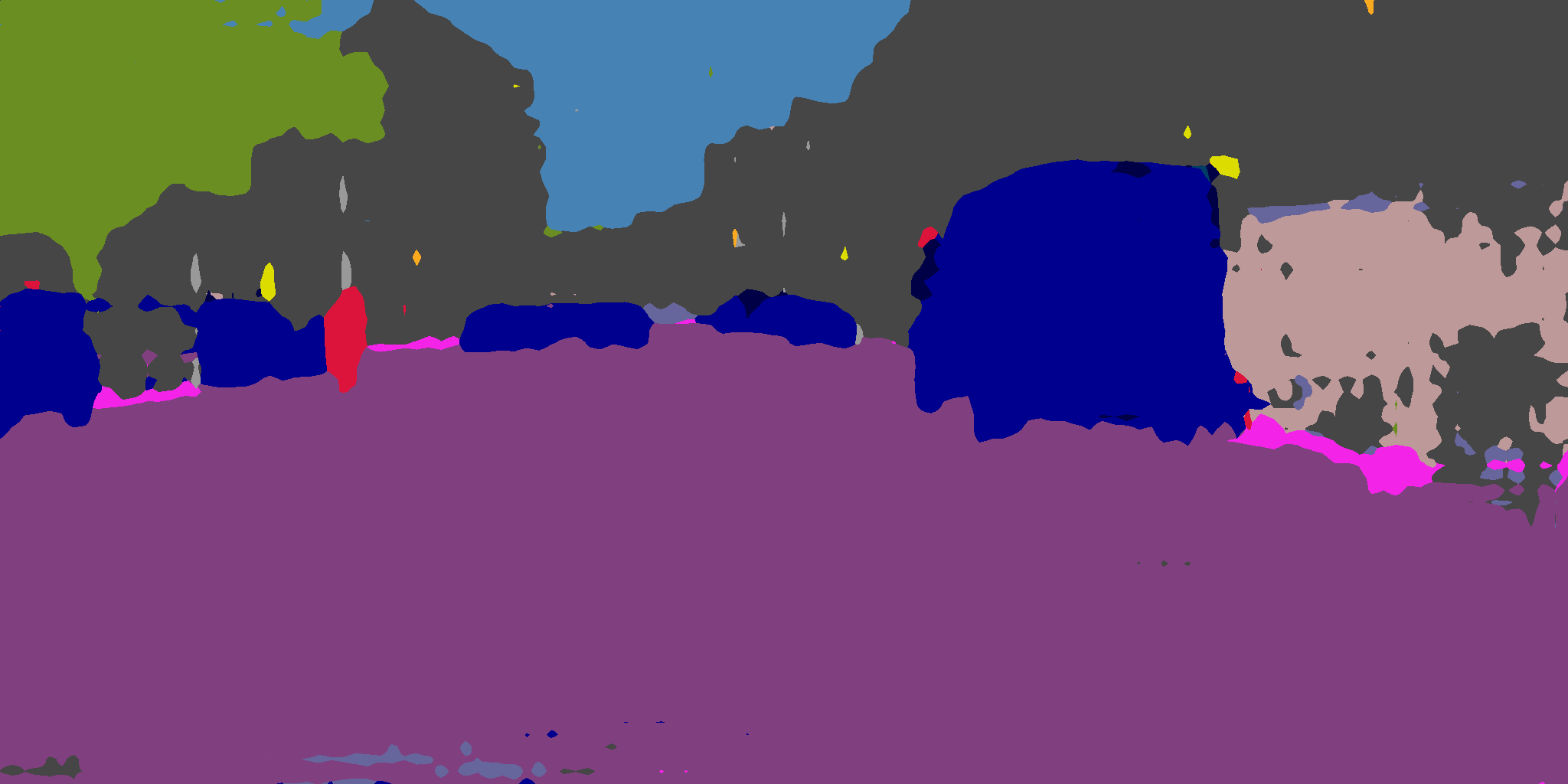} & \hspace{-.45cm}
			\includegraphics[width=.159\textwidth]{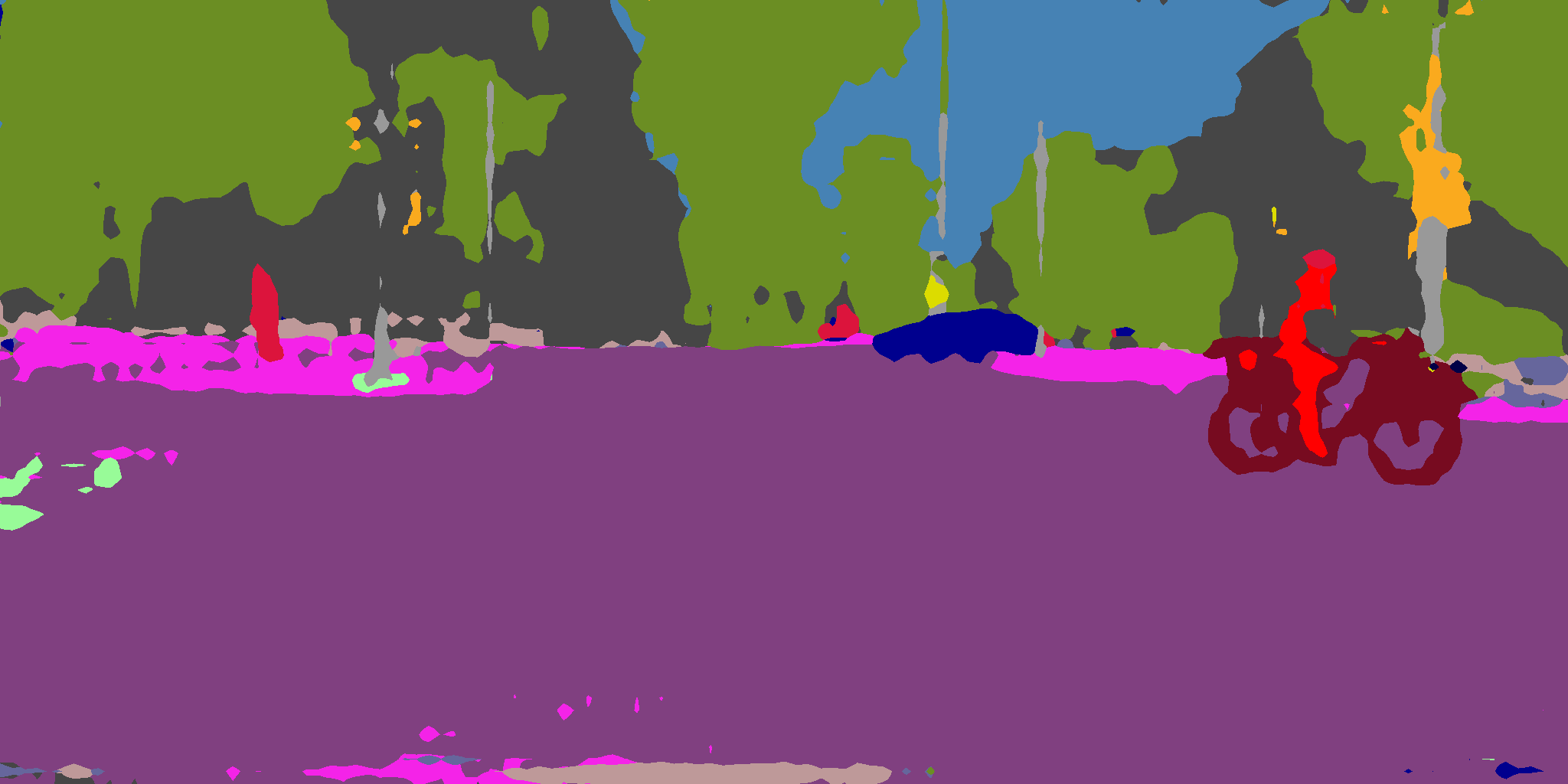} & \hspace{-.45cm}
			\includegraphics[width=.159\textwidth]{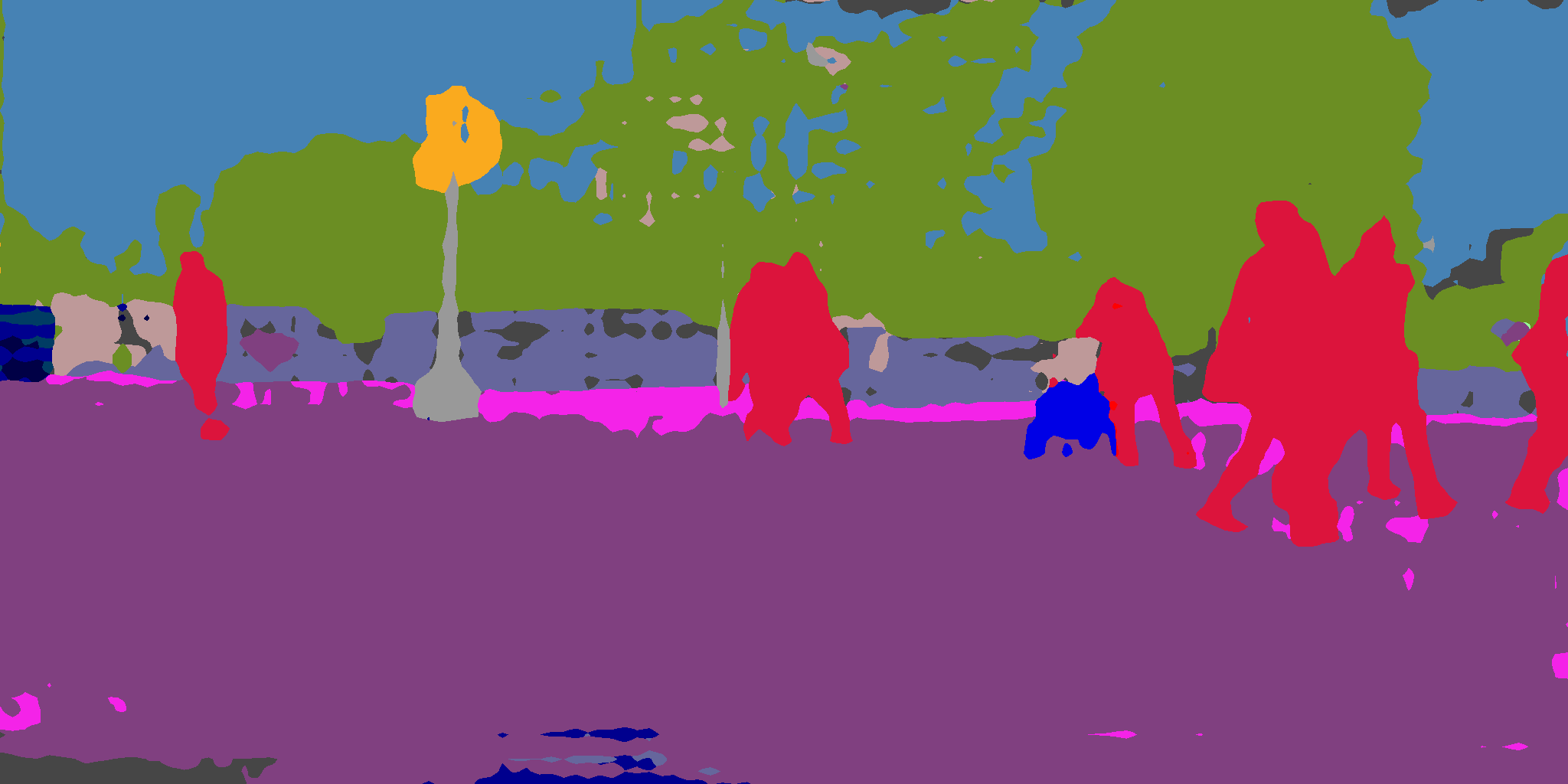} & \hspace{-.45cm}
			\includegraphics[width=.159\textwidth]{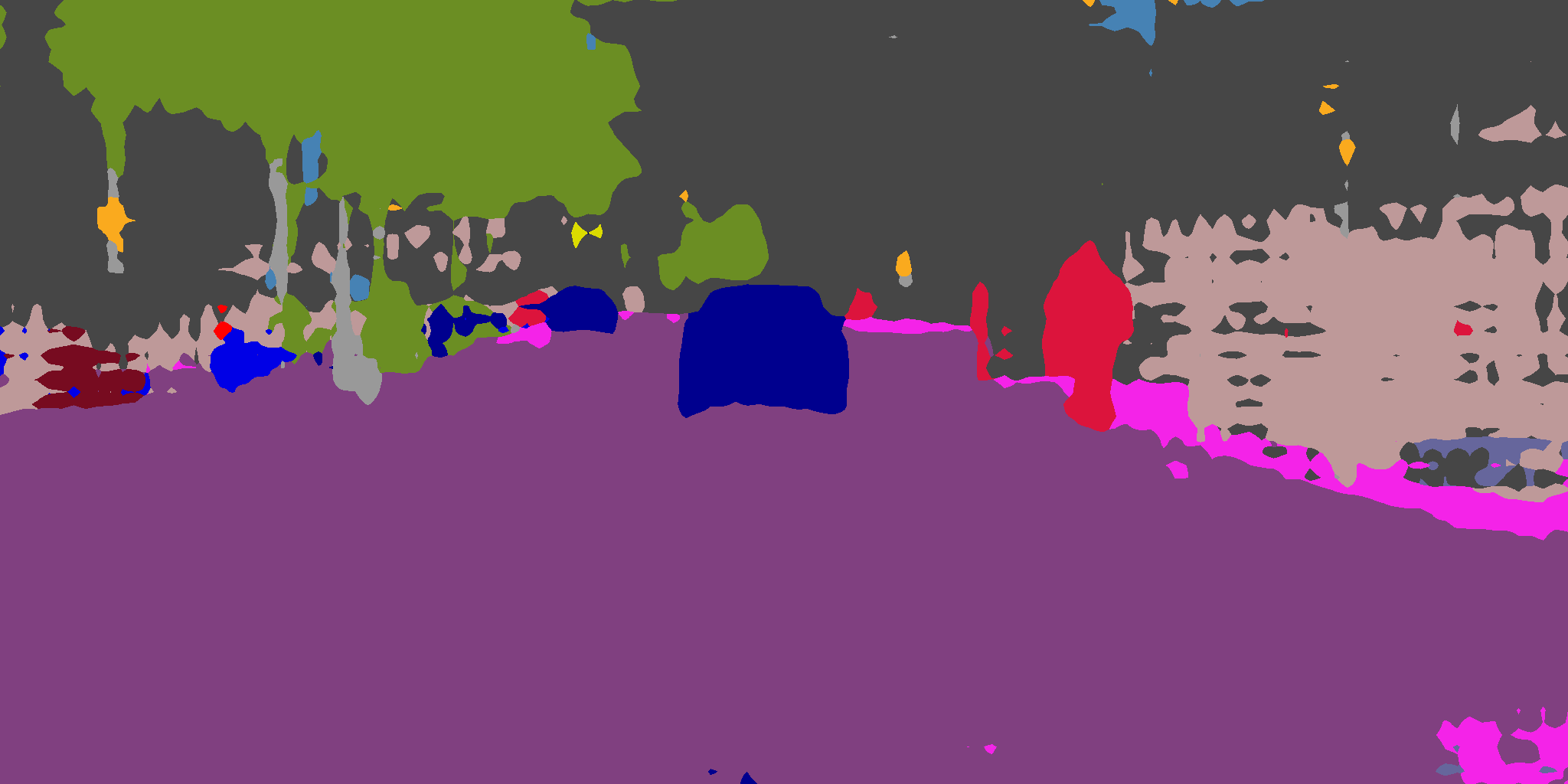} & \hspace{-.45cm}
			\includegraphics[width=.159\textwidth]{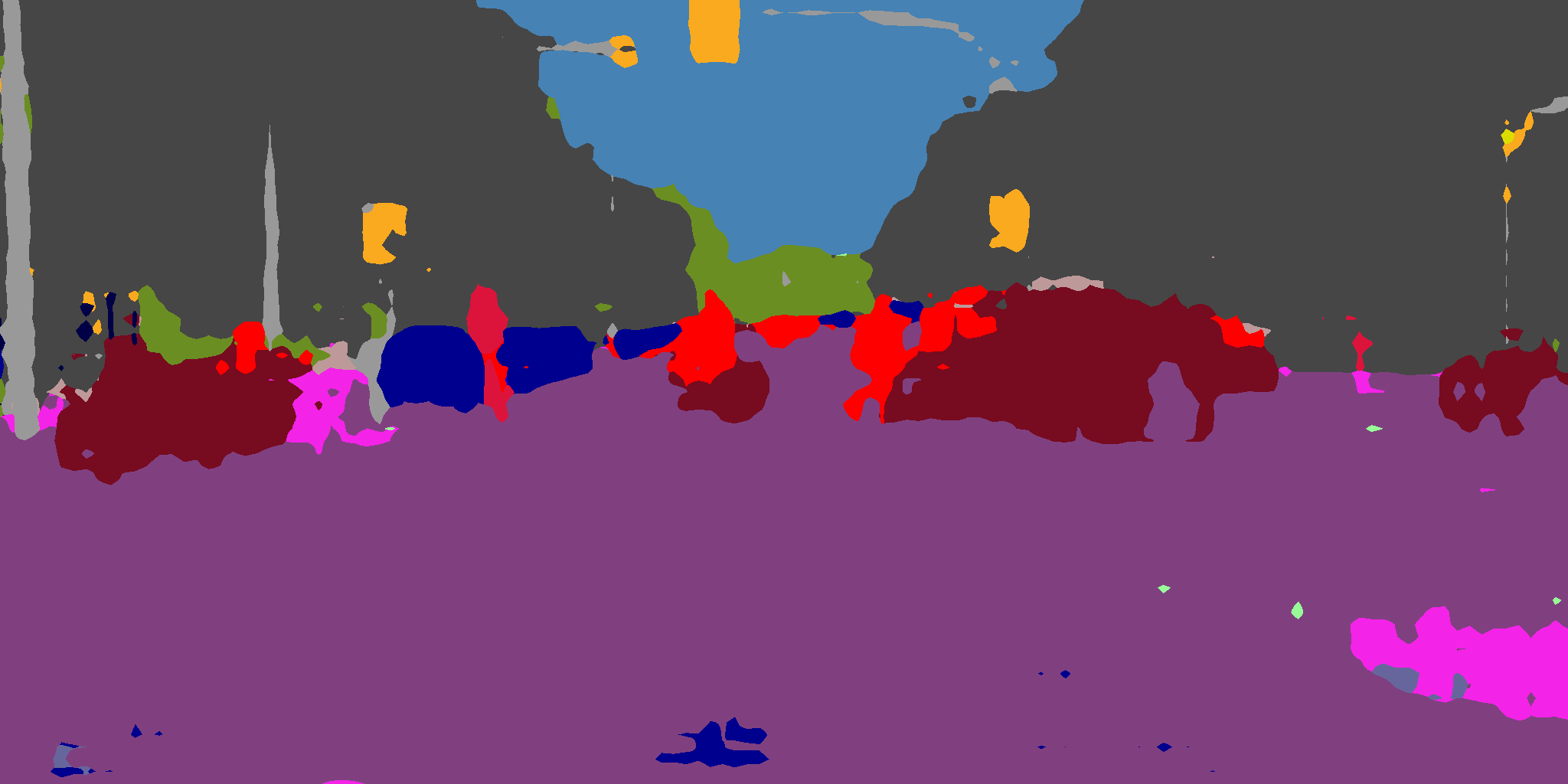} \vspace{-.05cm} \\
			\hspace{-.21cm} \rotatebox{90}{\ \ \footnotesize GT} & \hspace{-.45cm}
			\includegraphics[width=.159\textwidth]{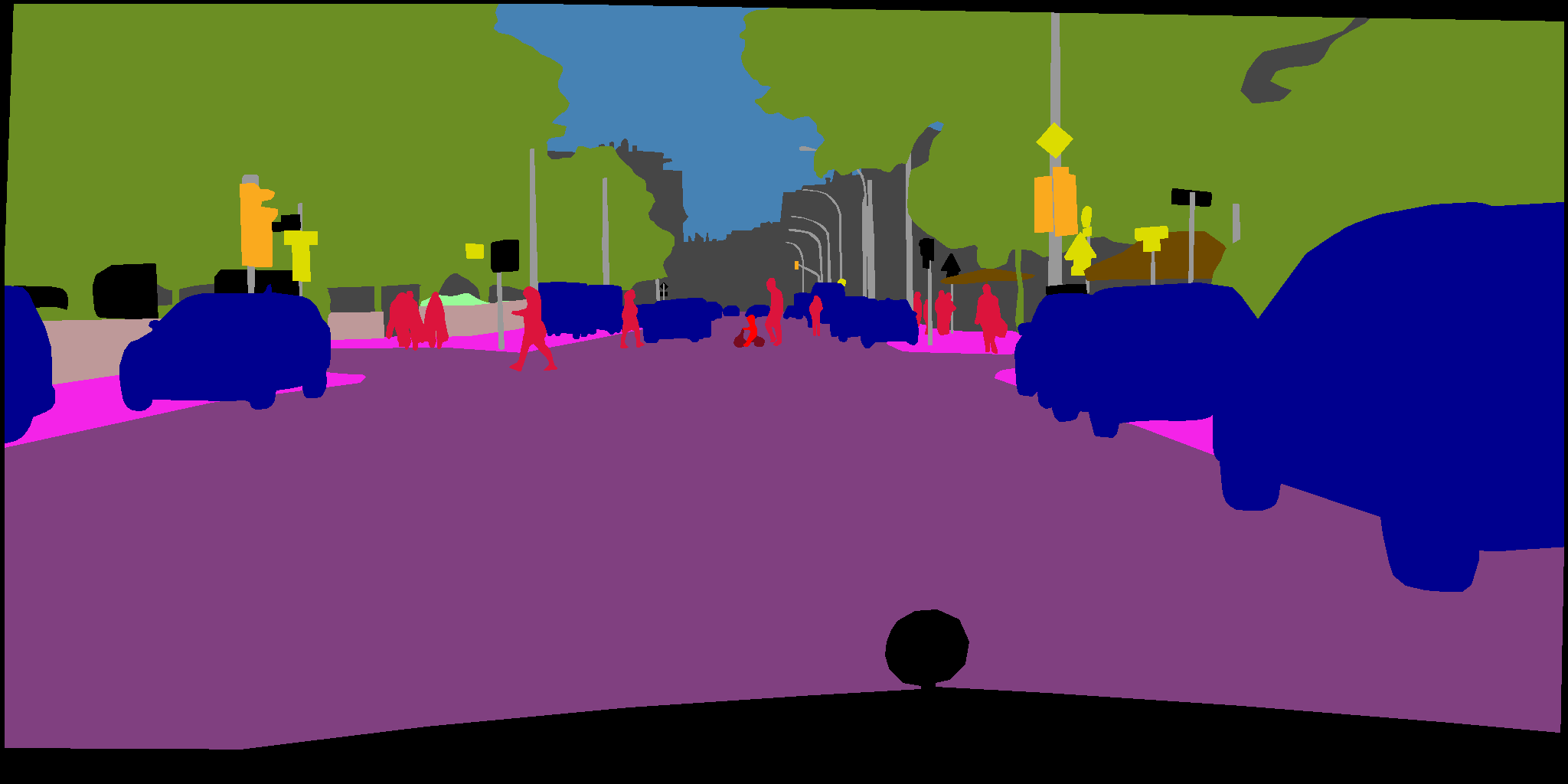} & \hspace{-.45cm}
			\includegraphics[width=.159\textwidth]{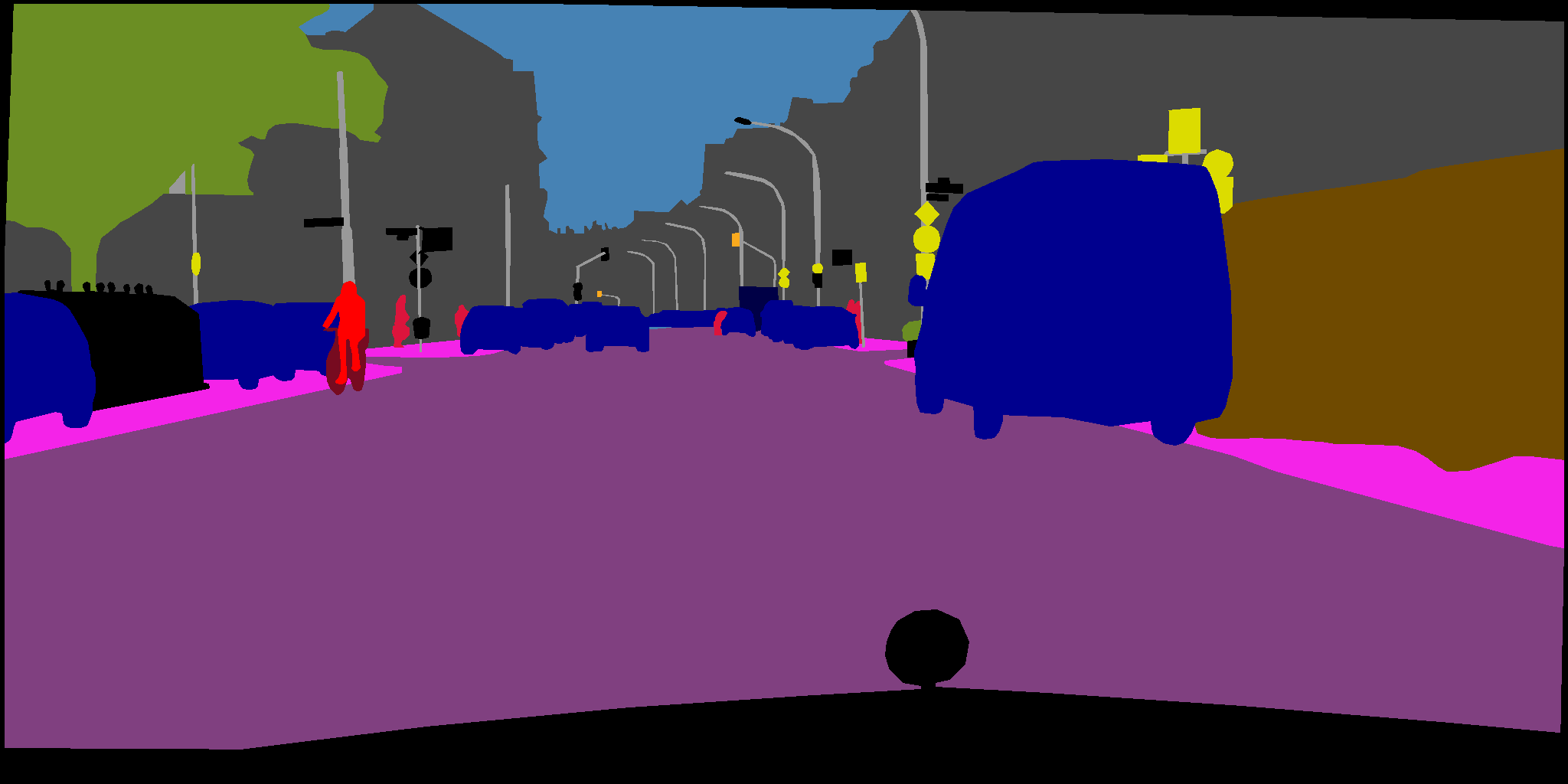} & \hspace{-.45cm}
			\includegraphics[width=.159\textwidth]{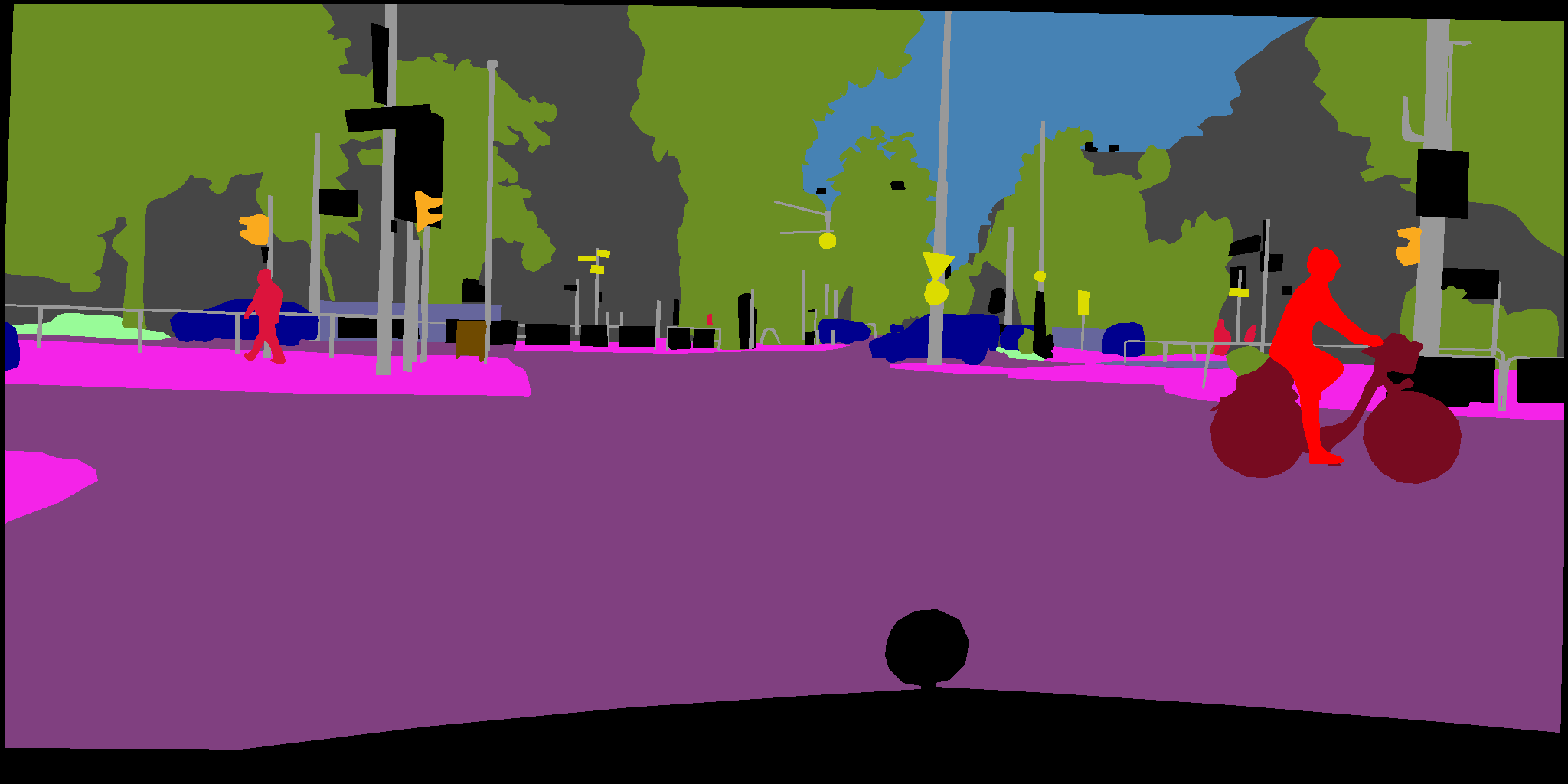} & \hspace{-.45cm}
			\includegraphics[width=.159\textwidth]{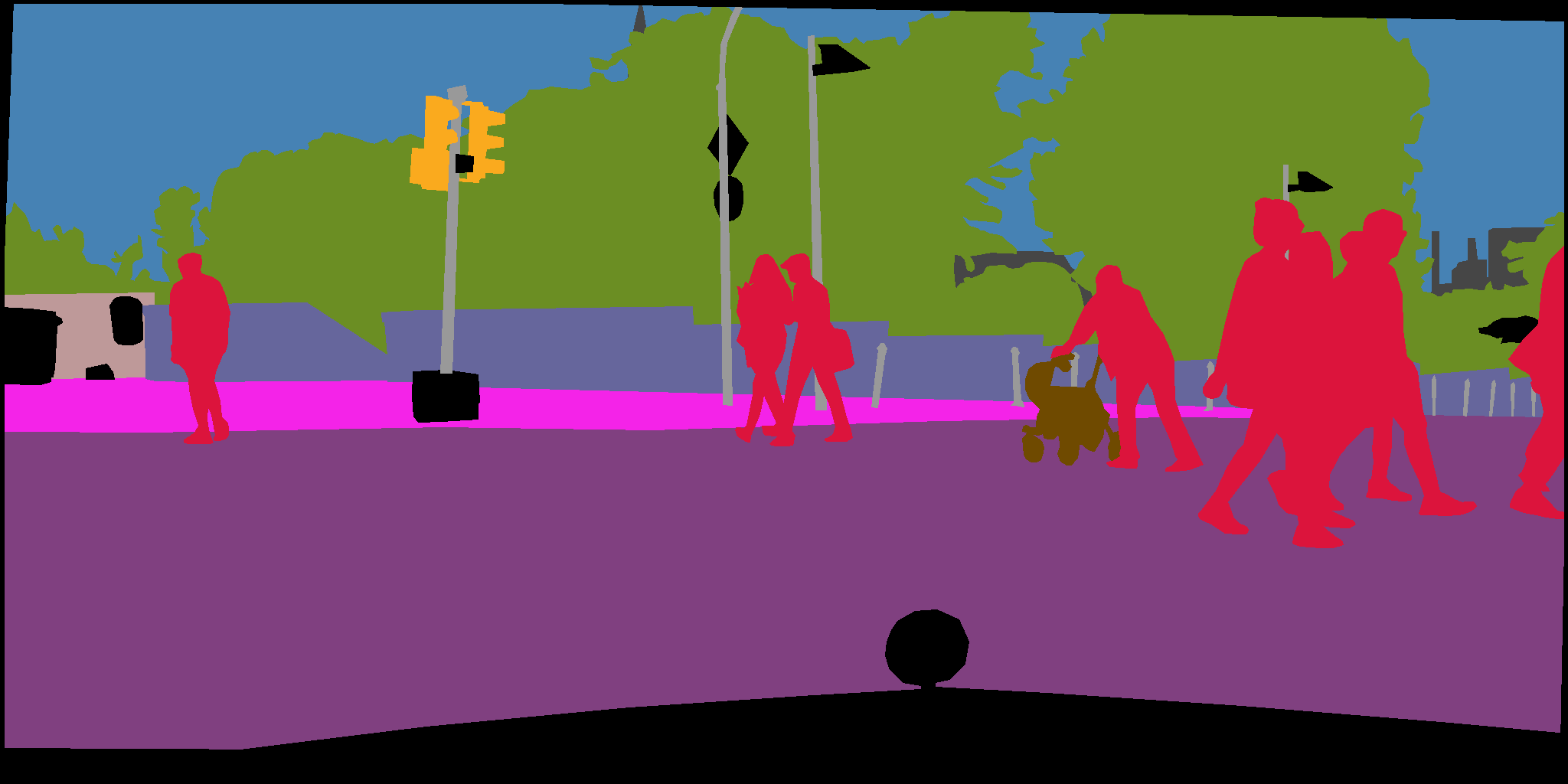} & \hspace{-.45cm}
			\includegraphics[width=.159\textwidth]{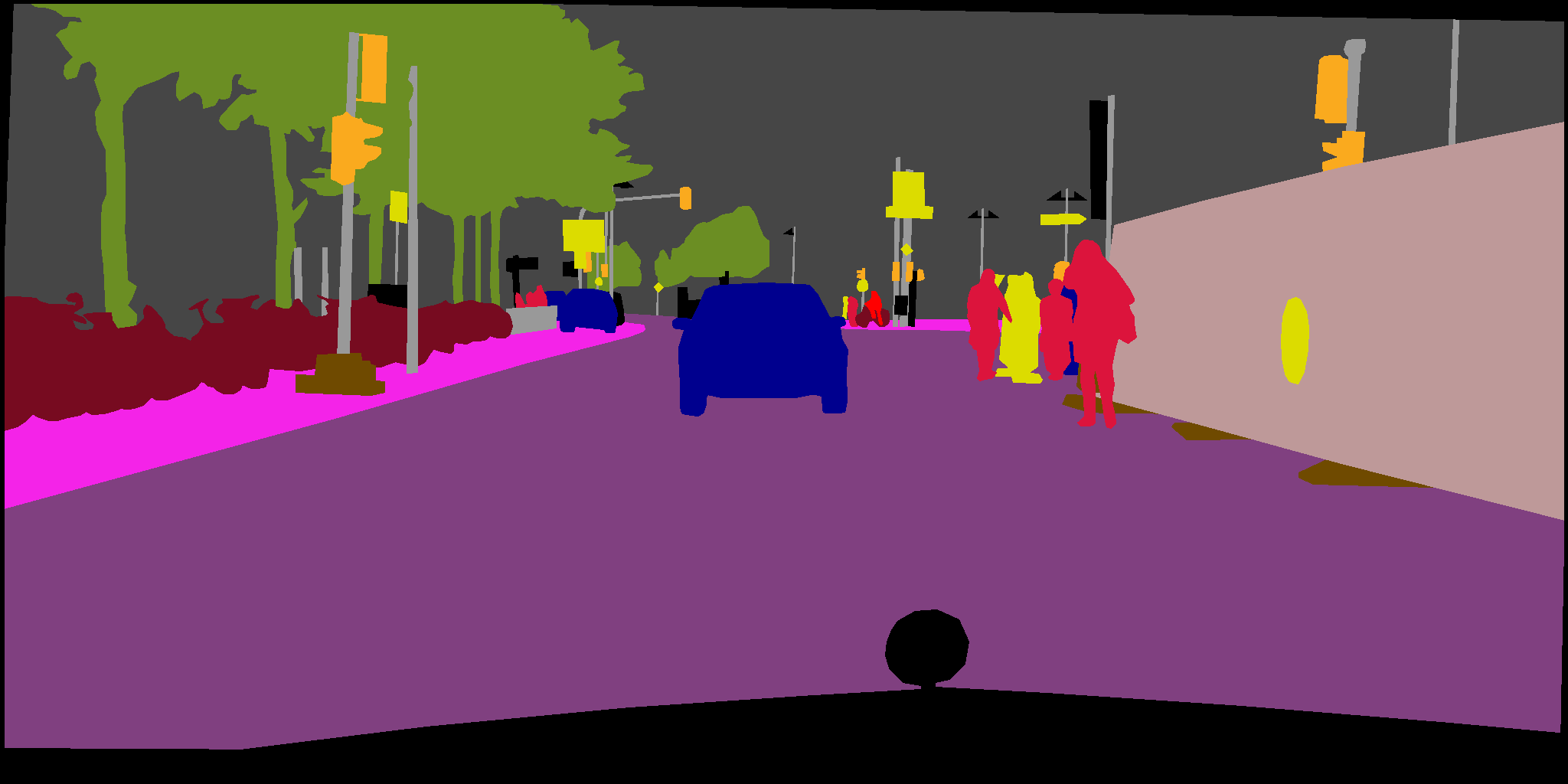} & \hspace{-.45cm}
			\includegraphics[width=.159\textwidth]{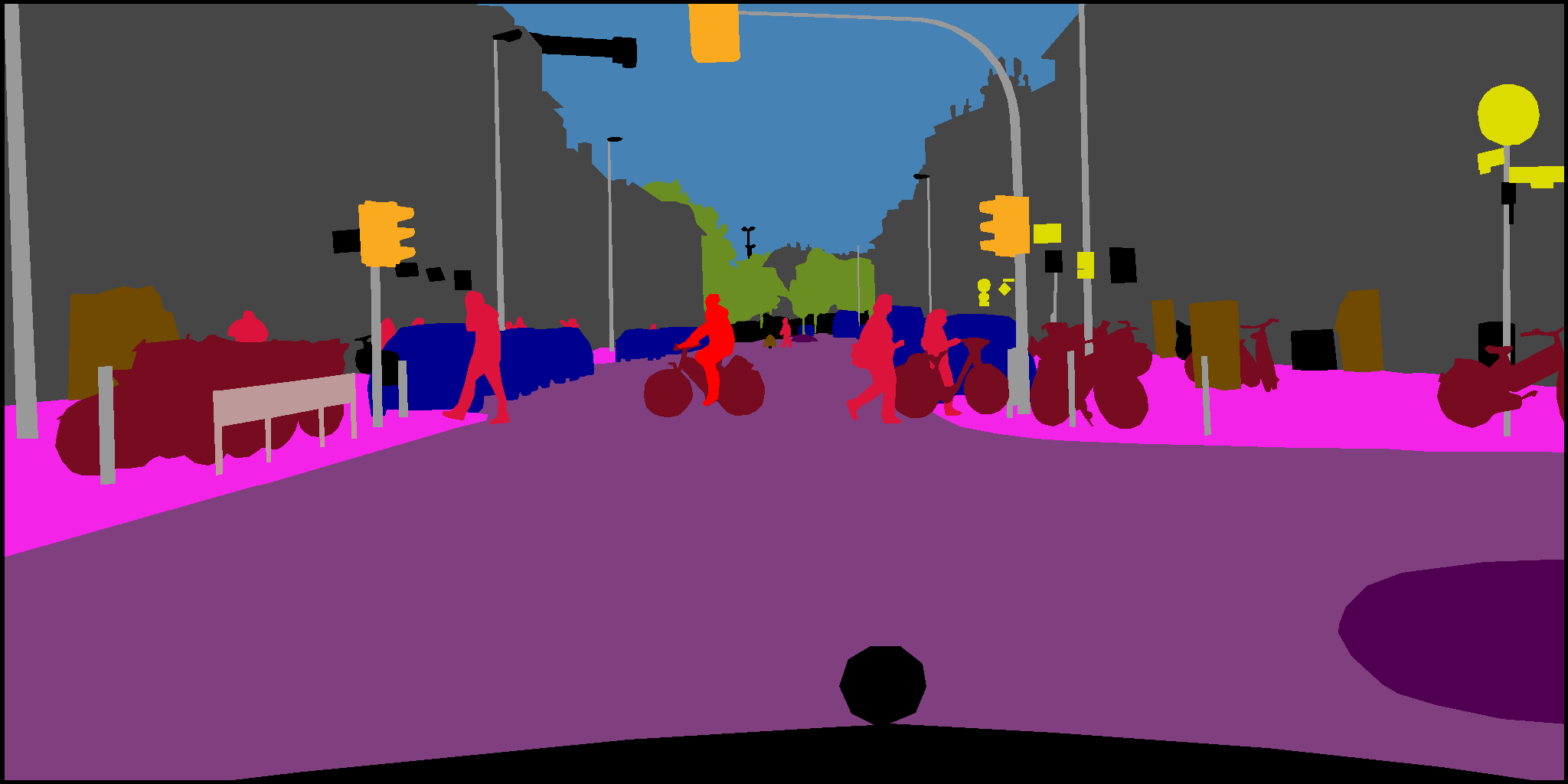} \vspace{-.05cm} \\
			& \hspace{-.45cm}(a) &\hspace{-.45cm}(b)&\hspace{-.45cm}(c)& \hspace{-.45cm}(d)& \hspace{-.45cm}(e) & \hspace{-.45cm}(f)\\
		\end{tabular}
		\caption{Some qualitative comparison results for domain adaptation from GTA5 $\rightarrow$ Cityscapes.}
		\label{Qualitative_gta5}
	\end{center}
\end{figure*}

\subsection{Comparison results}
In Table~\ref{gta5-city}, we present the comparison results with other state-of-the-art approaches for the GTA5 $\rightarrow$ Cityscapes experiment. The compared method can be divided into three camps based on the data samples except for the source domain that are needed for adaptation: 1) one unlabeled target image only (denoted by O); 2) style image dataset (denoted by S); (3) both 1) and 2) (denoted by O+S). It can be observed that our method achieves the best performance in the first camp. Obviously, the general UDA approaches Adaptseg \cite{Tsai_adaptseg_2018}, CLAN \cite{luo2019taking} and CBST~\cite{zou2018unsupervised} are not working well in the one-shot setting and some even get worse results than the source only. Methods CycleGAN ~\cite{CycleGAN2017} and OST \cite{Benaim2018OneShotUC} are proved to be more robust to this setting which indicates the usefulness of the style transfer strategy. ASM~\cite{Luo2020ASM} is the first method that tackles the OSUDA which is the most related one to ours. To make a fair comparison, we reproduce the results of ASM using the same backbone\footnote{https://github.com/RoyalVane/ASM/issues/2} as us and the reported mIoU is also based on the model saved in the last iteration (not selecting the best one). Especially, our method does not need an additional dataset to pre-train a style transfer model while ASM needs, and ours runs only for 500 iterations for domain adaptation to achieve these comparable results. 
We also find that domain generalization approaches DRPC~\cite{yue2019domain} and FSDR \cite{huang2021fsdr} using additional style image datasets also achieve comparable results or even better than the methods using one target image. This indicates that using more images than only one target image can be more helpful as expected. However, they need to spend more time exploring the desired domains and the style references also need to be properly chosen. 
The results for SYNTHIA $\rightarrow$ Cityscapes experiment are reported in Table.~\ref{syn2city}, where our method achieves the best performance across all of the three settings and surpasses the second-best by 4.5\% mIoU in the one-shot only setting.

\begin{figure*}[!ht]
	\begin{center}
		\begin{tabular}{ccccccc}
			\hspace{-.21cm} \rotatebox{90}{\ \ \footnotesize Target} & \hspace{-.45cm}
			\includegraphics[width=.159\textwidth]{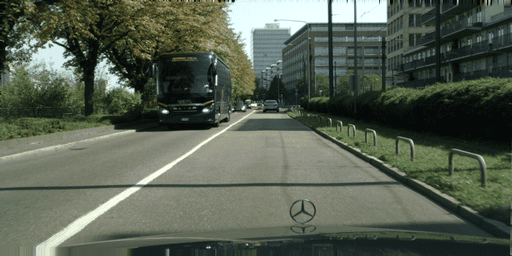} & \hspace{-.45cm}
			\includegraphics[width=.159\textwidth]{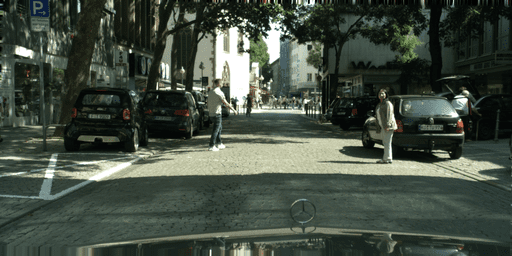} & \hspace{-.45cm}
			\includegraphics[width=.159\textwidth]{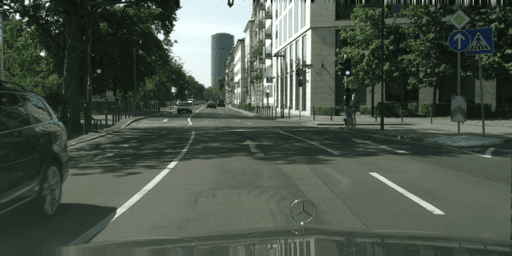} & \hspace{-.45cm}
			\includegraphics[width=.159\textwidth]{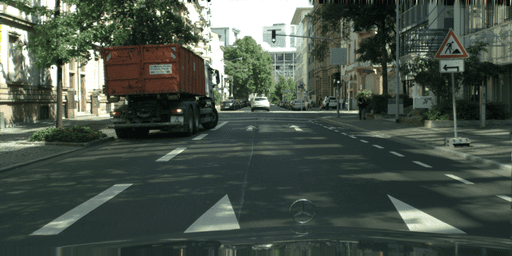} & \hspace{-.45cm}
			\includegraphics[width=.159\textwidth]{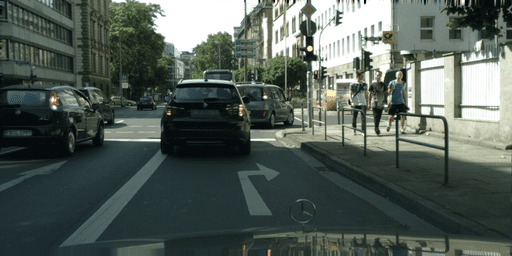} & \hspace{-.45cm}
			\includegraphics[width=.159\textwidth]{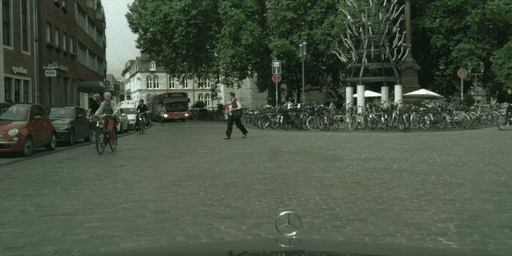} \vspace{-.05cm} \\			
			\hspace{-.21cm} \rotatebox{90}{\ \ \footnotesize ASM} & \hspace{-.45cm}
			\includegraphics[width=.159\textwidth]{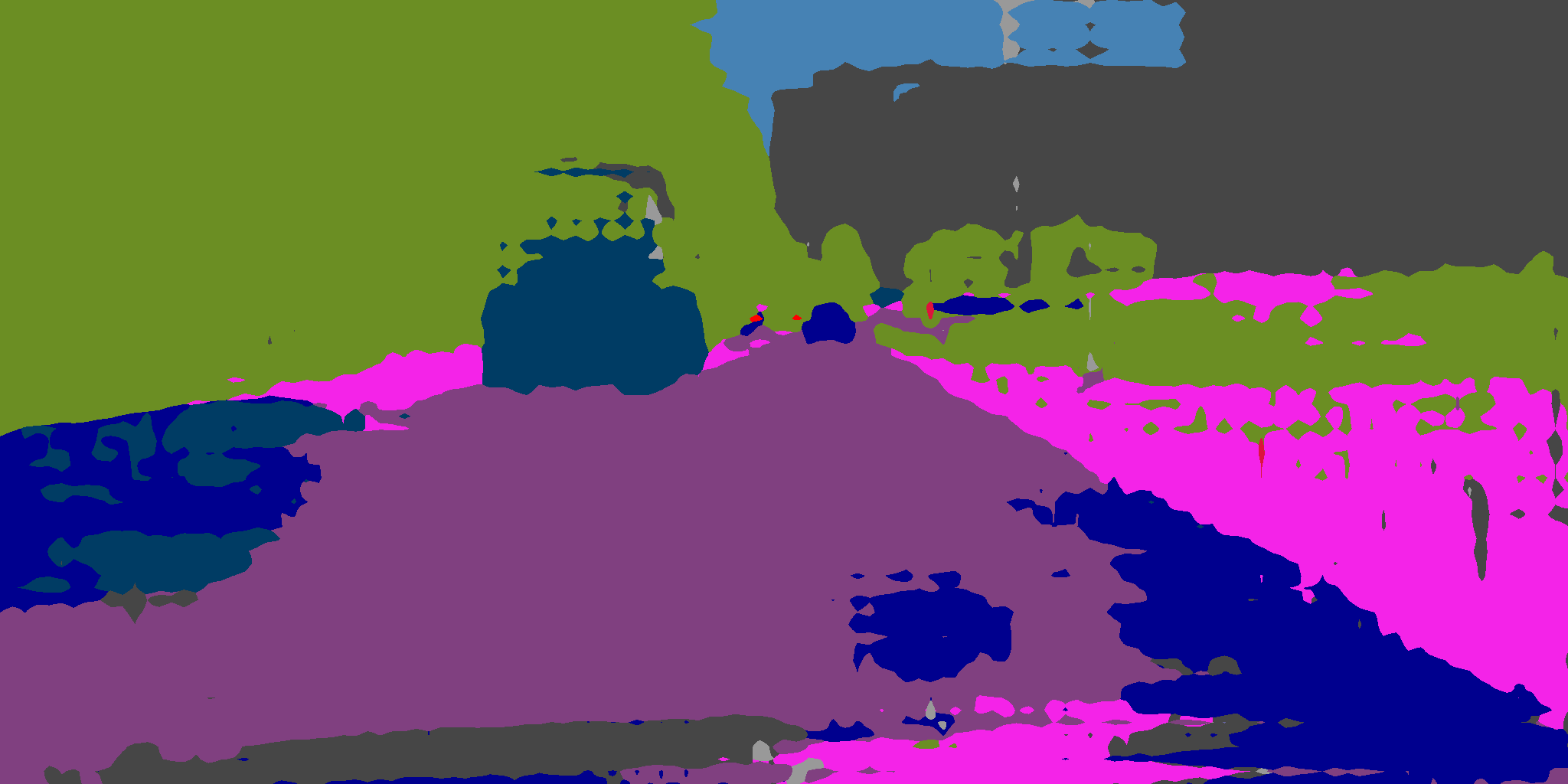} & \hspace{-.45cm}
			\includegraphics[width=.159\textwidth]{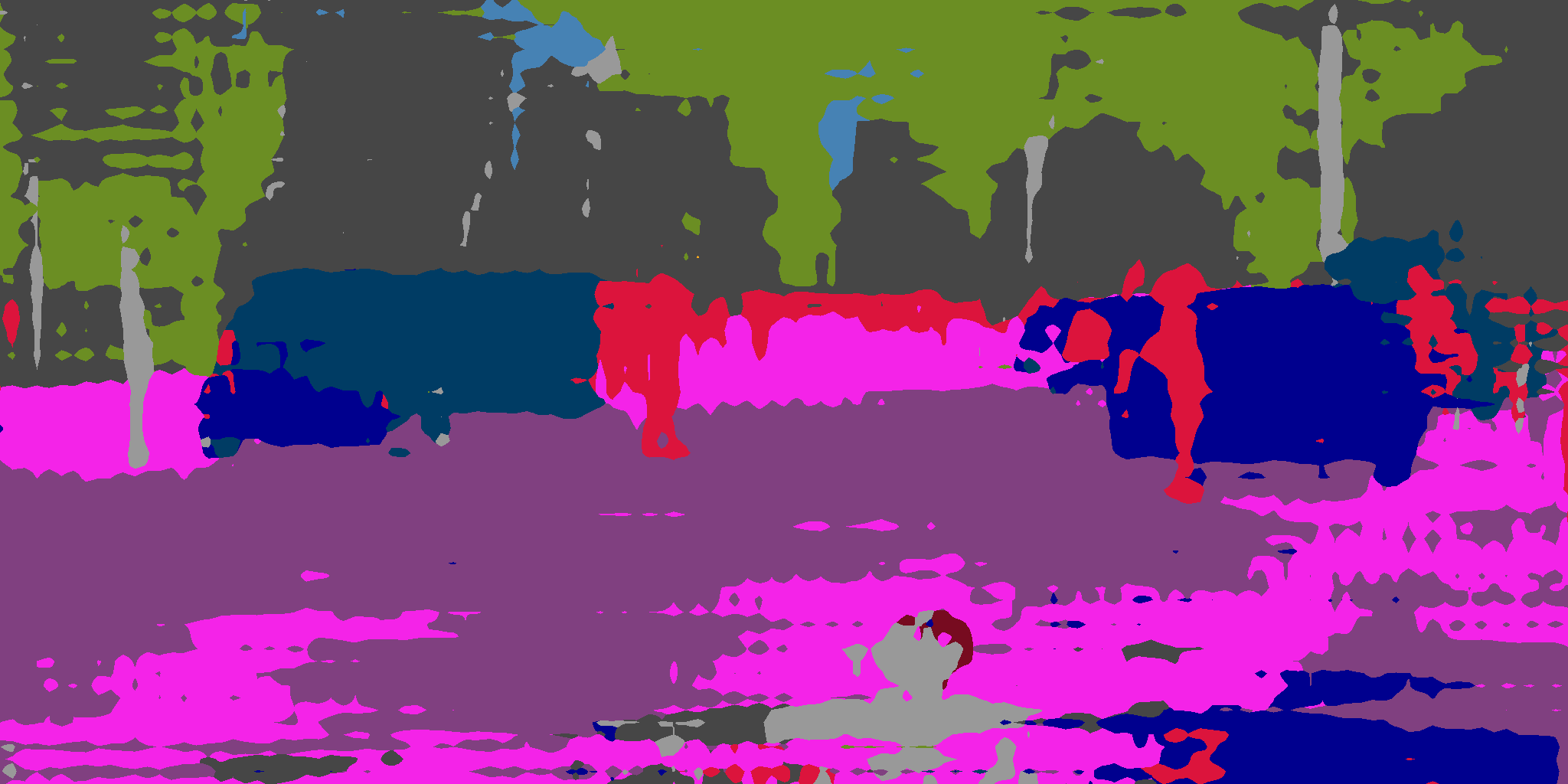} & \hspace{-.45cm}
			\includegraphics[width=.159\textwidth]{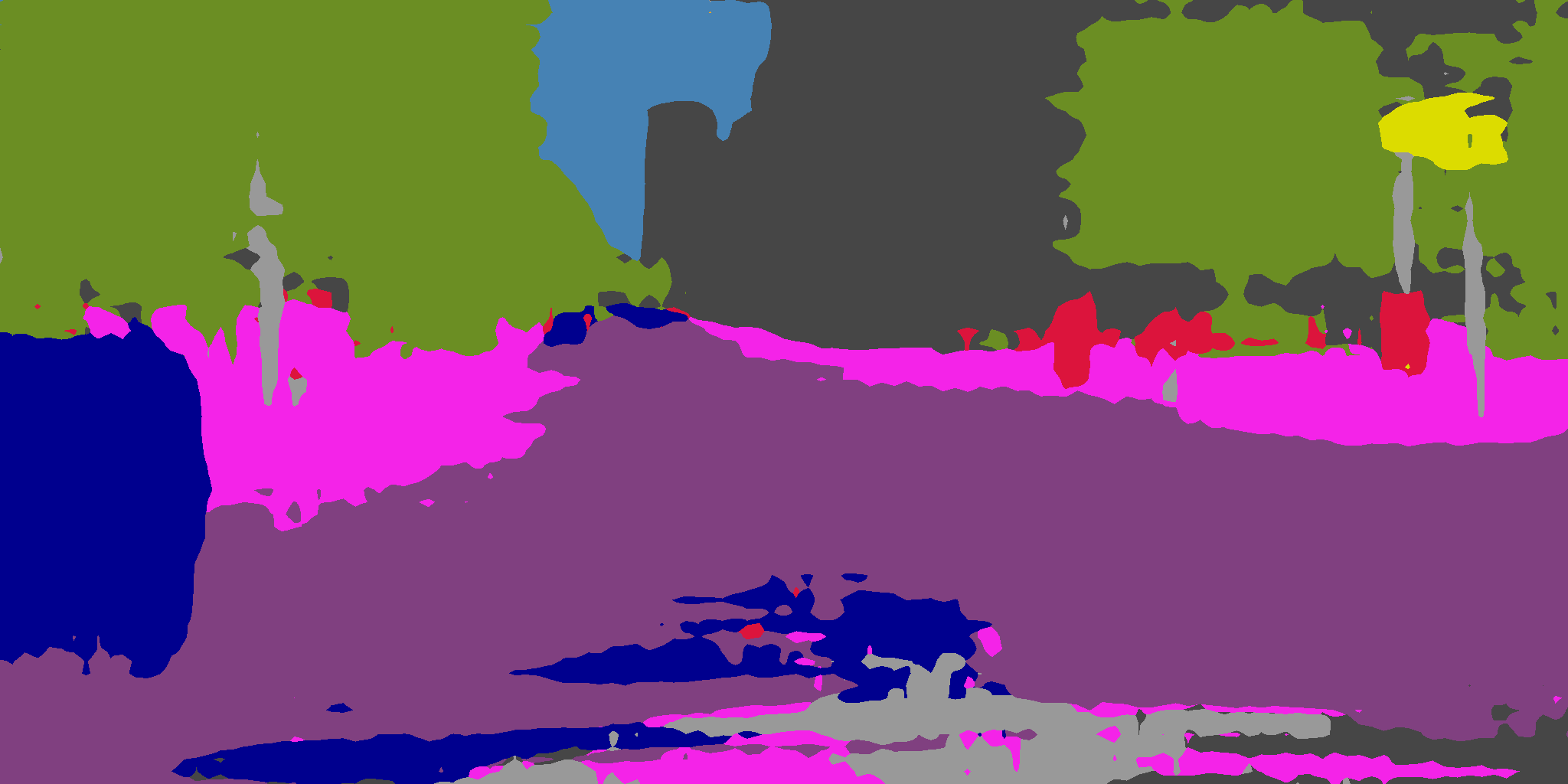} & \hspace{-.45cm}
			\includegraphics[width=.159\textwidth]{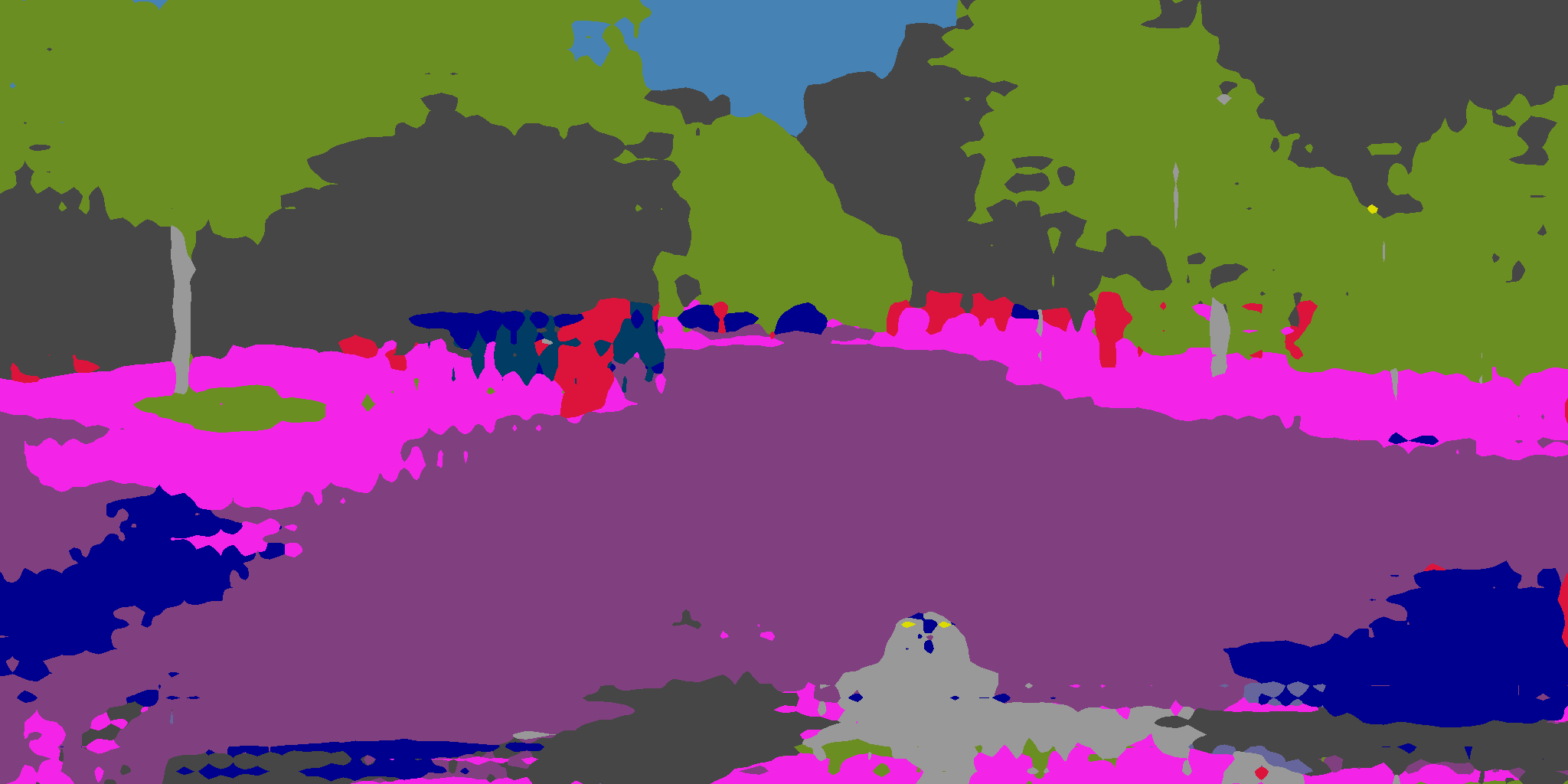} & \hspace{-.45cm}
			\includegraphics[width=.159\textwidth]{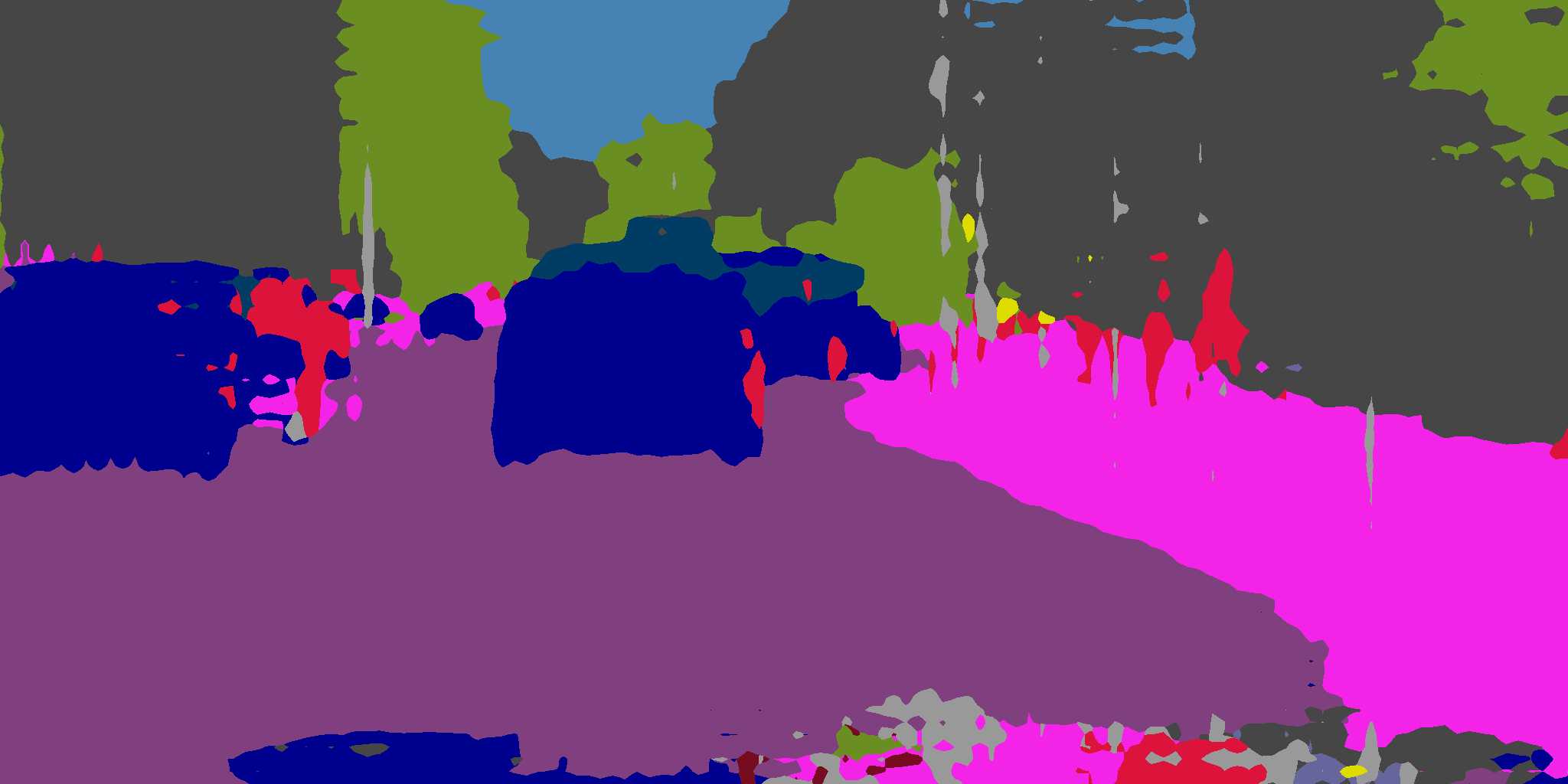} & \hspace{-.45cm}
			\includegraphics[width=.159\textwidth]{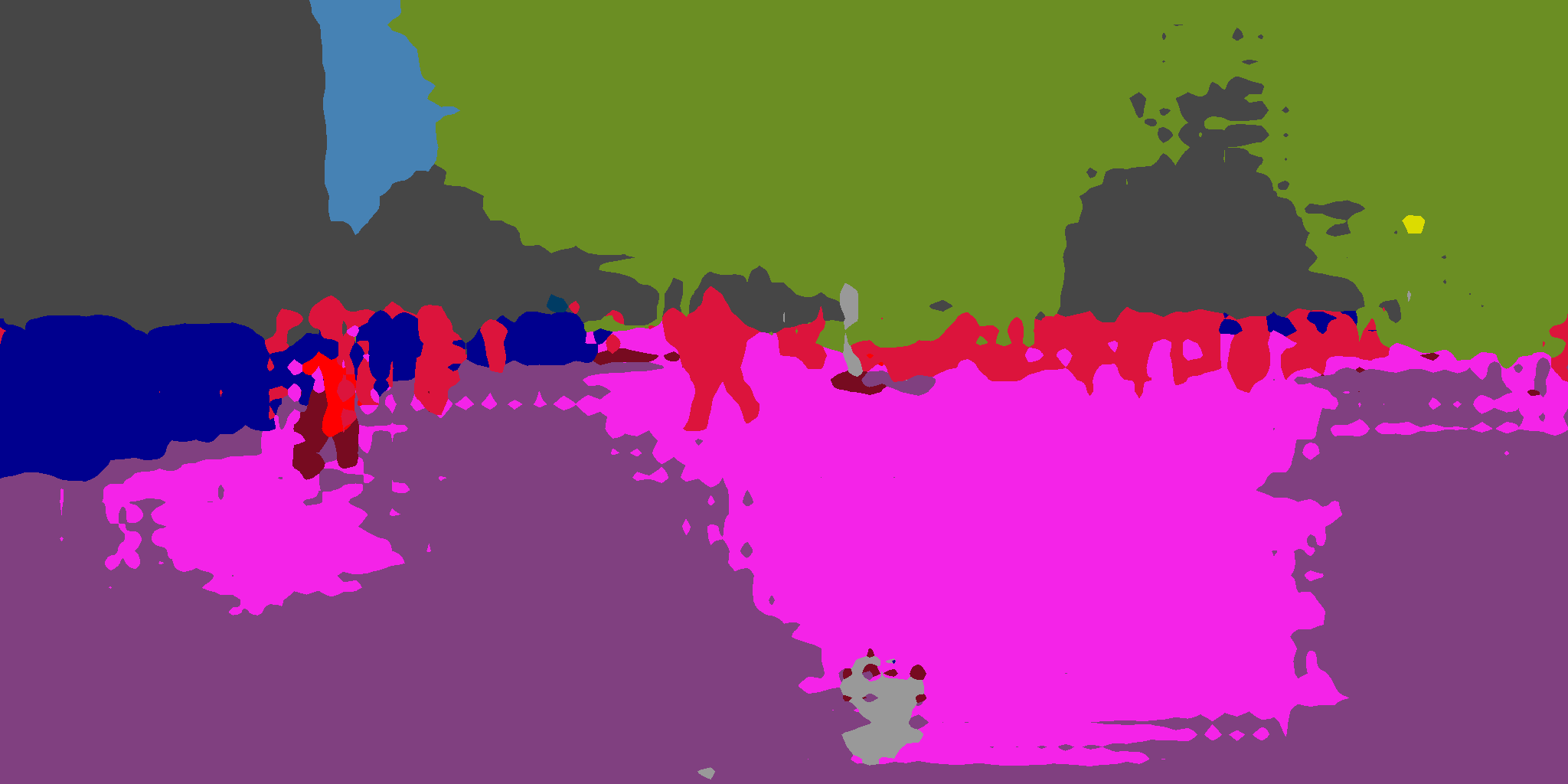} \vspace{-.05cm} \\
			\hspace{-.21cm} \rotatebox{90}{\ \ \footnotesize Ours} & \hspace{-.45cm}
			\includegraphics[width=.159\textwidth]{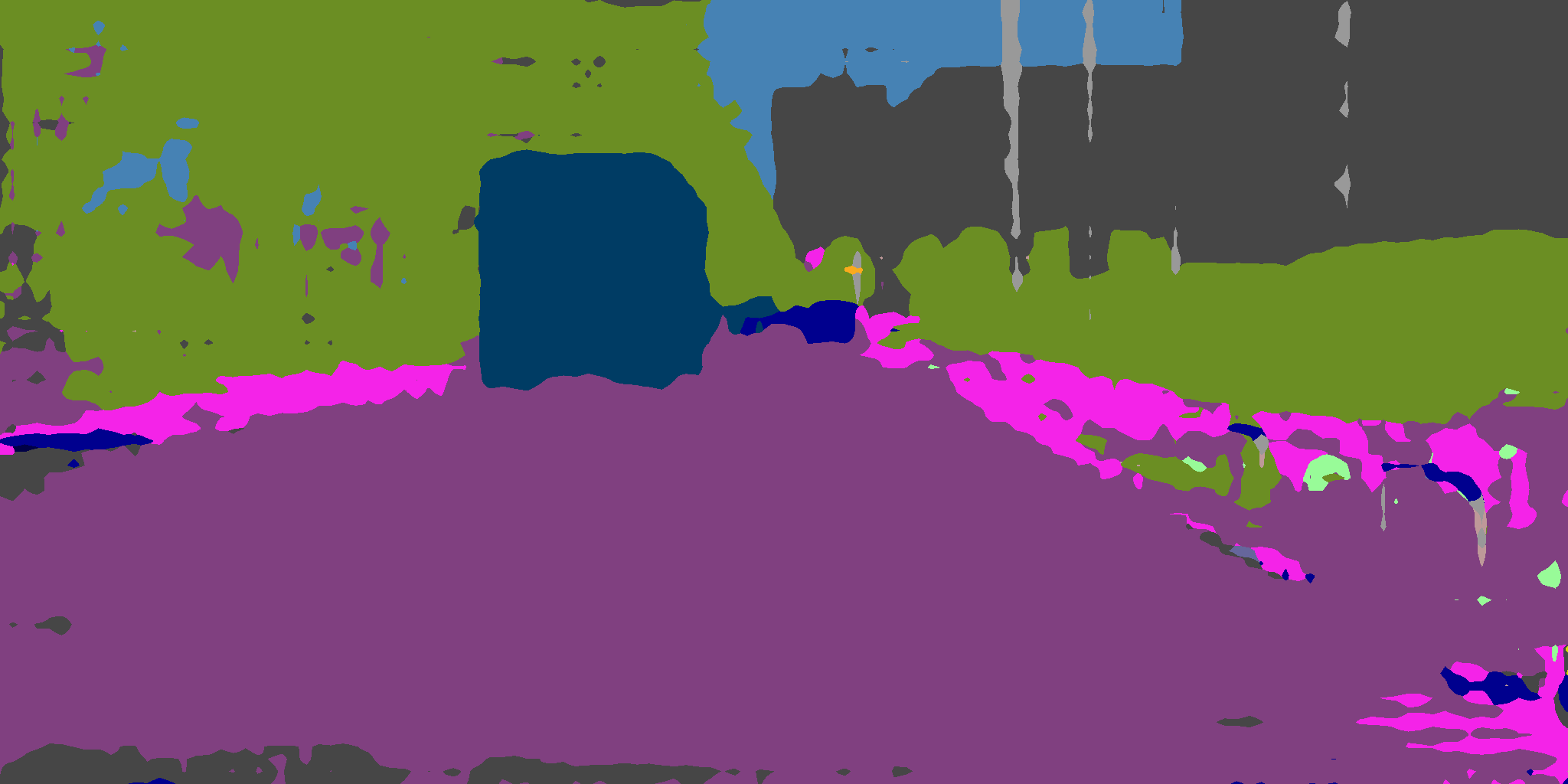} & \hspace{-.45cm}
			\includegraphics[width=.159\textwidth]{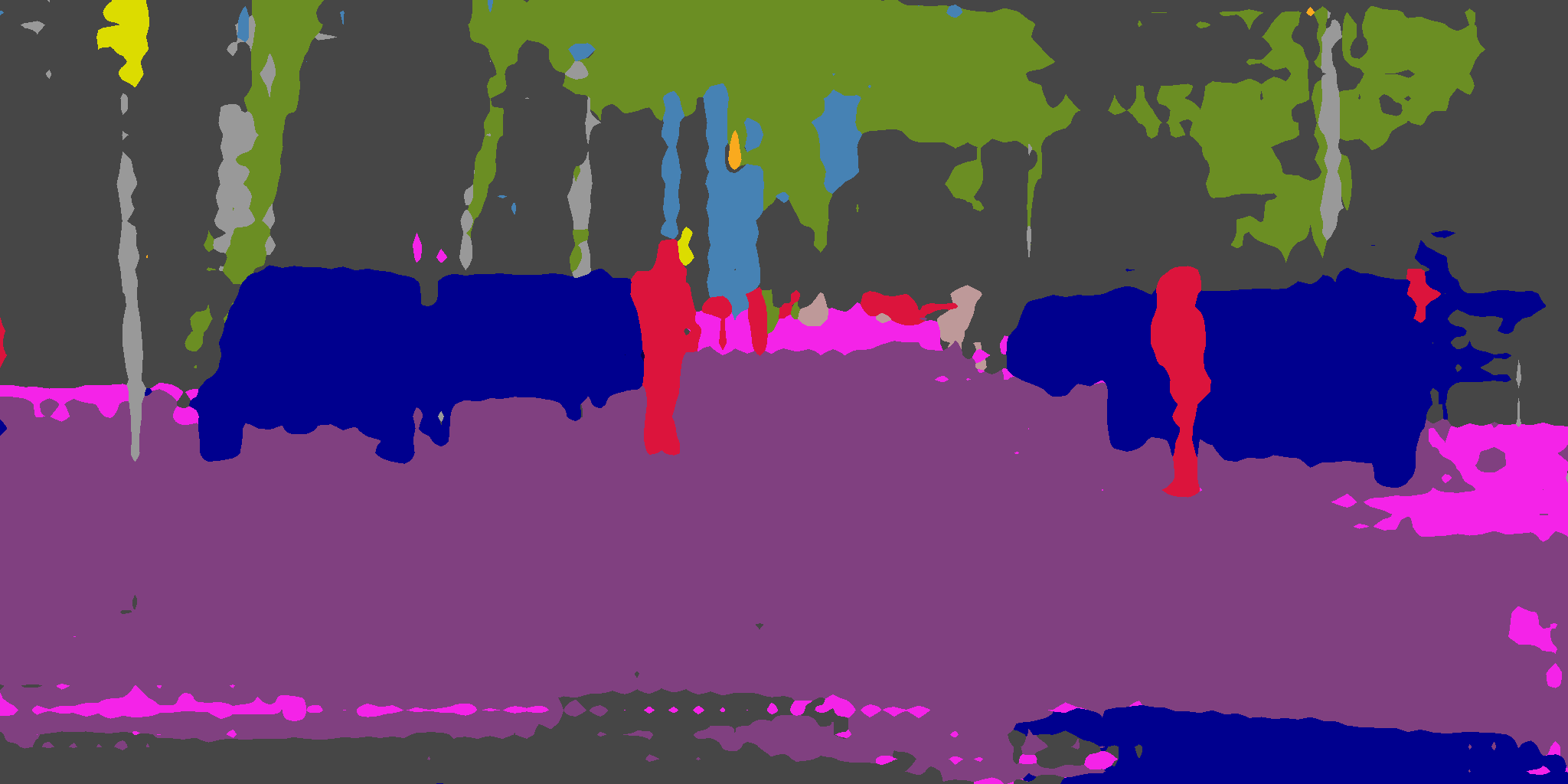} & \hspace{-.45cm}
			\includegraphics[width=.159\textwidth]{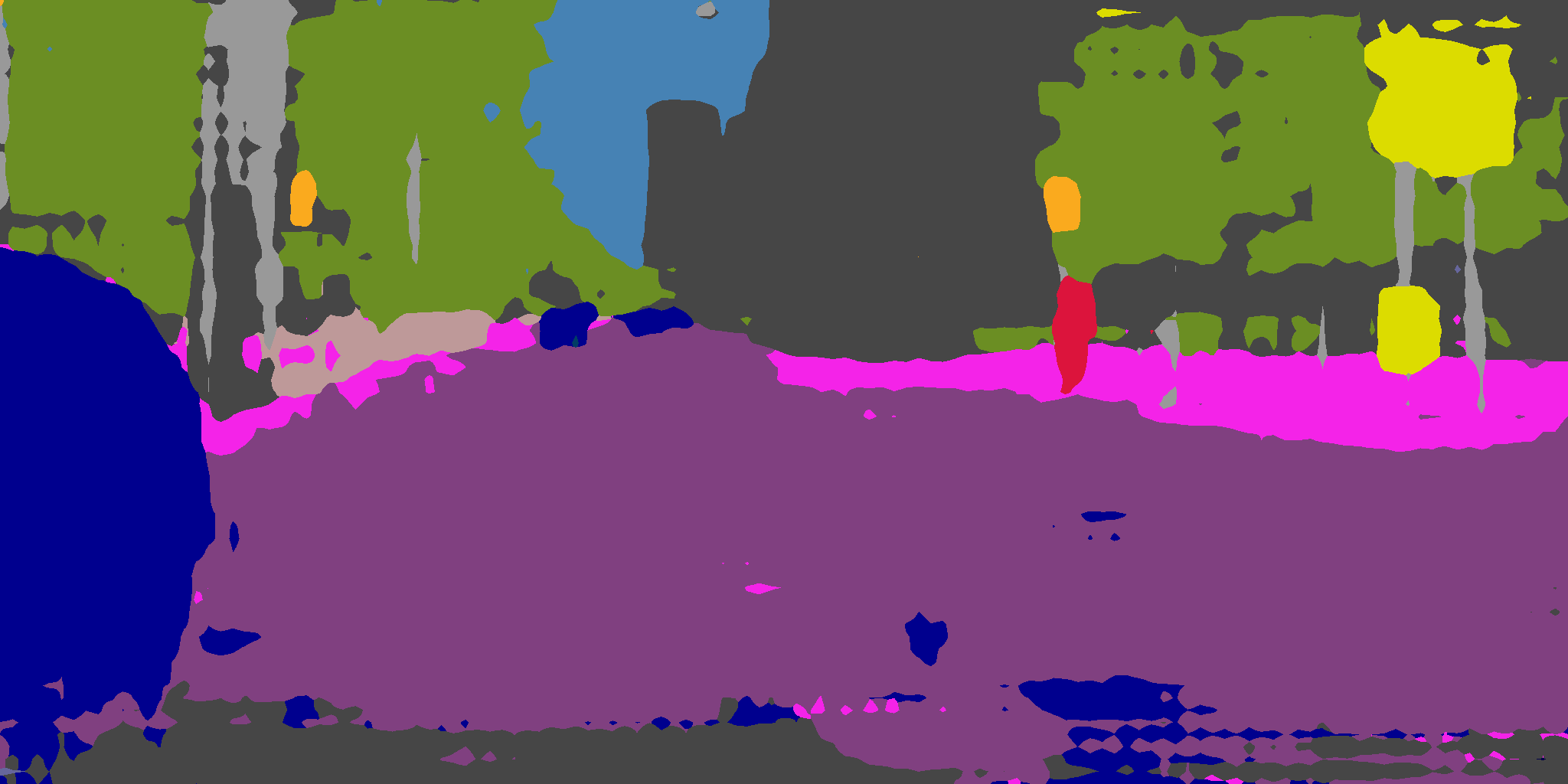} & \hspace{-.45cm}
			\includegraphics[width=.159\textwidth]{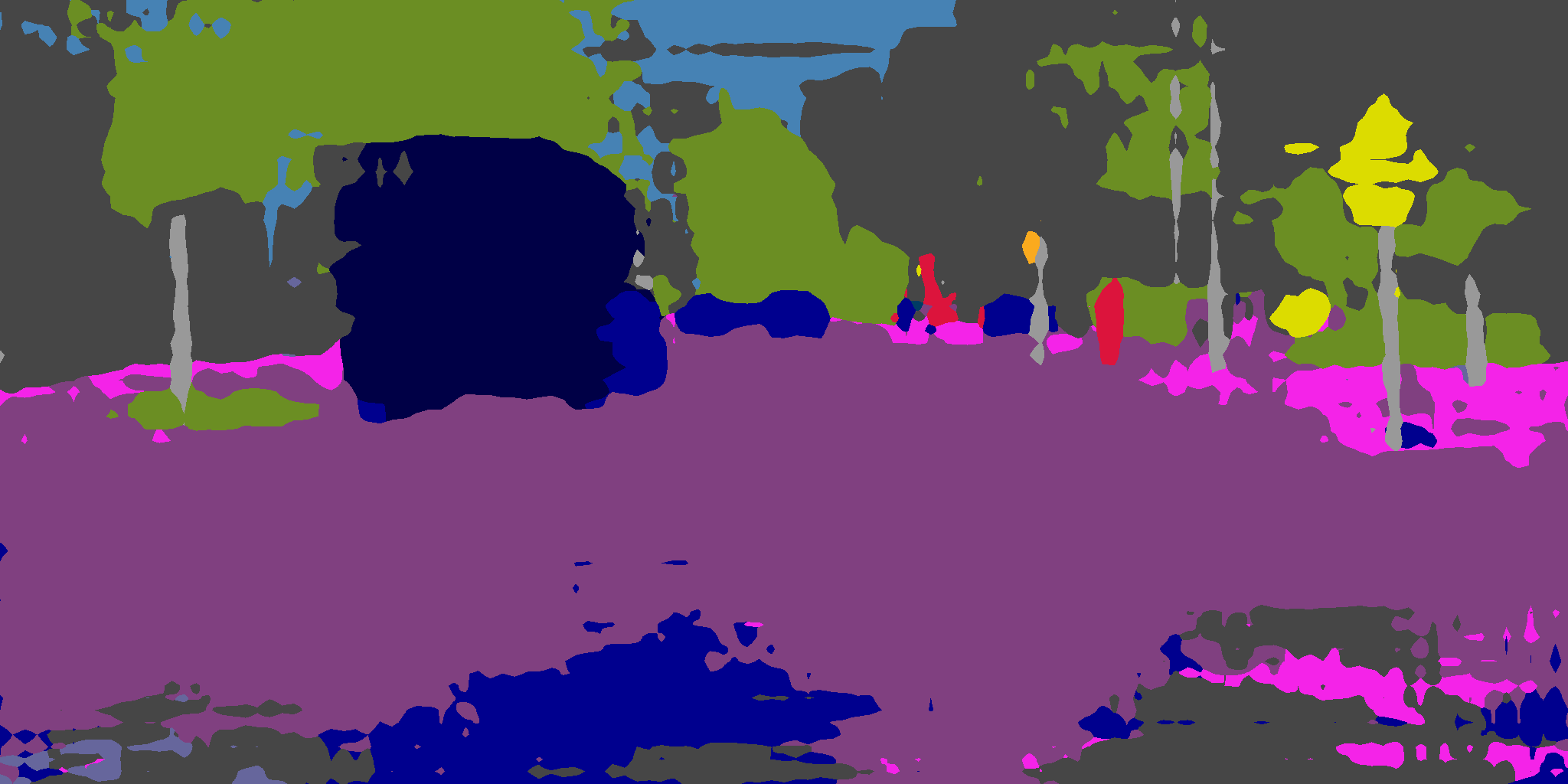} & \hspace{-.45cm}
			\includegraphics[width=.159\textwidth]{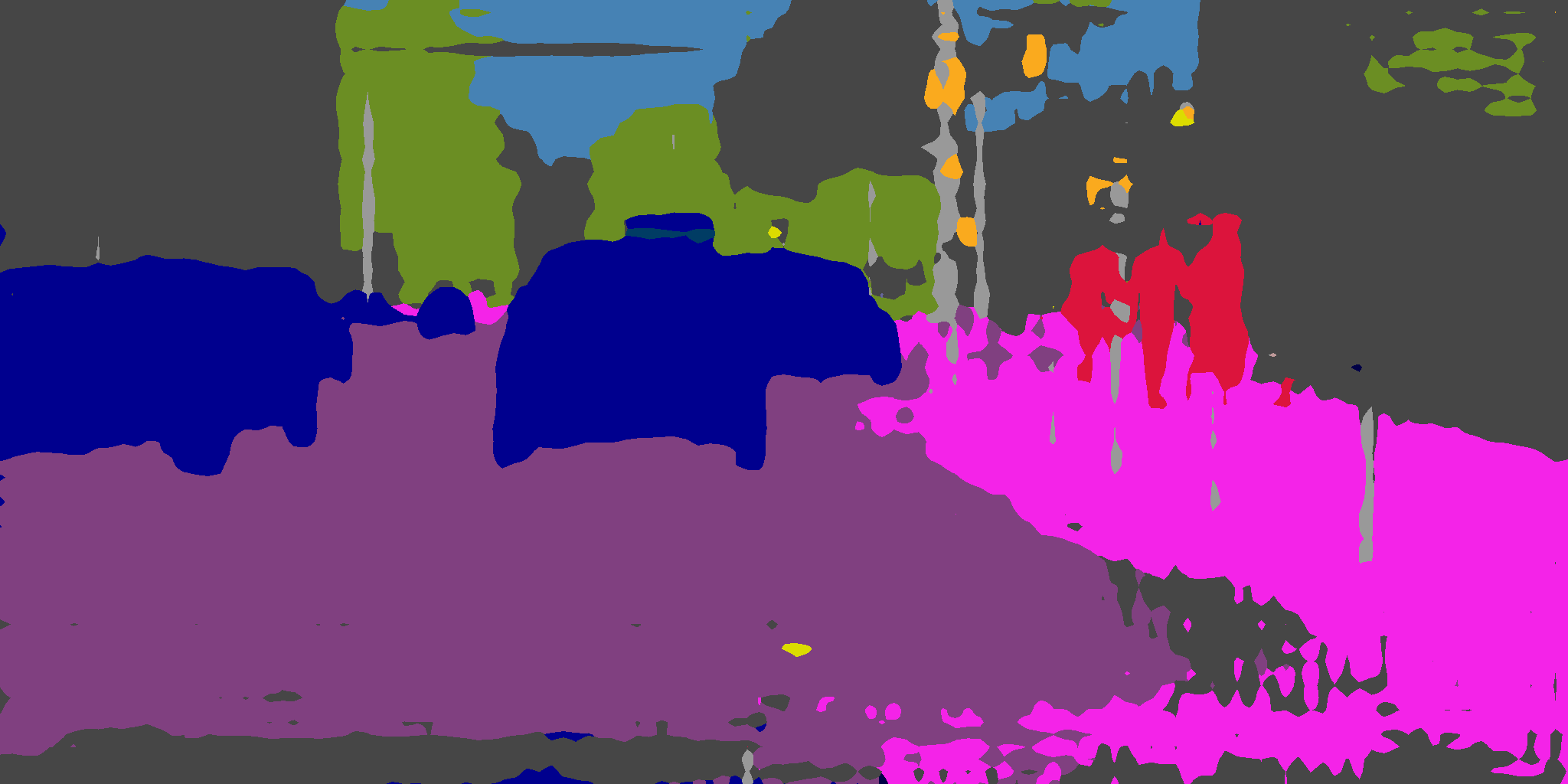} & \hspace{-.45cm}
			\includegraphics[width=.159\textwidth]{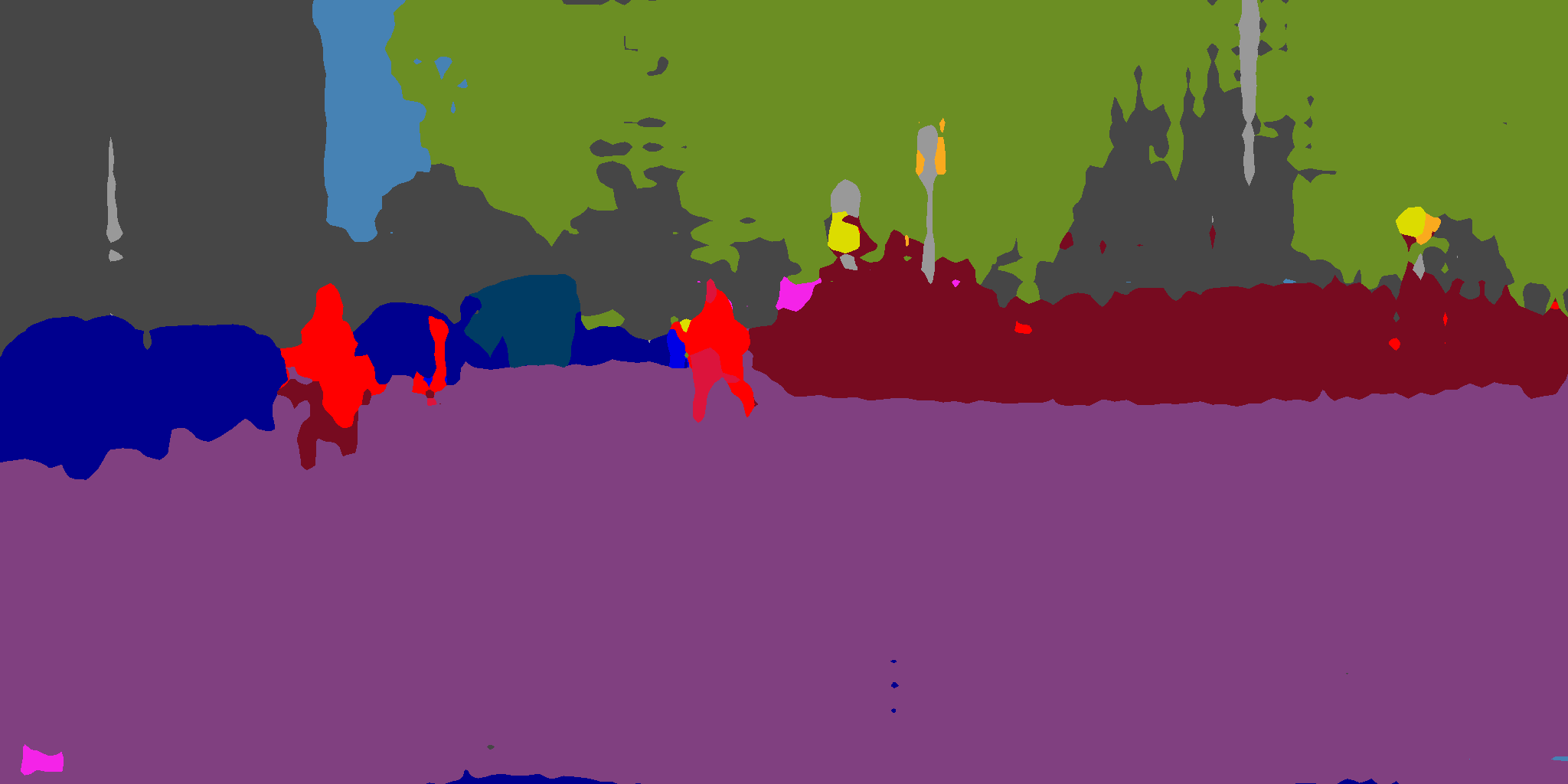} \vspace{-.05cm} \\
			\hspace{-.21cm} \rotatebox{90}{\ \ \footnotesize GT} & \hspace{-.45cm}
			\includegraphics[width=.159\textwidth]{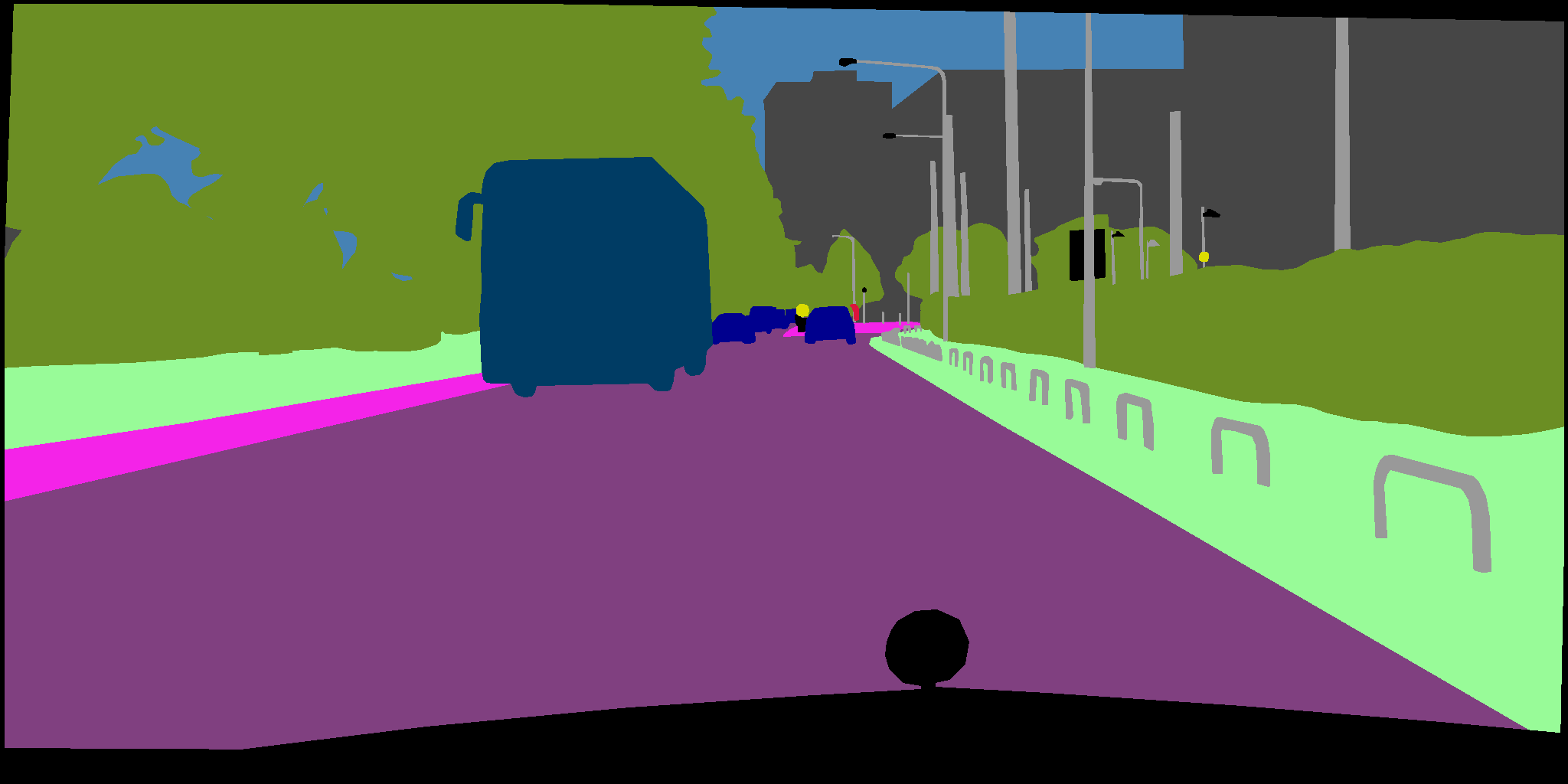} & \hspace{-.45cm}
			\includegraphics[width=.159\textwidth]{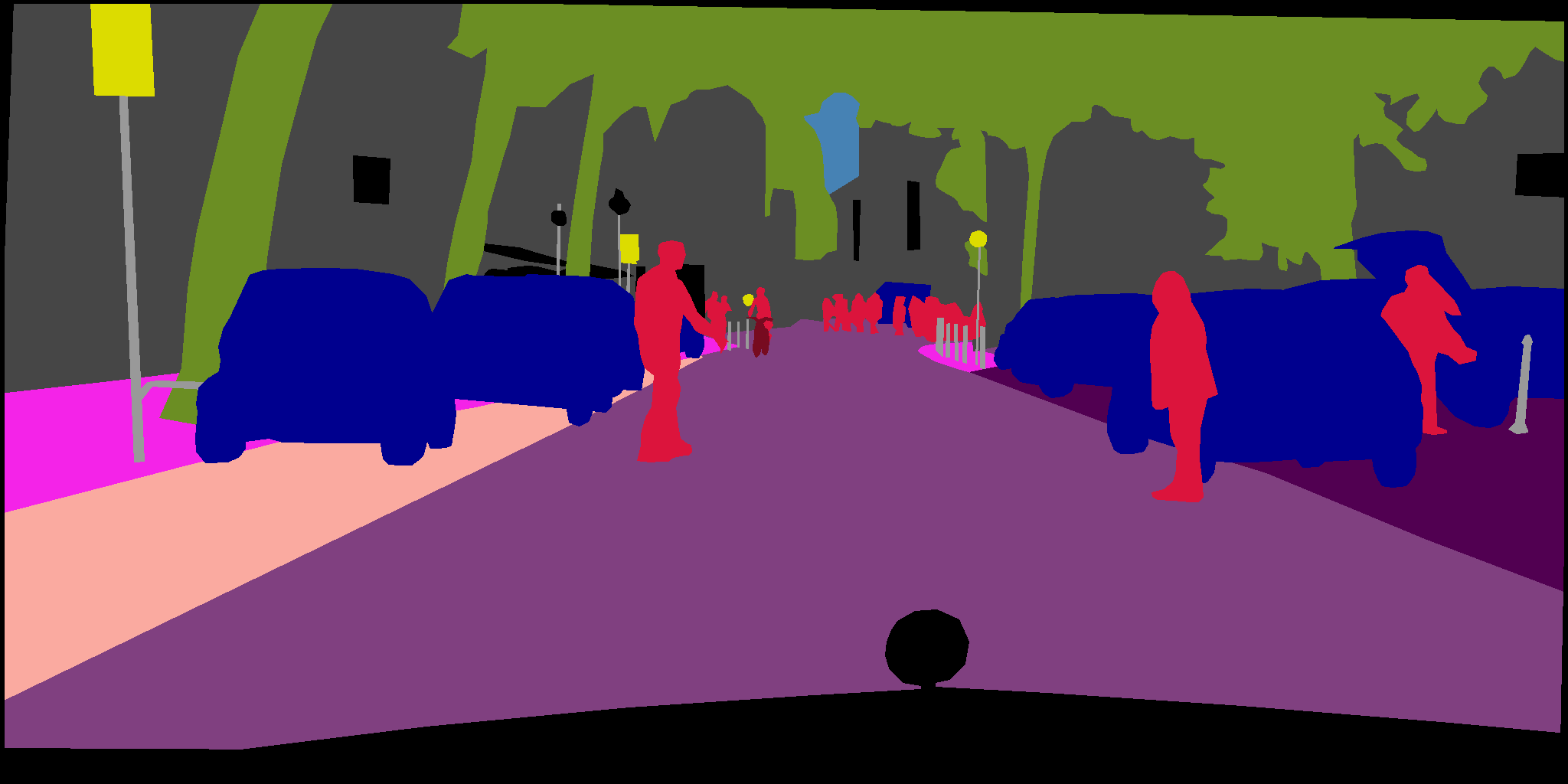} & \hspace{-.45cm}
			\includegraphics[width=.159\textwidth]{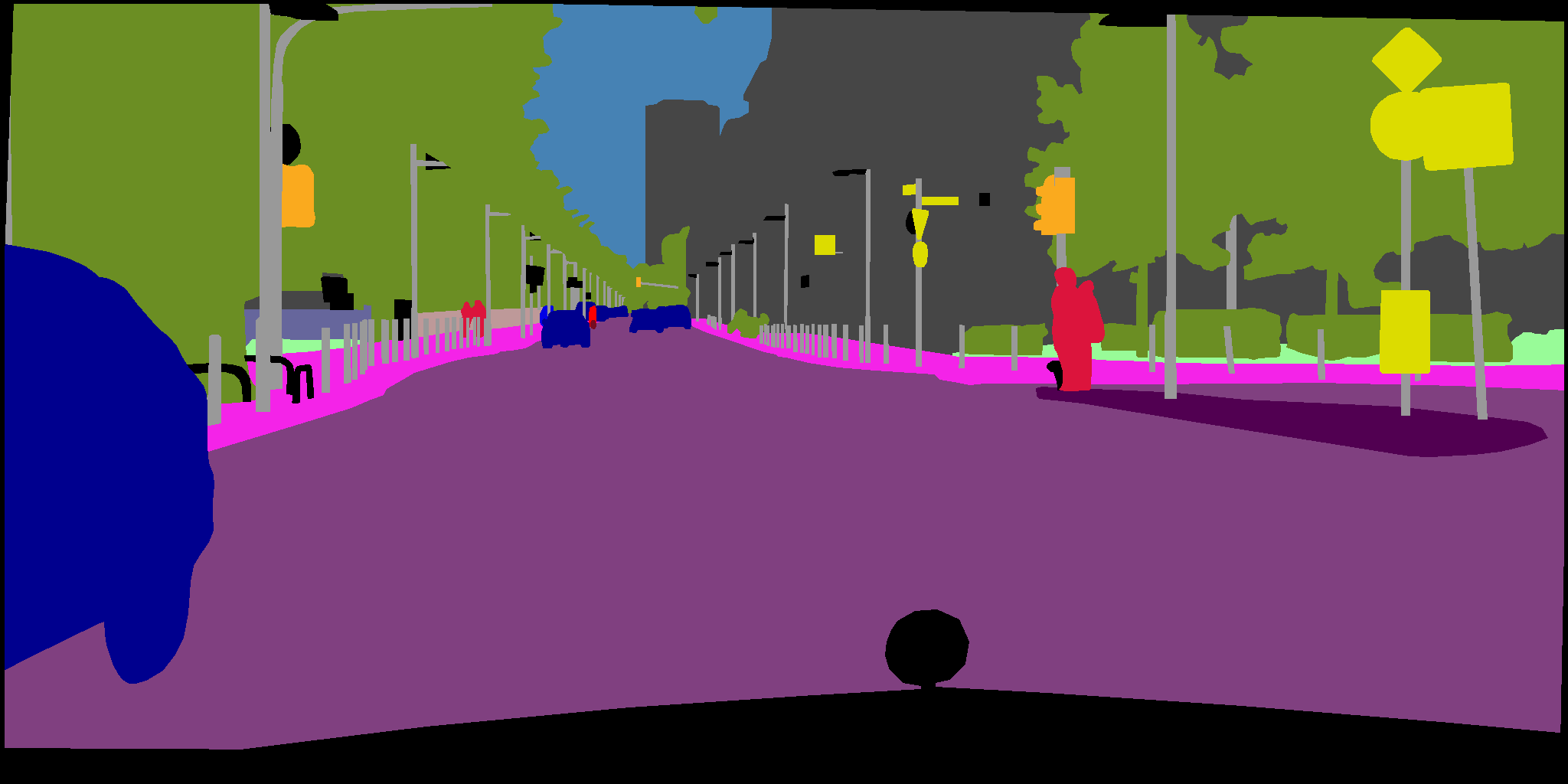} & \hspace{-.45cm}
			\includegraphics[width=.159\textwidth]{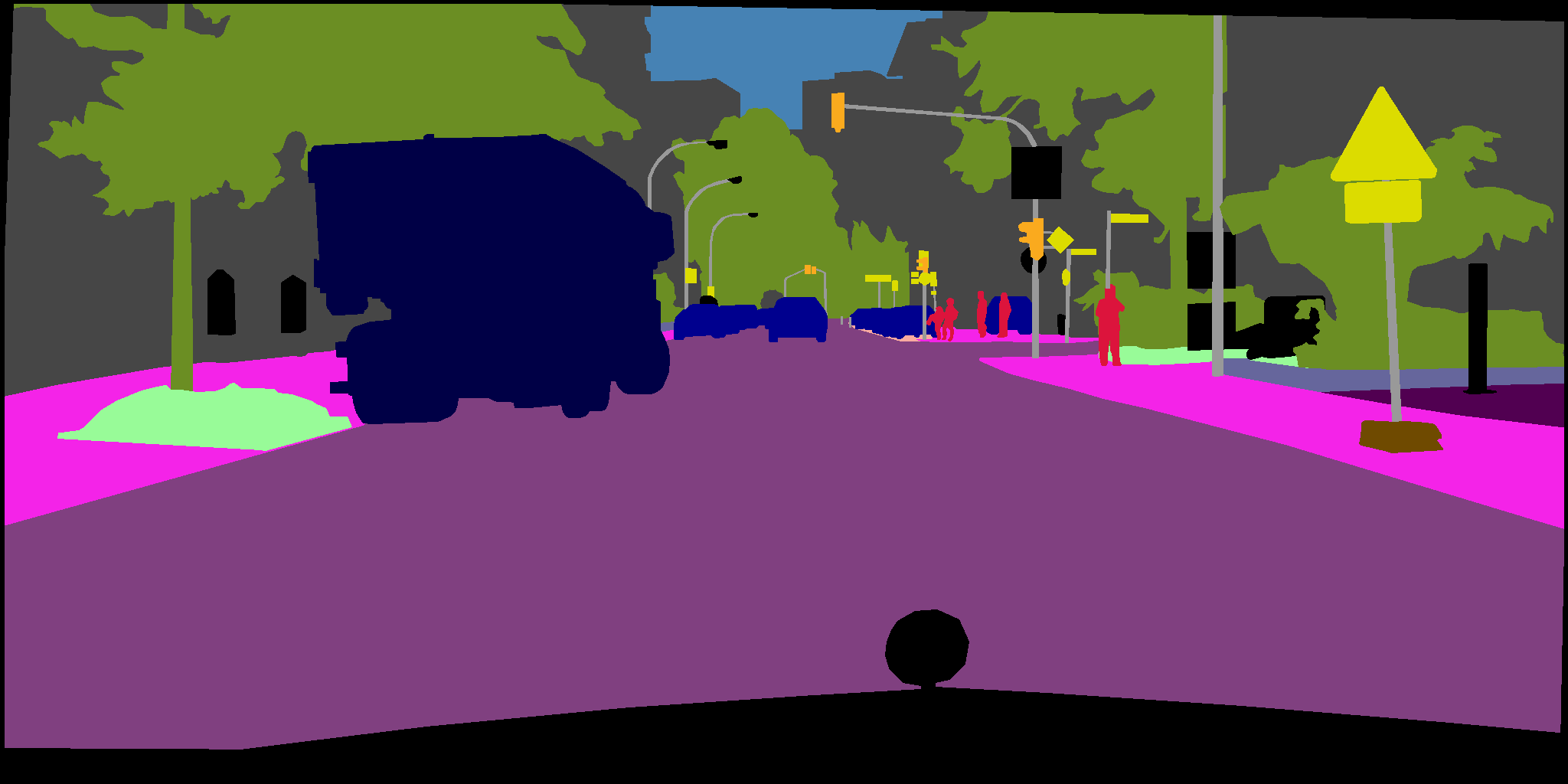} & \hspace{-.45cm}
			\includegraphics[width=.159\textwidth]{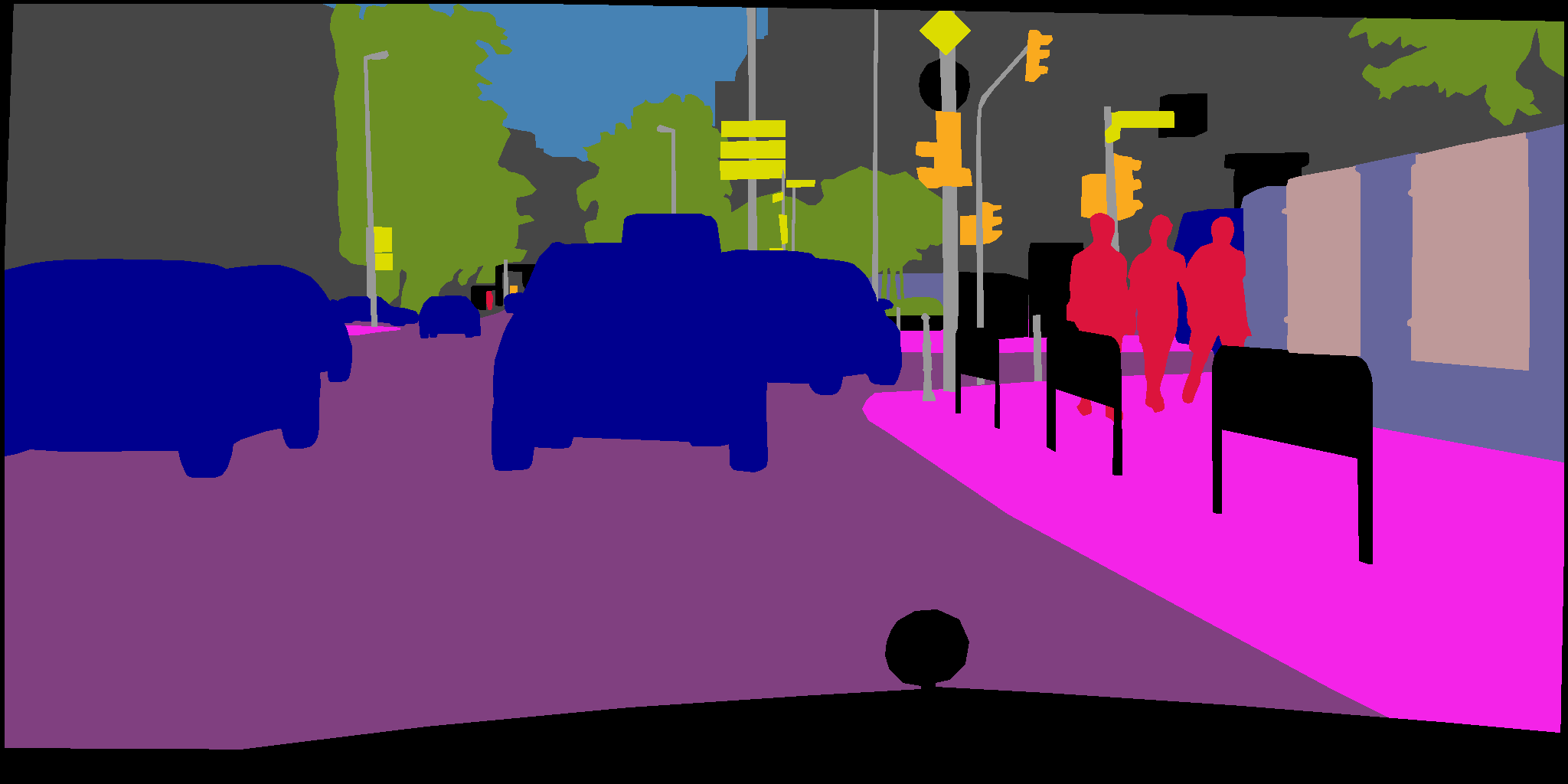} & \hspace{-.45cm}
			\includegraphics[width=.159\textwidth]{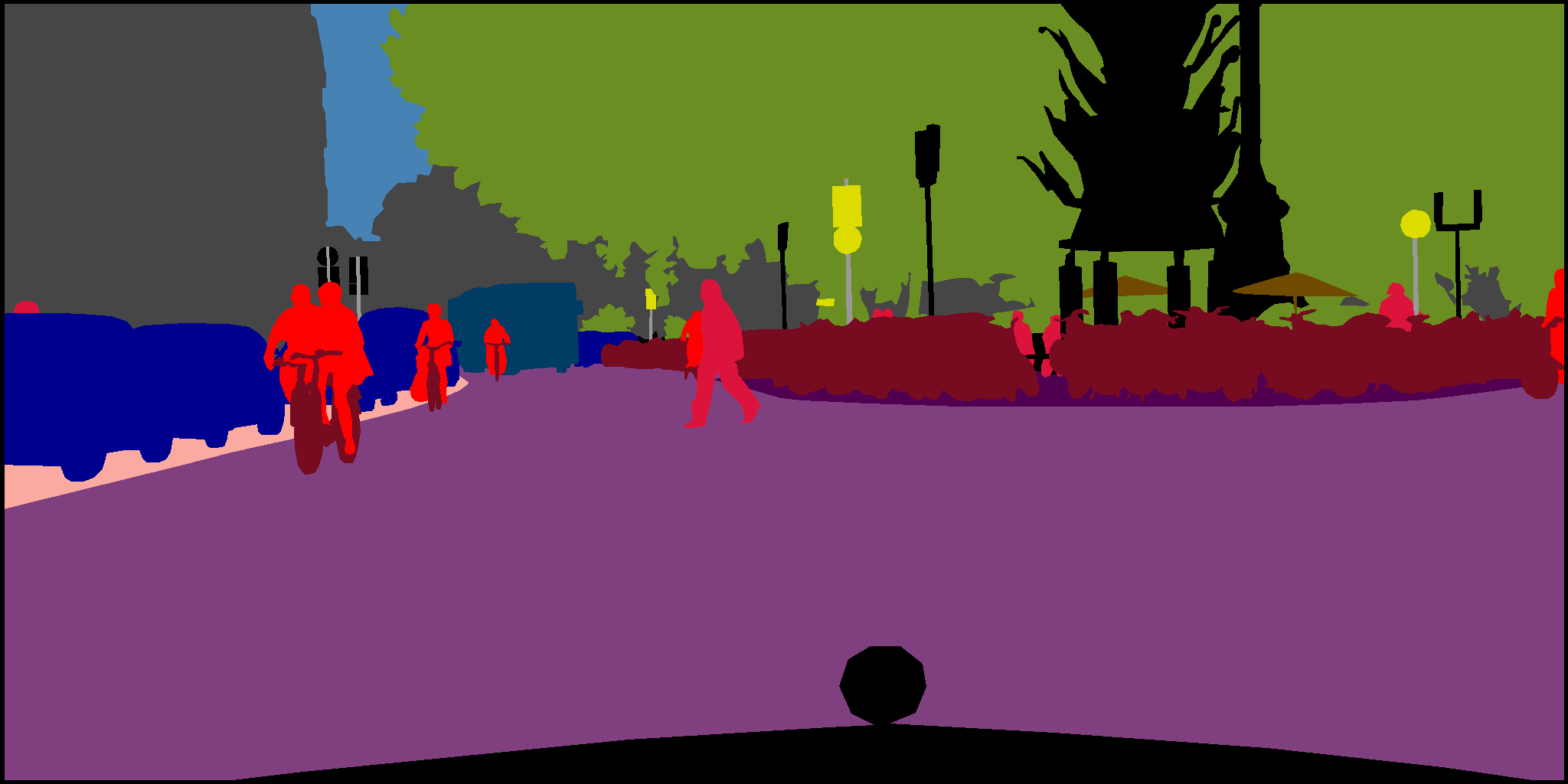} \vspace{-.05cm} \\
			& \hspace{-.45cm}(a) &\hspace{-.45cm}(b)&\hspace{-.45cm}(c)& \hspace{-.45cm}(d)& \hspace{-.45cm}(e) & \hspace{-.45cm}(f)\\
		\end{tabular}
		\caption{Some qualitative comparison results for domain adaptation from SYNTHIA $\rightarrow$ Cityscapes.}
		\label{Qualitative_syn}
	\end{center}
\end{figure*}

We also show qualitative results for GTA5 $\rightarrow$ Cityscapes and SYNTHIA $\rightarrow$ Cityscapes each on 5 samples from the Cityscapes-val set in Fig.~\ref{Qualitative_gta5} and Fig.~\ref{Qualitative_syn}, respectively. It can be observed that our method achieves comparable visualization results as ASM in the two domain adaptation scenarios and even better on some categories such as train (Fig.~\ref{Qualitative_syn}(a)), rider and bicycle (Fig.~\ref{Qualitative_gta5}(c)) and truck (Fig.~\ref{Qualitative_gta5}(d)).

\begin{table*}[htbp]
	\centering
	\caption{ Quantitative comparison results for domain adaptation from Cityscapes to Dark Zurich. The per-category mIoU (\%) of the Dark Zurich validation set are reported. }
	\small
	\label{day2night}
		\begin{tabular}{p{15mm}|p{6mm}|*{18}{p{2.8mm}}p{5mm}|p{6mm}}
			\toprule
			Method &Extra  & \rotatebox{90}{road} & \rotatebox{90}{sidewalk} & \rotatebox{90}{building} & \rotatebox{90}{wall} & \rotatebox{90}{fence} & \rotatebox{90}{pole} & \rotatebox{90}{traffic light \ } & \rotatebox{90}{traffic sign} & \rotatebox{90}{vegetation} & \rotatebox{90}{terrain} & \rotatebox{90}{sky} & \rotatebox{90}{person} & \rotatebox{90}{rider} & \rotatebox{90}{car} & \rotatebox{90}{truck} & \rotatebox{90}{bus} & \rotatebox{90}{train} & \rotatebox{90}{motorcycle} & \rotatebox{90}{bicycle}  & \bf mIoU\\ 
			\midrule
			Source only                            & - &69.5 &12.4 &44.9 &3.0 &18.5 &17.8 &15.6 &7.0 &34.2 &7.4 &8.4 &12.3 &0.8 &22.8 &0.0 &0.0 &0.0 &3.4 &0.2 &14.6\\            
			ASM                 & O+S &75.2 &31.7 &38.6 &7.7 &17.6 &16.9 &12.6 & 4.9 &24.4 &8.0 & 9.8 &16.7 &1.4 &42.9 &0.0 &0.0 &0.0 &7.7 &2.1 &16.7\\
			\bf{Ours}                              & O &63.3 &16.6 &46.5 &5.1 &22.9 &9.0 &15.5 &7.7 &39.7 &10.1 &31.5 &10.2 &0.7 &38.7 &0.0 &0.0 &0.0 &11.0 &4.2 &17.5\\
			\bottomrule
	\end{tabular}
\end{table*}
\begin{figure*}[!ht]
	\begin{center}
		\begin{tabular}{ccccc}
			\hspace{-.21cm} \includegraphics[width=.195\textwidth]{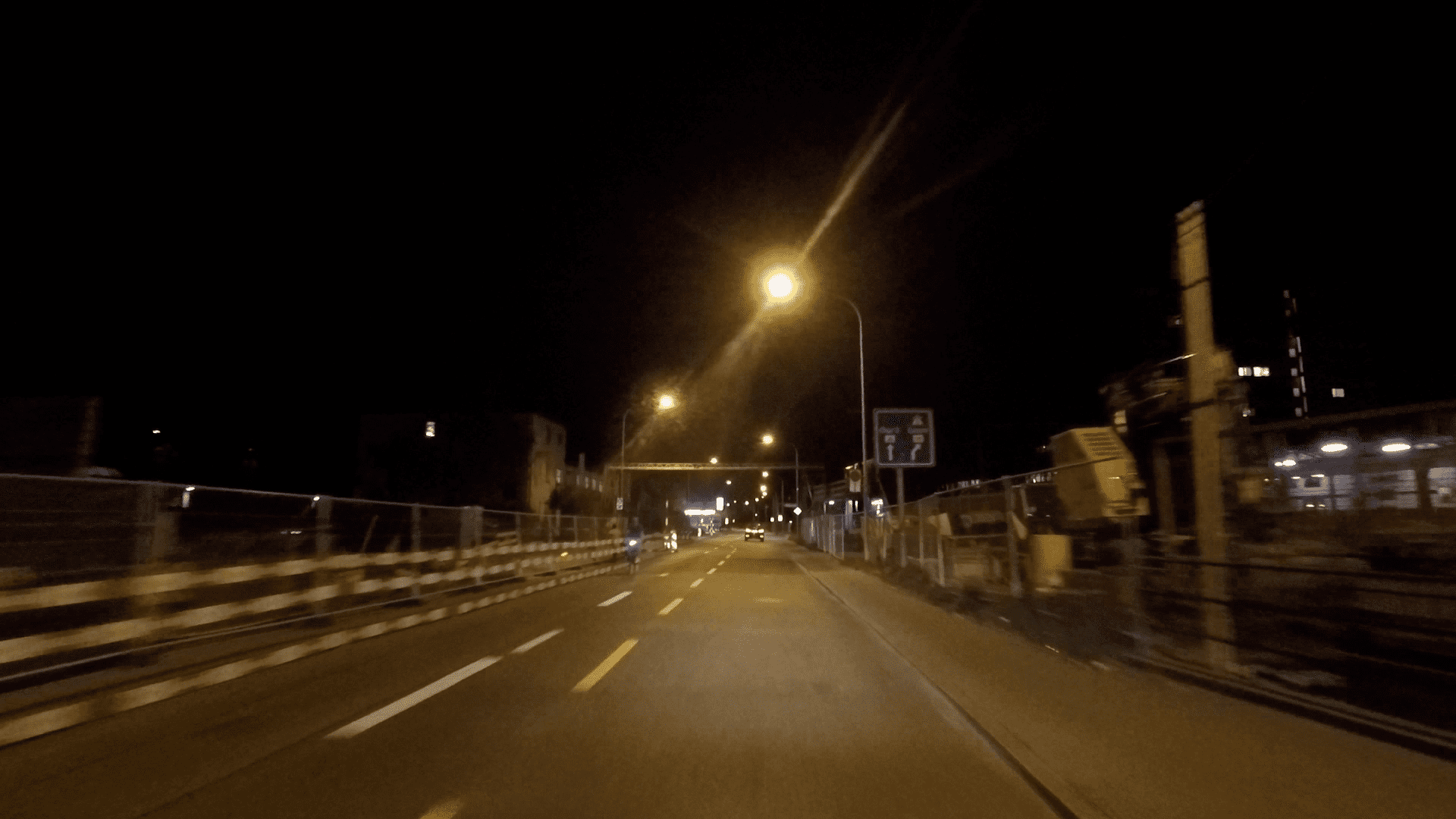} & 
			\hspace{-.45cm} \includegraphics[width=.195\textwidth]{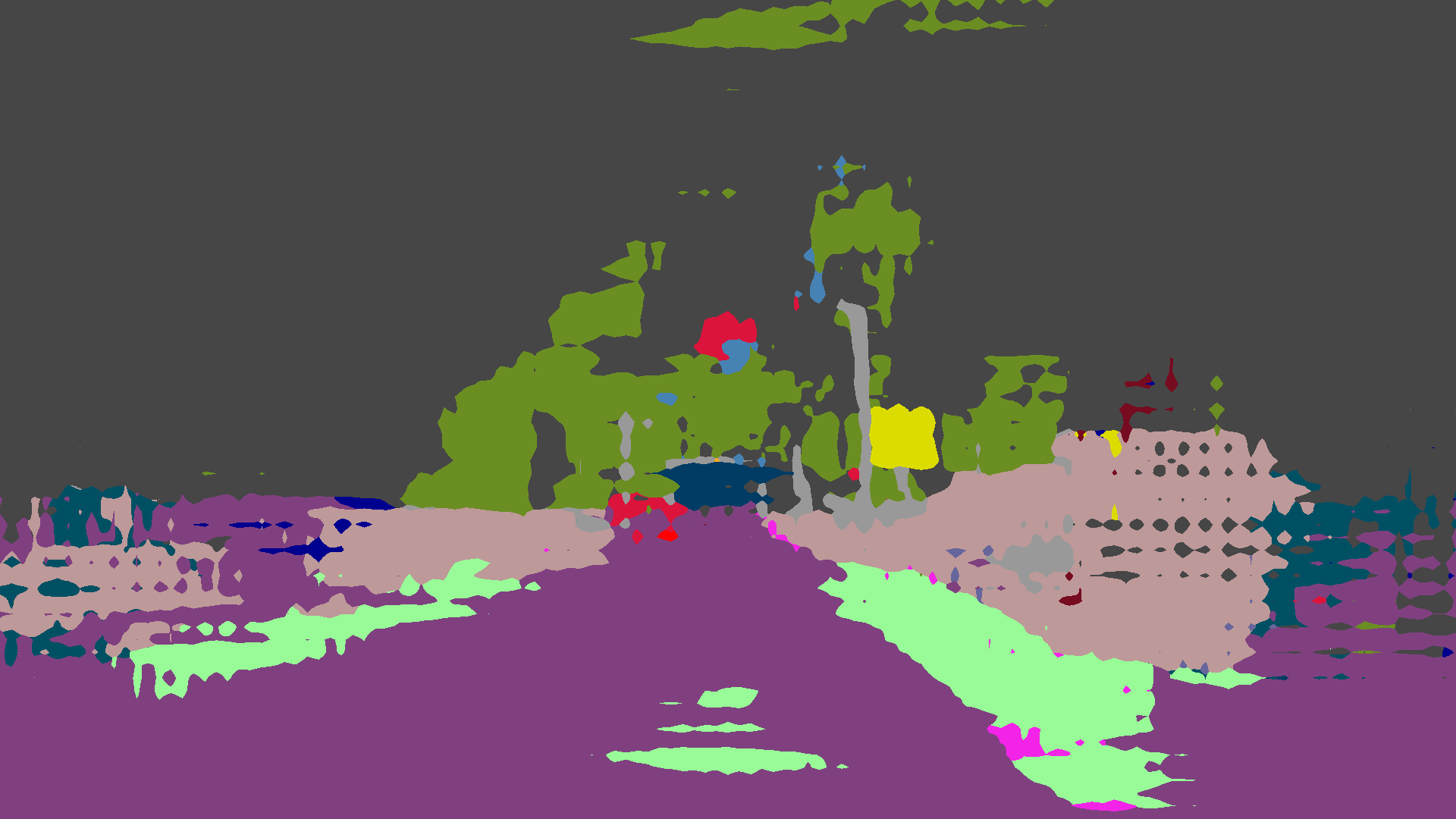} & 
			\hspace{-.45cm} \includegraphics[width=.195\textwidth]{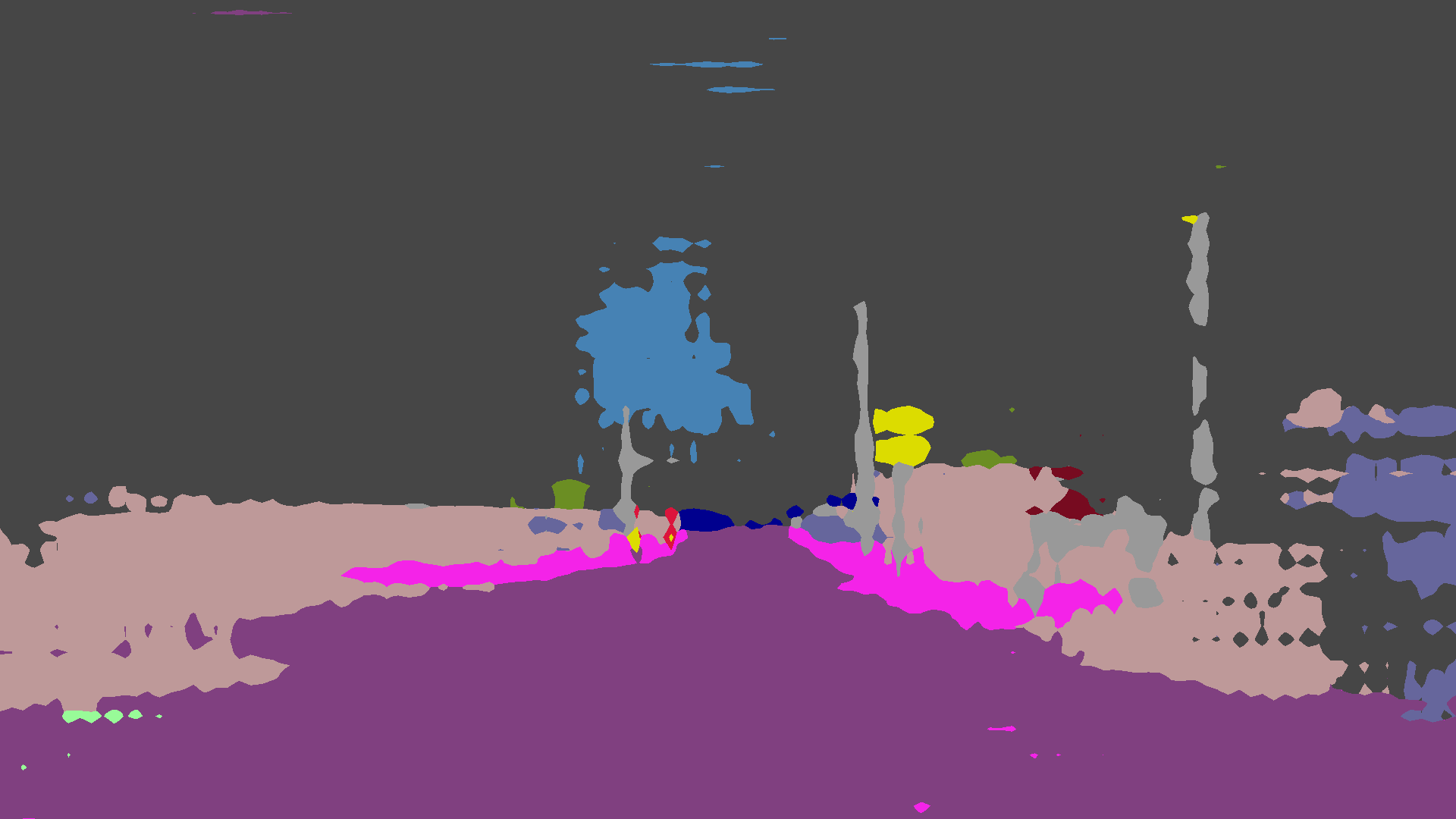} & 
			\hspace{-.45cm} \includegraphics[width=.195\textwidth]{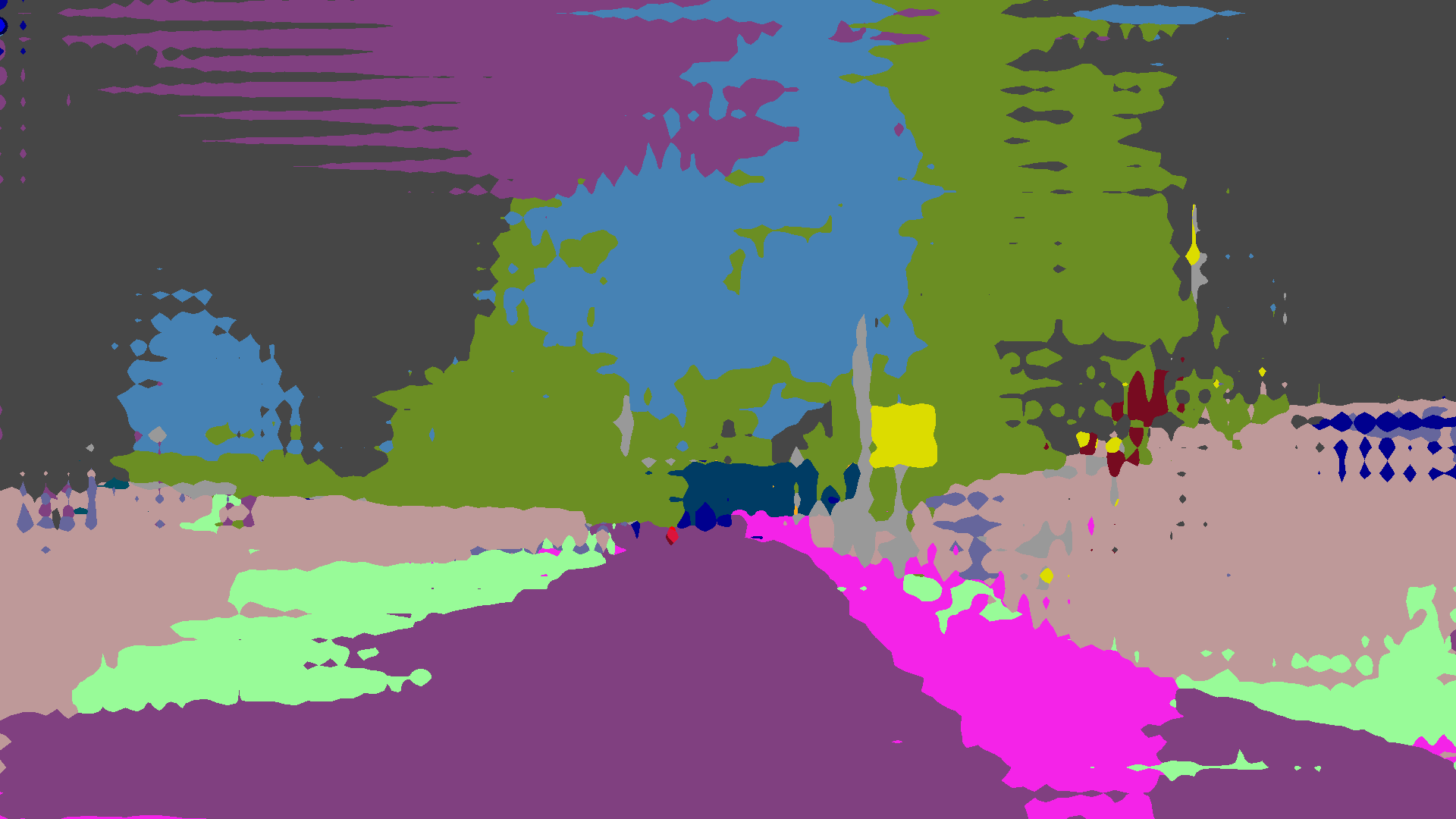} & 
			\hspace{-.45cm} \includegraphics[width=.195\textwidth]{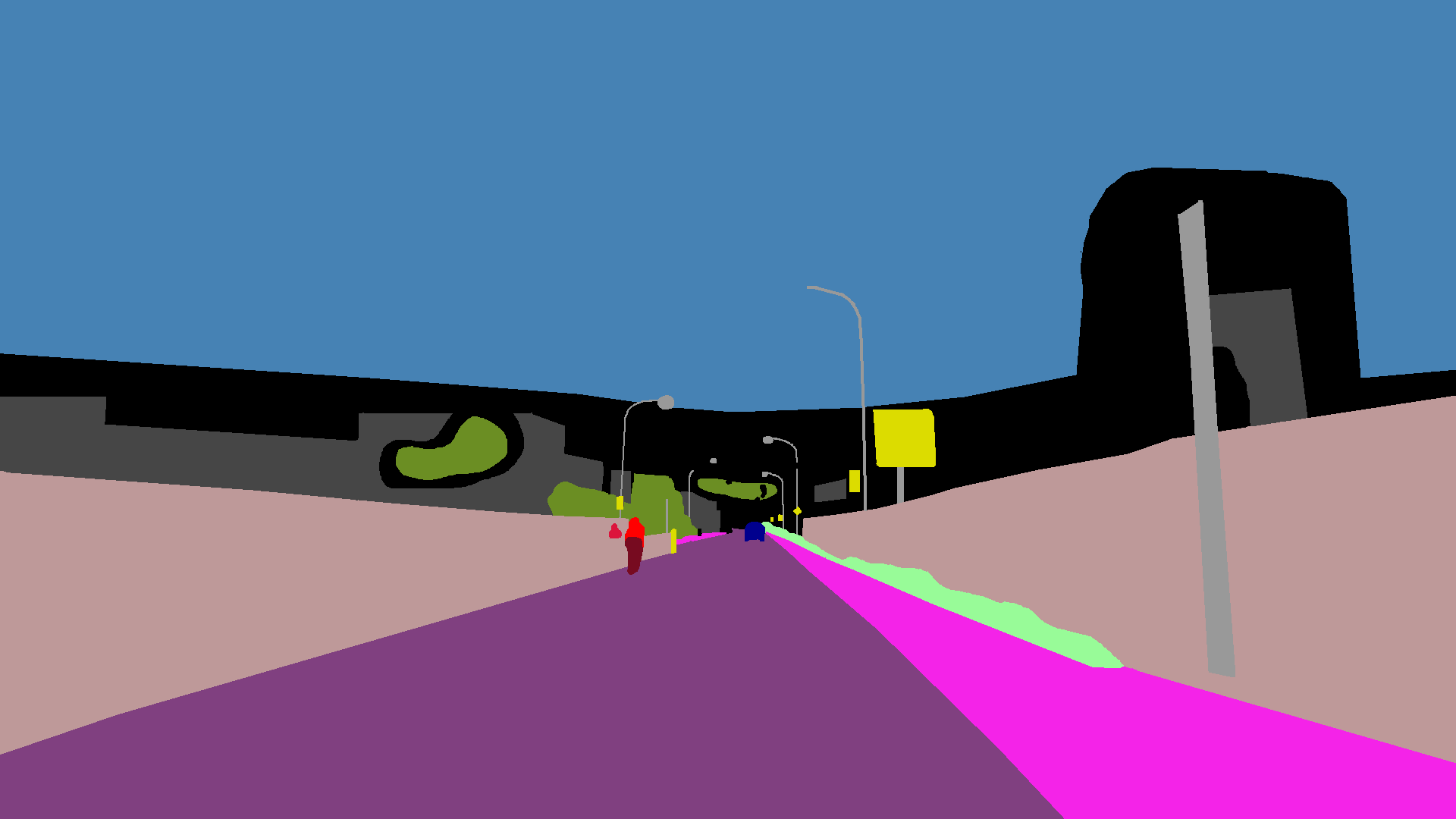} \vspace{-.05cm} \\
			\hspace{-.21cm} \includegraphics[width=.195\textwidth]{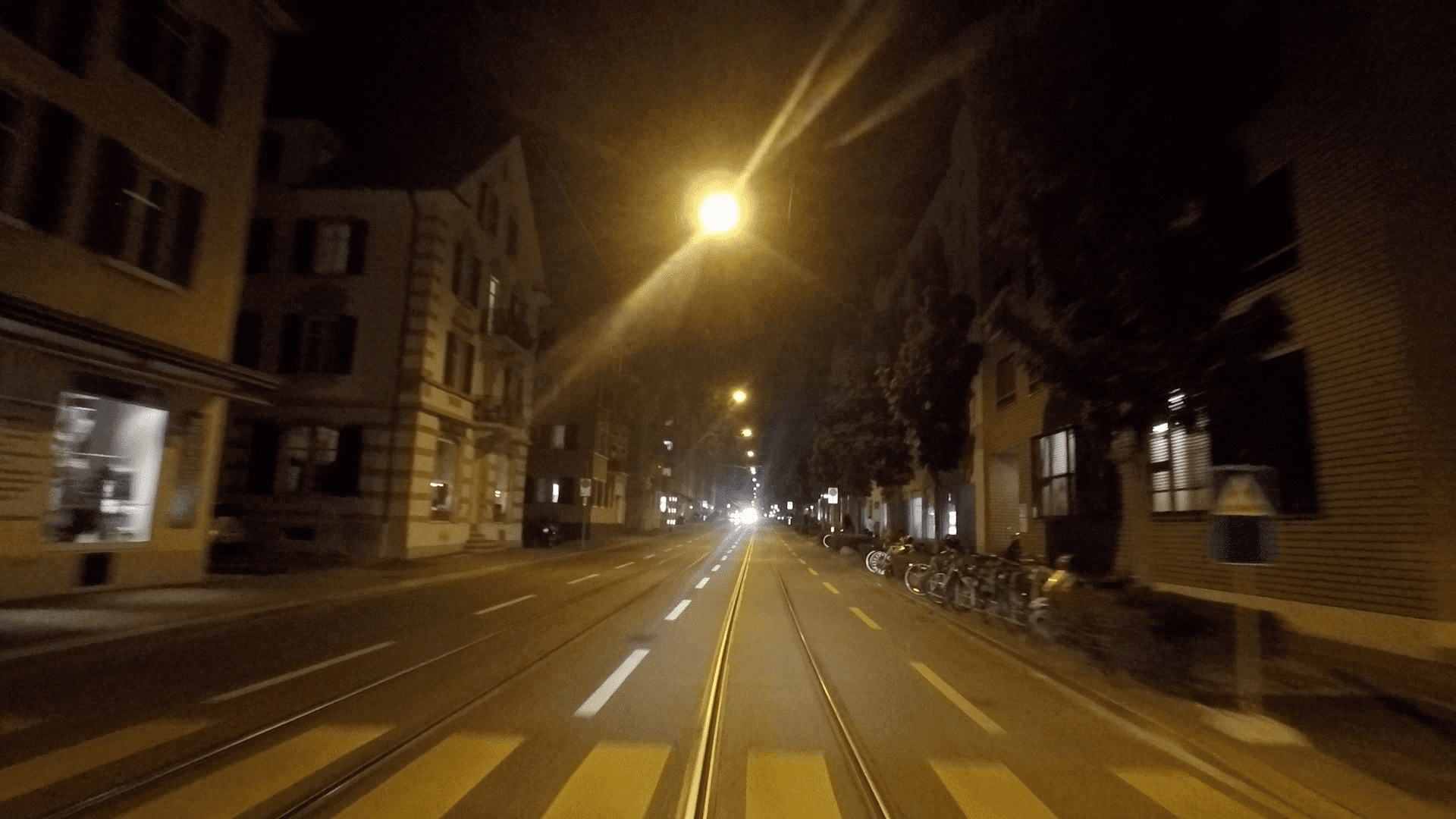} & 
			\hspace{-.45cm} \includegraphics[width=.195\textwidth]{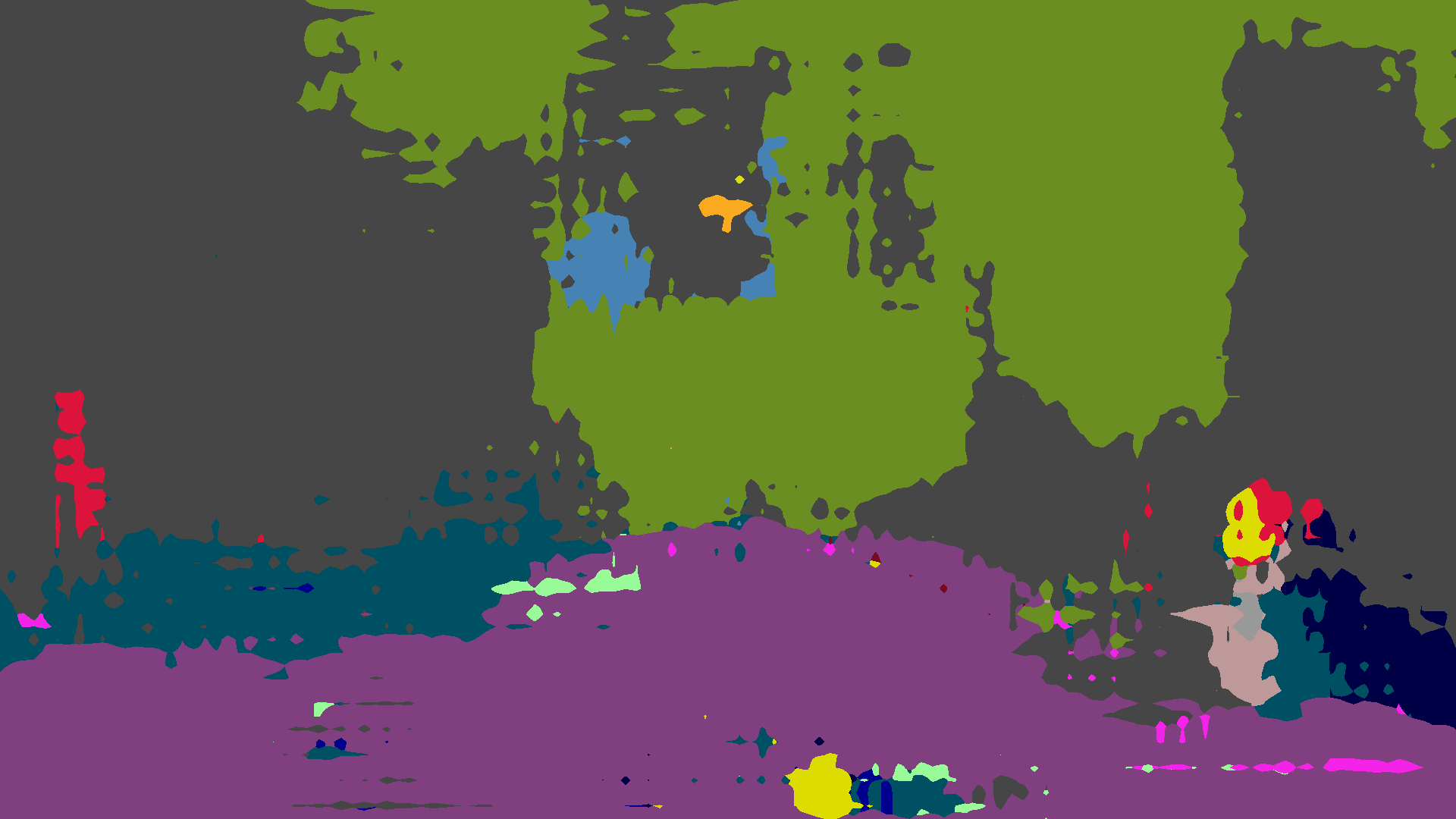} & 
			\hspace{-.45cm} \includegraphics[width=.195\textwidth]{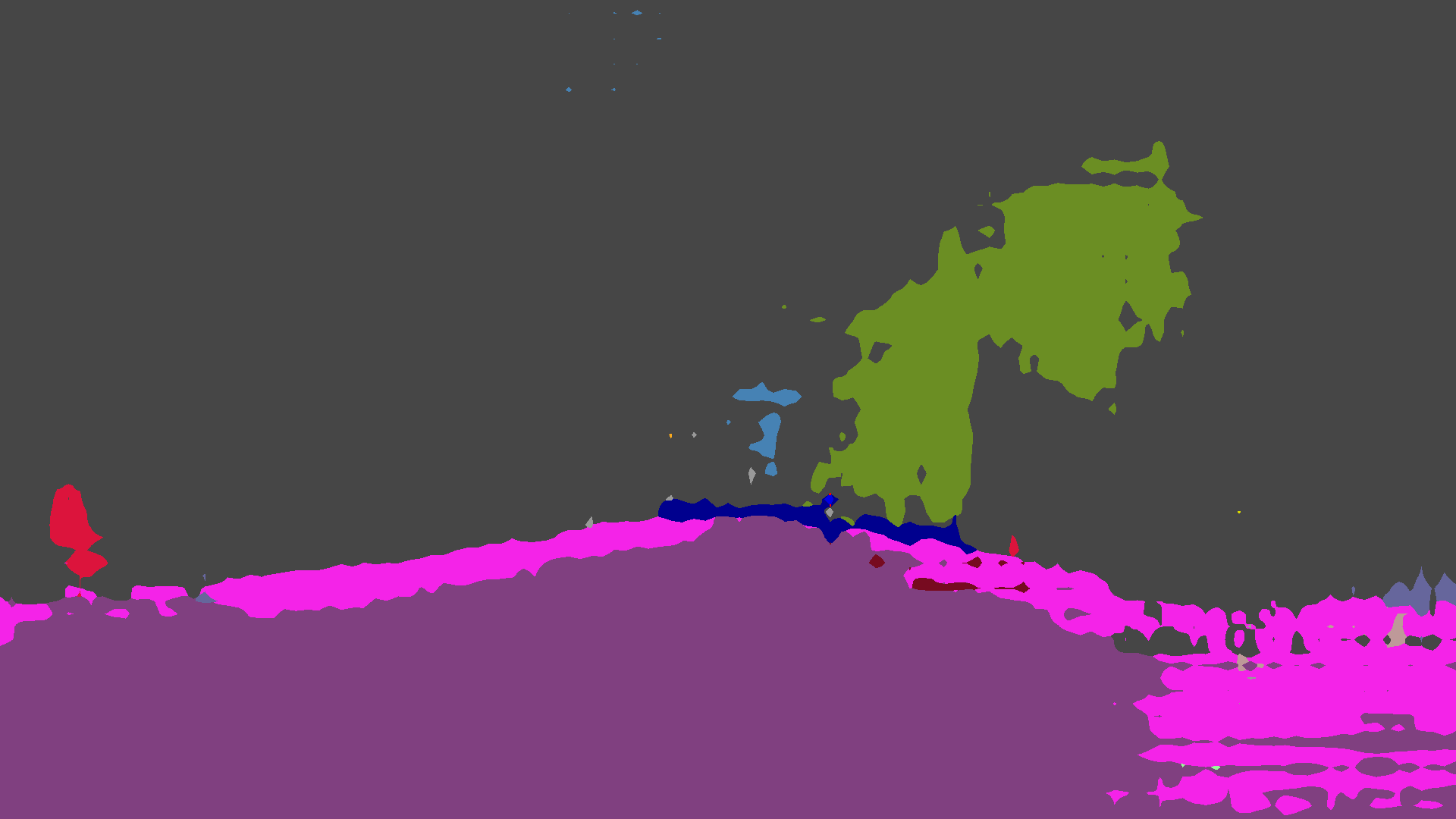} & 
			\hspace{-.45cm} \includegraphics[width=.195\textwidth]{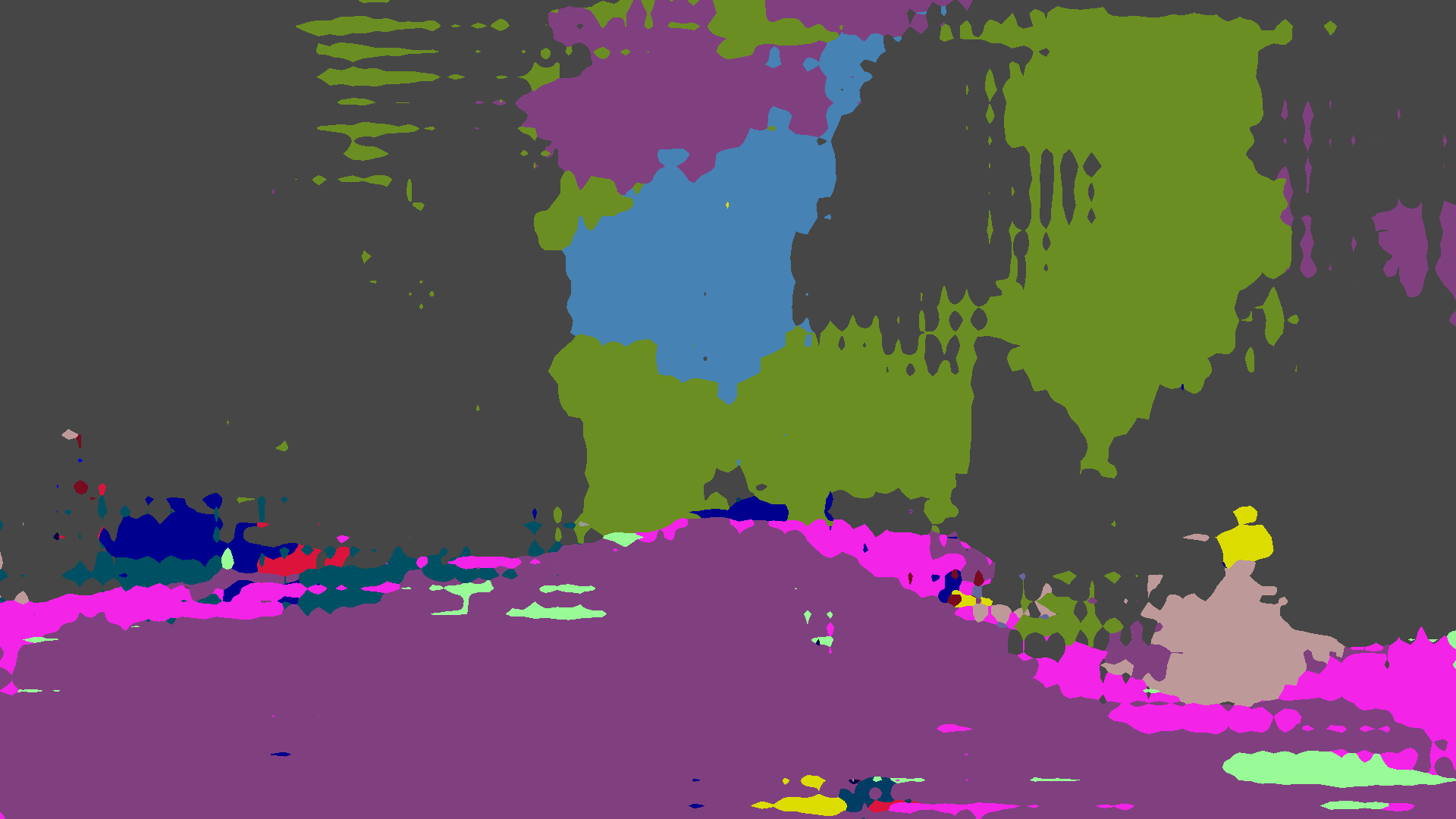} & 
			\hspace{-.45cm} \includegraphics[width=.195\textwidth]{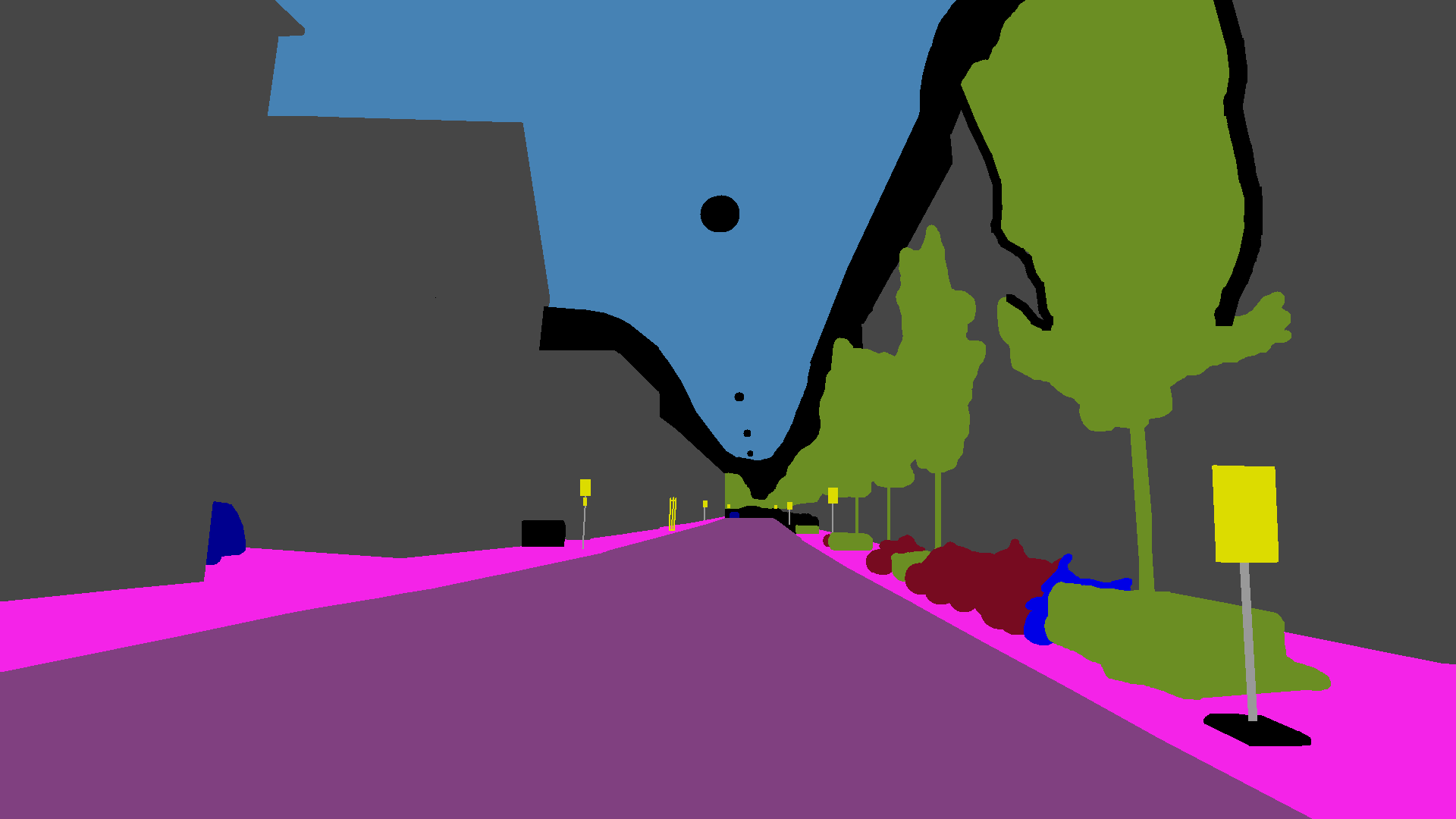} \vspace{-.05cm} \\
			\hspace{-.21cm}Target &\hspace{-.45cm}Source-only &\hspace{-.45cm}ASM&\hspace{-.45cm}Ours& \hspace{-.45cm}GT\\
		\end{tabular}
		\caption{Some qualitative comparison results for domain adaptation from Cityscapes to Dark-Zurich.}
		\label{Qualitative_night}
	\end{center}
\end{figure*}

\subsection{One-Shot Day-to-Night Domain Adaptation}
We further evaluate the proposed method on the more challenging day-to-night setting. In this experiment, we pick Cityscapes~\cite{Cordts2016Cityscapes} as the source and the Dark Zurich~\cite{SDV19} as the target.

The Dark Zurich dataset is carefully collected by Sakaridis {et.al} for unsupervised nighttime semantic segmentation. It consists of 3,041 daytime, 2,920 twilight and 2,416 nighttime images that are all unlabeled and can be used for domain adaptation. There are also 201 labeled nighttime images including 50 images for validation whose labels are provided and the rests serve as an online benchmark. The resolution of all images is 1,920 $\times$ 1,080. 

In our experiments, only one nighttime image is used for domain adaptation and the Dark Zurich validation set is used for performance evaluation. We run this experiment with 4 images (one for each experiment) and 5 times for each image. Finally, we report the average mIoU of the 20 runs computed using the model weights saved in the last running iteration in Table \ref{day2night}. Here, the source-only model is obtained by training the DeepLabV2-Res101~\cite{chen2017deeplab} on the Cityscapes training set for 150K iterations. By applying the source-only model weights, our method can achieve 17.5\% with an additional 500 training iterations. We run ASM with the same 4 images for 50K iterations and compute the average of the 4 experiments as their results. It can be observed that both of the two OSUDA approaches obtain performance gains over the source-only results and our method get better results without training an explicit style transfer model with additional dataset. Some qualitative results are shown in Fig.~\ref{Qualitative_night} where we can see that our method gets better visualization results.

\subsection{Ablation studies}

\textbf{Variants of the loss functions.}
We first investigate several variants of Eq. (\ref{eq5}) as shown in Table.~\ref{ab1:loss}. We directly use the model weights provided by~\citeauthor{Tsai_adaptseg_2018}~\shortcite{Tsai_adaptseg_2018} and fine-tune it with the source data using the standard cross-entropy loss to obtain the source-only results. Note that all these variants are equipped with the original segmentor instead of the style-mixing one. We can observe that the confidence obtained via PPM is the most important component in this equation, without which the mIoU drops 2.66\% on GTA5 $\rightarrow$ Cityscapes and 9.66\% on SYNTHIA $\rightarrow$ Cityscapes. Compared with $\widehat{Conf}$, $E$ has an inconsistent effect on the two experiments. 
\begin{table}[htbp]
	\centering
	\caption{Variants of Eq. (\ref{eq5}) in both GTA5 $\rightarrow$ Cityscapes (G $\rightarrow$ C) and SYNTHIA $\rightarrow$ Cityscapes (S $\rightarrow$ C) scenarios. The mIoU (\%)  scores are reported.}
	\small
	\label{ab1:loss}
		\begin{tabular}{c|cccc}
			\toprule
			Variants & Source-only & w/o $\widehat{Conf}$ & w/o $E$ &  Eq. (\ref{eq5})\\
			\midrule
			G $\rightarrow$ C &36.67 &38.38 &40.95 &\bf41.04 \\
			S $\rightarrow$ C &35.26 &34.26 &\bf44.14 &43.92 \\
			\bottomrule
	\end{tabular}
\end{table}	

\noindent \textbf{Variants of the style-mixing segmentor.}
We study several variants of the style-mixing segmentor as shown in Table.~\ref{ab2:style}. The original Deeplab-V2-Res101 without style-mixing (with PPM) serves as the baseline which can obtain 41.04\% mIoU. Applying the style-mixing layer in image-level ($x^\mathtt S$ only) can obtain 1.39\% for GTA5 $\rightarrow$ Cityscapes and 3.04\% for SYNTHIA $\rightarrow$ Cityscapes. In addition, we try different feature-level style-mixing and we observe that using $f_3^\mathtt S$ only is better than using other levels for both adaptation scenarios. Therefore, we choose to apply the style-mixing layer to both $x^\mathtt S$ and $f_3^\mathtt S$ (Ours). Other combinations might result in similar performance. Compared with the original version of AdaIN, our modified version achieves better performance.
\begin{table}[htbp]
	\centering
	\caption{Variants of the style-mixing segmentor in both GTA5 $\rightarrow$ Cityscapes (G $\rightarrow$ C) and SYNTHIA $\rightarrow$ Cityscapes (S $\rightarrow$ C) scenarios. The mIoU (\%)  scores are reported.}
	\small
	\label{ab2:style}
		\begin{tabular}{p{9mm}|p{5mm}*{6}{p{4.5mm}}p{5mm}}
			\toprule
			Variants & base & $x^\mathtt S$   & $f_1^\mathtt S$ & $f_2^\mathtt S$  & $f_3^\mathtt S$  & $f_4^\mathtt S$  & AdaIN & Ours\\
			\midrule
			G$\rightarrow$C &41.04 &42.43 &40.74 &41.13 &41.16 &40.24 &41.98 &\bf42.77 \\
			S$\rightarrow$C &43.92 &46.96 &43.65 &43.71 &44.63 &43.92 &46.82 &\bf47.33 \\
			\bottomrule
	\end{tabular}
\end{table}	

\noindent \textbf{Variants of the patch size.}
We also study different choices of patch size as shown in Table.~\ref{ab3:patchsize}, where ``no patch'' means that we don't split the target image into patches and use the prototype of the whole image to calculate the confidence map. We find that the patch size 32 performs the best size for both domain  adaptation experiments. 
\begin{table}[htbp]
	\centering
	\caption{Variants of the patch size for both GTA5 $\rightarrow$ Cityscapes (G $\rightarrow$ C) and SYNTHIA $\rightarrow$ Cityscapes (S $\rightarrow$ C) scenarios. The mIoU (\%) scores are reported.}
	\small
	\label{ab3:patchsize}
		\begin{tabular}{c|ccccc}
			\toprule
			Patch size & 8 & 16 & 32 & 64 & no patch\\
			\midrule
			G $\rightarrow$ C & 42.43 & 42.30 &\bf 42.77 &42.38 &42.26\\
			S $\rightarrow$ C &44.83 &45.91 &\bf 47.33 &45.52 &46.03 \\
			\bottomrule
	\end{tabular}
\end{table}	

\begin{figure}[h]
	\centering
	\includegraphics[width=0.3\textwidth]{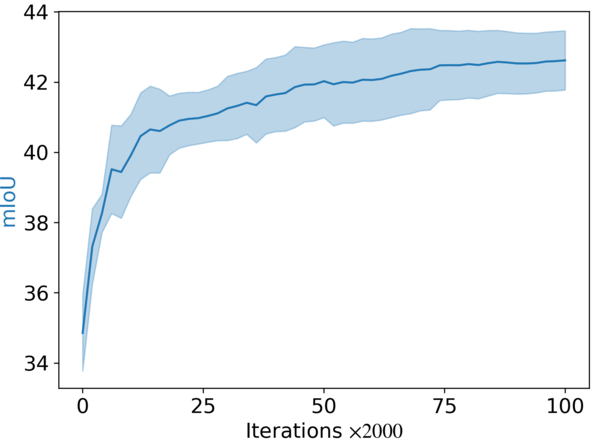}
	\caption{ The mIoU (\%) performance over varying adaptation iterations without using pertrained model for GTA5 $\rightarrow$ Cityscapes.}
	\label{fig:SCRATCH}
\end{figure}

\noindent \textbf{Ablation Study on the Pre-training Model} We study the effect of the usage of the pre-training model. From Fig.~\ref{fig:SCRATCH}, we find that our method can still obtain similar mIoU results without using the pre-training model for GTA5 $\rightarrow$ Cityscapes with more training iterations. Compared with ASM, our method does not need additional dataset and time to train a style-transfer model and uses fewer adaptation iterations with a pre-trained source-only model. And our method can save the memory usage, for example, it only needs about 10G GPU memory while ASM requires around 25G.
\section{Conclusion}
In this paper, we have developed a novel method for the challenging one-shot semantic segmentation in unsupervised domain adaptation. The proposed style-mixing segmentor has the ability to explore more styles around the target sample and perform the semantic segmentation at the same time. This implicit style transfer based on feature-level statistics can significantly reduce memory usage and improve the efficiency of domain adaptation. In addition, patchwise prototypical matching, which is proposed for relieving the negative adaptation and weighting more the positive adaptation, is also shown to be very effective for this task. Various experiments demonstrate that our method can achieve better or comparable results to the current state-of-the-arts in the one-shot setting with much fewer iterations. 

\section{Acknowledgments}
Dr. Lili Ju's work is partially supported by U.S. Department of Energy, Office of Advanced Scientific Computing Research through Applied Mathematics program under grant DE-SC0022254.

\bibliography{egbib} 

\end{document}